\documentclass[11pt]{article}

\usepackage[preprint]{acl}

\usepackage{times}
\usepackage{latexsym}

\usepackage[T1]{fontenc}

\usepackage[utf8]{inputenc}

\usepackage{microtype}

\usepackage{inconsolata}

\usepackage{graphicx}

\usepackage{pifont}
\usepackage{multirow}
\usepackage{arydshln}
\usepackage{algorithm}
\usepackage{algpseudocode}
\usepackage{amsmath}
\usepackage[normalem]{ulem}

\usepackage{booktabs}
\usepackage{multirow}
\usepackage[table,xcdraw]{xcolor}
\usepackage{bbold}
\newcommand{\metricid}[1]{\scalebox{0.85}{\textbf{#1}}}

\title{\textit{Ready Jurist One}: Benchmarking Language Agents for Legal Intelligence \\ in Dynamic Environments}

\author{\textbf{Zheng Jia\textsuperscript{1}\protect\footnotemark[2]},
 \textbf{Shengbin Yue\textsuperscript{1}\protect\footnotemark[2]}, \\
  \textbf{Wei Chen\textsuperscript{3}},
 \textbf{Siyuan Wang\textsuperscript{4}},
  \textbf{Yidong Liu\textsuperscript{5}},
  \textbf{Zejun Li\textsuperscript{1}},
 \textbf{Yun Song\textsuperscript{6}\protect\footnotemark[1]},   
 \textbf{Zhongyu Wei\textsuperscript{1}\textsuperscript{2}\protect\footnotemark[1]}
 \\
 \textsuperscript{1}Fudan University
\textsuperscript{2}Shanghai Innovation Institute \\
  \textsuperscript{3}Huazhong University of Science and Technology
 \textsuperscript{4}University of Southern California \\
  \textsuperscript{5} Midu Technology
 \textsuperscript{6}Northwest University of Political and Law 
\\
\normalsize\texttt{\{zjia24,sbyue23\}@m.fudan.edu.cn, zywei@fudan.edu.cn}
\normalsize\texttt{
}}

\begin{document}
\maketitle
\begin{abstract}
The gap between existing benchmarks and the dynamic nature of real-world legal practice poses a key barrier to advancing legal intelligence.
To this end, we introduce {\textit{\textbf{J1-E}\scalebox{0.8}{\textbf{NVS}}}}, the first interactive and dynamic legal environment tailored for LLM-based agents. Guided by legal experts, it comprises six representative scenarios from Chinese legal practices at three levels of environmental complexity. 
We further introduce {\textit{\textbf{J1-E}\scalebox{0.8}{\textbf{VAL}}}}, a dual-metric evaluation framework, designed to assess both task performance and procedural compliance across varying levels of legal proficiency. 
Extensive experiments on 17 LLM agents reveal that while many models demonstrate solid legal knowledge, they struggle with procedural execution in dynamic settings. Even the SOTA model is below 60\% overall performance \. These findings highlight persistent challenges in achieving dynamic legal intelligence and offer valuable insights to guide future research. \textit{Resources will be available upon acceptance.}
\end{abstract}

\section{Introduction}

Although large language models (LLMs) have shown remarkable capabilities across conventional legal NLP tasks, \textit{e.g.}, legal information extraction and statute memorization  ~\cite{yao2022leven,huang2021dependency},
genuine legal intelligence requires agentic capabilities: procedural awareness, multi-stakeholder interaction, and decision-making in evolving environments.
Recent advancements in LLM-based agent frameworks offer a viable solution for the paradigmatic shift from static task completion to dynamic environmental interaction~\cite{yao2022react,mou-etal-2025-agentsense,yue2025synergistic}, paving promising avenues for the development of \textit{general legal intelligence} (GLI).

However, the primary obstacle to achieving GLI lies in the lack of reliable experimental platforms and quantitative evaluation tools to expose current deficiencies and guide optimization efforts. 
Realistic legal services inherently require  \textit{multi-turn interactions} to provide step-by-step guidance for the laypeople, alongside a strict adherence to \textit{procedural legality}. 
In contrast,
As shown in Table~\ref{tab:Comparison Benchmark}, existing benchmarks are confined to a \textbf{static evaluation paradigm}~\cite{chalkidis2021lexglue,guha2023legalbench,fei2024lawbench,li2024lexeval,dai2023laiw} with several limitations. \textbf{(1) Narrow scenarios}: 
They are restricted to predefined static environments with limited coverage, failing to support sustained agent-environment interactions.
\textbf{(2) Over-simplified tasks}: 
They primarily focus on QA tasks adapted from traditional NLP tasks, lacking the definition of composite targets that reflect real-world legal demands.
\textbf{(3) Restricted evaluation metrics}: 
They focus solely on the accuracy of outcomes, neglecting the requirement of procedural justice inherent in the judiciary. To bridge these gaps, we introduce \textit{the first dynamic evaluation benchmark} for legal intelligence including two integrated modules: the interactive environment and the comprehensive assessment system.

\begin{table}[t]
\resizebox{\linewidth}{!}{
\begin{tabular}{@{}llllll@{}}
\toprule
\textbf{Benchmarks} 
& \textbf{MT?} 
& \textbf{Dyn?} 
& \textbf{Proc?} 
& \textbf{Type} 
& \textbf{Evaluation} \\ 
\midrule

\multicolumn{1}{l|}{LexGLUE}     
& \ding{55} & \ding{55} & \ding{55} 
& NLP task & Rule \\

\multicolumn{1}{l|}{LegalBench}  
& \ding{55} & \ding{55} & \ding{55} 
& NLP task & Rule \\

\multicolumn{1}{l|}{LawBench}    
& \ding{55} & \ding{55} & \ding{55} 
& NLP task & Rule \\

\multicolumn{1}{l|}{LexEval}     
& \ding{55} & \ding{55} & \ding{55} 
& NLP task & Rule \\

\multicolumn{1}{l|}{Law-Eval}    
& \ding{55} & \ding{55} & \ding{55} 
& NLP task & Rule \& LLM \\

\multicolumn{1}{l|}{LAiW}        
& \ding{55} & \ding{55} & \ding{55} 
& NLP task & Rule \& Human \\ 

\multicolumn{1}{l|}
{{\textit{\textbf{J1-E}\scalebox{0.8}{\textbf{VAL}}}}}
& \ding{51} & \ding{51} & \ding{51} 
& Scenario & Rule \& LLM \\

\bottomrule
\end{tabular}
}
\caption{Comparison between our {\textit{\textbf{J1-E}\scalebox{0.8}{\textbf{VAL}}}} and existing legal benchmarks, where \textbf{MT}, \textbf{Dyn}, and \textbf{Proc} denote Multi-turn, Dynamic, and Procedural Process, respectively. 
{\textit{\textbf{J1-E}\scalebox{0.8}{\textbf{VAL}}}} is scenario-based, dynamic, and interactive, aligning better with \textit{real-world legal practices}.}
\label{tab:Comparison Benchmark}
\end{table}

\begin{figure*}[t]
    \centering
\includegraphics[width=0.93\linewidth]{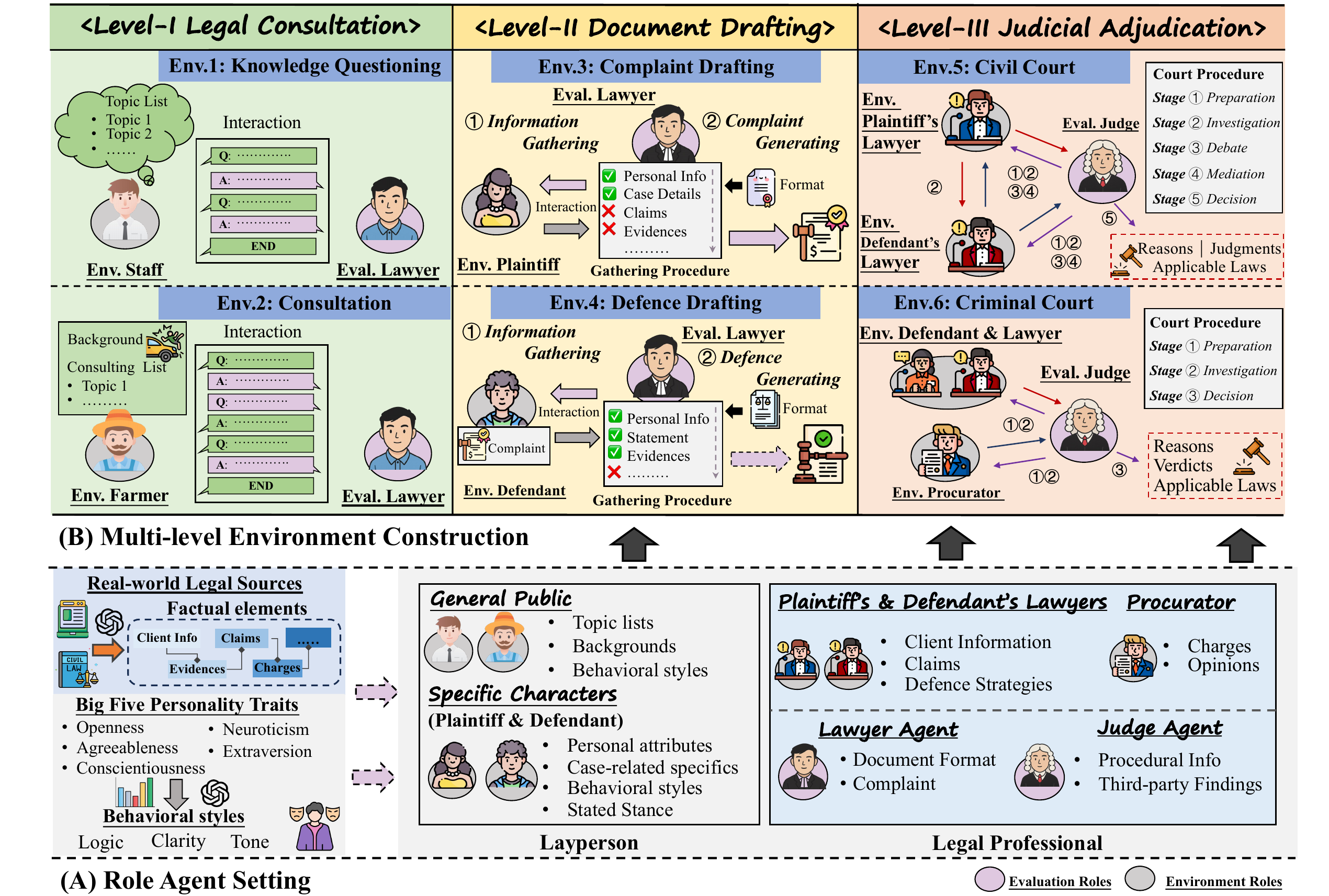}
    \caption{An illustration of {\textit{\textbf{J1-E}\scalebox{0.8}{\textbf{NVS}}}} construction pipeline. (A) Role Agent Setting: We synthesize real-world legal sources and personality theories to construct heterogeneous agents.
(B) Multi-level Environment Construction: We structure these roles within specific procedures and relationships to form environments.}
    \label{fig:model}
\end{figure*}

{\textit{\textbf{J1-E}\scalebox{0.8}{\textbf{NVS}}}}: Interactive Legal Environments. We construct an open-ended legal environment in which agents navigate various legal scenarios through dynamic and procedural interactions. 
As shown in Figure~\ref{fig:model}, we instantiate heterogeneous agents ranging from layperson to legal professionals, each endowed with distinct behavioral traits and initialized with interrelated factual elements from actual cases to ensure high-fidelity simulation.
These agents are then orchestrated into corresponding environments following the structure of legal practice,
which are categorized into three hierarchical levels based on increasing environmental complexity: 
\uline{legal consultation} (knowledge questioning and consultation), 
\uline{document drafting} (complaint and defence drafting), and \uline{judicial adjudication} (civil and criminal courts).

{\textit{\textbf{J1-E}\scalebox{0.8}{\textbf{VAL}}}}: Dual-Metric Assessment System. In order to enable comprehensive assessment, we construct 508 specific interactive environment instances from real-world Legal sources (\textit{Chinese Judgment Documents} and\textit{ Chinese legal Articles}).
Our framework utilizes a dual-metric system consisting of \textbf{Outcome-oriented} and \textbf{Process-oriented } metrics.
The former quantifies the quality of the final target, while the latter scrutinizes adherence to indispensable intermediate constraints, reasoning steps, and procedural norms. 
Assessment is conducted using rule-based or LLM-based methods, with each instance equipped with \textbf{explicit ground truth} for the final objectives, ensuring evaluation reliability and minimizing potential bias introduced by LLM-based evaluators. 
Together, {\textit{\textbf{J1-E}\scalebox{0.8}{\textbf{VAL}}}} can provide deeper insights into the skill sets required for legal task execution, and present a realistic and reliable overview of agent capabilities, facilitating further exploration into the similarities and differences between LLM-based agents and human learning mechanisms.

We perform an extensive assessment of 17 leading LLM agents, spanning proprietary, open-source, and legal-specific models.
Results reveal significant capability gaps in dynamic environments, with performance degrading markedly in complex, multi-agent, and long-horizon settings. Crucially, these failures stem not from knowledge deficits, but from limitations in procedural adherence and role interaction. Our analysis confirms that process-oriented competence is essential for valid outcomes, yet current models frequently misalign procedural execution with final objectives. 

\textbf{Contributions:} \textbf{(A)} We construct \textit{\textbf{J1-E}\scalebox{0.8}{\textbf{NVS}}}, the first open-ended interactive legal environment, orchestrating heterogeneous agents across three hierarchical stages and enabling dynamic agent–environment interaction.
\textbf{(B)} We introduce \textit{\textbf{J1-E}\scalebox{0.8}{\textbf{VAL}}}, a comprehensive assessment system comprising 508 real-world instances, featuring a dual-metric mechanism that integrates both procedural and outcome perspectives.
\textbf{(C)} Experimental results reveal the gaps between diverse agents and real-world legal demands, offering valuable insights to guide future progress.
\textbf{(D)} This work introduces a new paradigm for legal intelligence, shifting from static to dynamic settings. Beyond evaluation, \textit{\textbf{J1-E}\scalebox{0.8}{\textbf{NVS}}} can be extended to data generation and reinforcement learning training.

\section{{\textit{\textbf{J1-E}\textbf{\small NVS}}}: Interactive Legal Environments}

In this section, we detail the construction pipeline of {\textit{\textbf{J1-E}\scalebox{0.8}{\textbf{NVS}}}}, including various agents organized in six scenarios of three levels, namely, legal consultation, document drafting, and judicial adjudication.
\textbf{(A) Role Agent Setting} (§\ref{Role Agent Setting}) constructs diverse agents by synthesizing real-world legal sources and personality theories.
\textbf{(B) Multi-level Environment Construction} (§\ref{Multi-level Environment Construction}) integrates these roles into coherent legal environments governed by specific procedures and relationships.

\subsection{Role Agent Setting}
\label{Role Agent Setting}
Legal environments inherently involve heterogeneous roles, distinguishing between legal professionals and layperson with different backgrounds, targets, and behaviors. 
Without specialized grounding, generic LLM-driven simulators are prone to significant legal illusions \cite{shengbinyue2025multi}. Therefore, creating realistic individuals presents two challenges: 
\textit{diversity of layperson} and \textit{legal consistency among roles}.
We address this by modeling individuals with the Big Five personality traits and assigning interconnected legal elements, derived from real-world legal practice, to different agents within the shared environment.

\textbf{Role Agents Configuration.}
We establish two categories of role repositories to populate the environment:
(1) \textit{Layperson} includes the general public with a broad background and specific characters with explicit litigation-related needs.
The former are assigned topic lists, background, and behavioral styles.
The latter, \textit{i.e.,} plaintiffs and defendants, are assigned behavioral styles and individual profiles, including personal attributes (\textit{e.g.}, names, genders, and addresses) and case-related details (\textit{e.g.}, claims, statements of defence, and evidence).
(2) \textit{Legal Professionals} include lawyers, prosecutors, and judges who assume formal legal responsibilities. They are assigned authentic legal information corresponding to their current environments.
For example, Plaintiff’s Lawyers in the civil court possess detailed case information extracted from civil judgment documents, \textit{e.g.}, their client information, case details, claims \& defence. In criminal court scenarios, the Procurator agent is assigned to charges and opinions sourced from criminal judgment documents. 
The methodology for acquiring these interconnected factual elements from real-world legal sources is detailed in Sec. \ref{sec:legal_sources}.

\textbf{Setup with Personality Theory.} 
Behavioral diversity enhances the model to represent complex characters, further aligning with real-world behavior. Guided by social personality theory \cite{sun2024identity}, Big-5 Personality Traits are mapped to behavioral styles of non-legal roles (public, the plaintiff, and the defendant). 
Each trait is assigned one of three levels (high, medium, or low) and GPT-4o is prompted to generate corresponding behavioral descriptions based on the role's profile or background. 
Details are provided in Appendix \ref{sec:appendix:Personality Theory Construction}.

\subsection{Multi-level Environment Construction}
\label{Multi-level Environment Construction}
Building on role repositories, we organize six environments into a three-tiered hierarchy spanning inquiry to adjudication. Below, we detail the participants and interaction logic, with pseudo-code provided in the Appendix \ref{sec:pseudo-code}.
\paragraph{Level-I: Legal Consultation.}
This level involves two participants, the general public and a legal agent, aiming to simulate the response capabilities of the legal trainee in dynamic environments.
(1)
\textbf{Knowledge Questioning }({\textbf{K}\scalebox{0.85}{\textbf{Q}}}). 
Through progressive dialogue, the legal agent resolves user confusion, moving from broad themes to specific issues. 
The public, assigned a topic list and occupation, continues the interaction until all topics are addressed. 
(2) \textbf{Consultation}({\textbf{L}\scalebox{0.85}{\textbf{C}}}). This presents a complex counseling setting grounded in a specific incident. Since legal agents lack information about the public, they need actively ask for relevant details to provide better responses. 
\paragraph{Level II: Document Drafting.}
This level involves a litigant with a concrete legal demand and a legal agent responsible for task execution, and is designed to simulate how a practicing lawyer independently gathers relevant information and completes a litigation task according to a given agenda.
(1) \textbf{Complaint Drafting} ({\textbf{C}\scalebox{0.85}{\textbf{D}}}). 
Guided by a predefined format, the legal agent gathers information through interaction with the plaintiff, autonomously plans the required content, and generates a complete complaint. 
(2) \textbf{Defence Drafting} ({\textbf{D}\scalebox{0.85}{\textbf{D}}}). 
Unlike above, this scenario starts from an existing complaint. The legal agent interacts with the defendant, formulates counterarguments based on the complaint, and generates the corresponding defence document. 
\paragraph{Level III: Judicial Adjudication.}
This level involves multiple participants and is governed by strict court procedural norms.
(1) \textbf{Civil Court} ({\textbf{C}\scalebox{0.85}{\textbf{I}}}).
The Civil Court addresses disputes concerning rights and obligations between individuals or organizations, involving three participants: the plaintiff, the defendant, and the judge. 
It includes five stages: court preparation, court investigation, court debate, court mitigation, and court decision. 
The judge agent oversees the process by following the procedural framework and interacting with both parties' attorneys. In the final stage, the agent renders a judgment in response to the plaintiff’s claims.
(2) \textbf{Criminal Court} ({\textbf{C}\scalebox{0.85}{\textbf{R}}}).
The Criminal Court centers on adjudicating criminal charges against individuals or groups, involving four participants: the defendant, the defendant's lawyer, the procurator, and the judge. It includes three stages, court preparation, court investigation, and decision. Compared to civil court,
legal agents in criminal proceedings interact with a broader range of participants, eliciting statements from the defendant and the lawyer, requesting evidence from the procurator, and working toward the establishment of criminal liability.

\begin{table*}[t]
\centering
\small
\setlength{\tabcolsep}{3.6pt}
\begin{tabular}{lllp{4.8cm}lll}
\toprule
\textbf{Scenario} & \textbf{ID} & \textbf{Metric} & \textbf{Definition} & \textbf{Range} & \textbf{Type} & \textbf{Method} \\
\midrule
\multirow{2}{*}{\centering Level-I}
& I-1 & Binary Accuracy
& Binary (Yes/No) answer accuracy
& [0,0.5,1] & {O\scalebox{0.85}{O}} & EM \\

& \cellcolor[HTML]{ECF4FF}I-2
& \cellcolor[HTML]{ECF4FF}Non-binary Score
& \cellcolor[HTML]{ECF4FF}Score for open-ended answers
& \cellcolor[HTML]{ECF4FF} [0-1]
& \cellcolor[HTML]{ECF4FF}{O\scalebox{0.85}{O}}
& \cellcolor[HTML]{ECF4FF}LLM \\
\midrule
\multirow{2}{*}{\centering Level-II}
& II-1 & Format-following Score
& Document structure alignment
& [0-1] & {P\scalebox{0.85}{O}} & EM \\

& \cellcolor[HTML]{ECF4FF}II-2
& \cellcolor[HTML]{ECF4FF}Document Score
& \cellcolor[HTML]{ECF4FF}Average component-level accuracy
& \cellcolor[HTML]{ECF4FF}[0,1] or [0-1]
& \cellcolor[HTML]{ECF4FF}{O\scalebox{0.85}{O}}
& \cellcolor[HTML]{ECF4FF}EM+LLM \\
\midrule
\multirow{6}{*}{\centering Level-III}
& \cellcolor[HTML]{ECF4FF}III-1
& \cellcolor[HTML]{ECF4FF}Procedural-following Score
& \cellcolor[HTML]{ECF4FF}Procedural stage completeness
& \cellcolor[HTML]{ECF4FF}[0-1]
& \cellcolor[HTML]{ECF4FF}{P\scalebox{0.85}{O}}
& \cellcolor[HTML]{ECF4FF}LLM \\

& \cellcolor[HTML]{ECF4FF}III-2
& \cellcolor[HTML]{ECF4FF}Judgment Score
& \cellcolor[HTML]{ECF4FF}Final judgment quality
& \cellcolor[HTML]{ECF4FF}[0-1]
& \cellcolor[HTML]{ECF4FF}{O\scalebox{0.85}{O}}
& \cellcolor[HTML]{ECF4FF}LLM \\

& III-3 & Crime Accuracy
& Charge prediction accuracy
& [0-1] & {O\scalebox{0.85}{O}} & EM \\

& III-4 & Penalty Deviation
& Log-distance penalty error
& [0-1] ($\downarrow$)& {O\scalebox{0.85}{O}} & Dist. \\

& \cellcolor[HTML]{ECF4FF}III-5
& \cellcolor[HTML]{ECF4FF}Reason Score
& \cellcolor[HTML]{ECF4FF}Legal reasoning quality
& \cellcolor[HTML]{ECF4FF}[0-1]
& \cellcolor[HTML]{ECF4FF}{P\scalebox{0.85}{O}}
& \cellcolor[HTML]{ECF4FF}LLM \\

& III-6 & Law Accuracy
& Cited law precision
& $[0,1]$ & {P\scalebox{0.85}{O}} & EM \\
\bottomrule
\end{tabular}
\caption{Metrics in \textit{\textbf{J1-ENVS}}. {O\scalebox{0.85}{O}}, {P\scalebox{0.85}{O}}, EM, and Dist. denote outcome-oriented metrics, process-oriented metrics, exact match, and nlog-distance, respectively. Except for III-1, all metrics are evaluated against explicit ground truth.}
\label{tab:metrics}
\end{table*}

\section{\textit{J1-Eval}: Dual-metric Assessment System } 
\subsection{Setup with Real-world Legal Sources}
\label{sec:legal_sources}
To ensure authenticity, we assign distinct legal elements from the same event to various roles, drawing from two primary sources:
(1) \textit{Chinese Judgment Documents}. 
For Level II and Level III scenarios, we rely on Chinese court judgments, as they naturally encapsulate the full procedural lifecycle of legal cases, including litigant information, disputes, factual findings, judicial reasoning, rulings, and applicable statutes. We collect both civil and criminal judgments and employ LLMs to extract structured components such as party identities, claims, detailed case facts, and cited legal provisions.
(2) \textit{Chinese Legal Articles}. 
For Level I scenarios, which prioritize progressive inquiry and logical authenticity, we curate expert-authored legal articles from the \textit{HUALV} website \footnote{https://www.66law.cn} and the book  \textit{Civil Trial Practice Q\&A}. 
Given that these articles typically feature step-by-step questions or thematic headings, we prompt LLMs to utilize this inherent structure to generate coherent topic lists (classified as binary or open-ended) alongside their corresponding answers. Further data construction details are provided in Appendix \ref{sec:appendix:Legal Sources Processing}.

\vspace{-1.5mm}

\begin{figure}[t]
    \centering
\includegraphics[width=0.9\linewidth]{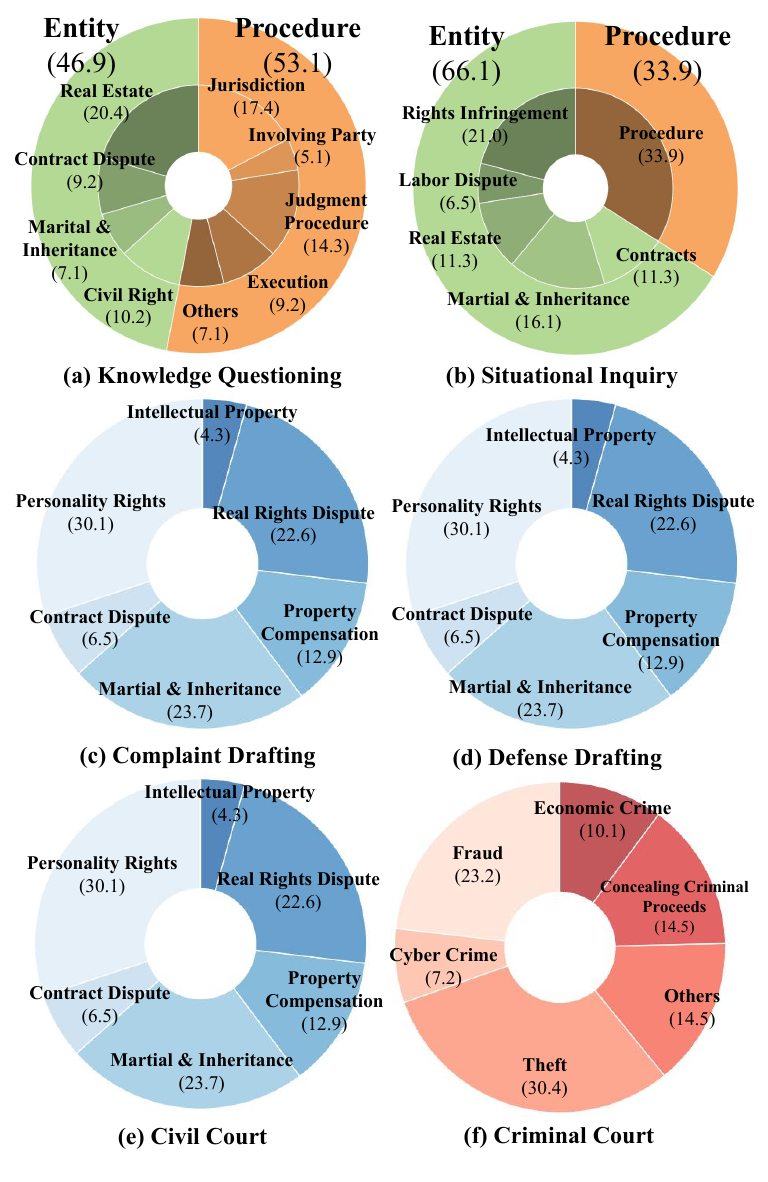}
    \caption{Distribution of legal attributes for six environments in {\textit{\textbf{J1-E}\scalebox{0.84}{\textbf{VAL}}}}, showing a wide range of coverage.}
    \label{fig:dist}
\end{figure}

\paragraph{Data Details.}
All documents are collected from 2025 onward to ensure legal timeliness. After preprocessing, {\textit{\textbf{J1-E}\scalebox{0.8}{\textbf{VAL}}}} consists of 508 instances, including 160 Level I, 186 Level II, and 192 Level III instances. Dataset statistics are reported in Appendix Table~\ref{tab:env-data}.
Our benchmark spans a diverse and nuanced set of legal attributes aligned with real-world legal practice, derived directly from raw case data.
As shown in Figure \ref{fig:dist}, 
{K\scalebox{0.85}{Q}} and {L\scalebox{0.85}{C}} involve different entities and procedural topics.{C\scalebox{0.85}{D}}, {D\scalebox{0.85}{D}} and {C\scalebox{0.85}{I}} address various types of civil rights disputes. {C\scalebox{0.85}{R}} covers different categories of criminal offenses. Detailed data analysis are provided in Appendix \ref{sec:appendix: Data statistics}.

\subsection{Evaluation Methods}
\label{sec: Evaluation Methods}
To comprehensively evaluate, we propose a dual evaluation framework that integrates \textbf{outcome-oriented} and \textbf{process-oriented} metrics. 
The former quantifies the quality of the final objectives (e.g., judgment accuracy), whereas the latter assesses the quality of essential intermediate constraints throughout task execution.
As shown in Table \ref{tab:metrics}, these metrics are strategically combined based on scenario characteristics to facilitate assessment across different environmental levels. 
Each metric adopts a score range in \textbf{[lower bound–upper bound]} form, utilizing either rule-based or LLM-based automatic methods. Crucially, each instance is equipped with an \textbf{explicit ground truth} for the final objective. This approach rigorously mitigates subjective judgment and potential bias, aligning with established methodologies for long-form generation tasks \cite{li2025generation}. See Appendix \ref{sec:appendix:Evaluation Metric} for details on evaluation metrics.

\paragraph{Outcome-oriented metrics.} 
These metrics evaluate the fidelity of the agent's final deliverables against the ground truth across three-level scenarios.
(1) \textit{Legal consultation}. Binary accuracy (\metricid{I-1}) evaluates yes/no responses via exact match, while Non-binary score (\metricid{I-2}) assesses open-ended answers based on factual accuracy and legal coverage.
(2) \textit{Document drafting}. Document score (\metricid{II-1}) measures content correctness, with personal information verified by exact match and substantive components (e.g., claims).
(3) \textit{Judicial adjudication}. Civil judgment score (\metricid{III-1}) and crime accuracy (\metricid{III-3}) evaluate civil rulings and criminal charges, respectively. Additionally, Penalty deviation (\metricid{III-4}) measures the accuracy of fines and sentences via log distance \cite{chen2019charge}, where lower values indicate higher accuracy.

\paragraph{Process-oriented metrics.}
These metrics assess compliance with procedural norms, formatting constraints, and logical reasoning pathways.
Format-following Score (\metricid{II-1}) evaluates document structure by verifying the presence, order, and labeling of mandatory components.
Procedural-following Score (\metricid{III-1}) measures trial completeness, where a stage is considered complete only if all predefined actions are correctly executed.
Reason Score (\metricid{III-5}) evaluates logical coherence of judicial reasoning. 
Law Accuracy (\metricid{III-6}) evaluates correctness of cited statutes via ground-truth matching.
Except for \metricid{III-1} metric, all process-oriented metrics are evaluated against explicit ground-truth references.

\begin{table*}[t]
\centering
\resizebox{\textwidth}{!}{
\begin{tabular}{lcc|cc|cc|cc|cccc|ccccc|cc}
\hline
\multicolumn{1}{l|}{\multirow{3}{*}{\textbf{Model}}} & \multicolumn{4}{c|}{\textbf{Level-I}}                                     & \multicolumn{4}{c|}{\textbf{Level-II}}                                   & \multicolumn{9}{c|}{\textbf{Level-III}}    & \multicolumn{2}{c}{\multirow{2}{*}{\textbf{AVG}}}                                             \\ \cline{2-18} 
\multicolumn{1}{l|}{}                       & \multicolumn{2}{c|}{\textbf{KQ}}         & \multicolumn{2}{c|}{\textbf{LC}}         & \multicolumn{2}{c|}{\textbf{CD}}        & \multicolumn{2}{c|}{\textbf{DD}}        & \multicolumn{4}{c|}{\textbf{CI}}                     & \multicolumn{5}{c|}{\textbf{CR}} & \multicolumn{1}{l}{} & \multicolumn{1}{l}{}          \\ \cline{2-20} 
\multicolumn{1}{l|}{}                       & I-1 & \multicolumn{1}{c|}{I-2} & I-1 & \multicolumn{1}{c|}{I-2} & II-1 & \multicolumn{1}{c|}{II-2} & II-1 & \multicolumn{1}{c|}{II-2} & III-1 & III-2 & III-5 & \multicolumn{1}{c|}{III-6} & III-1 & III-3 & \textcolor{red}{III-4} & III-5 & III-6 & OO & PO \\ \hline
\rowcolor{gray!25}\multicolumn{20}{c}{\textit{Multilingual LLMs (close source)}} \\
\multicolumn{1}{l|}{ GPT-4o}                  & \underline{69.7}                                     & 62.7                                      & 55.9                                     & 45.1                                      & 58.9                                     & \textbf{89.9}                                     & 45.4                                     & \textbf{84.8}                                      & \textbf{87.0}                                     & \textbf{28.5}                                      & \textbf{50.2}                                     & 13.9                                      & 49.3                                     & 60.9                    & 38.5  & 42.8                                     & 27.0  & \underline{62.2}  &  \underline{46.8}                        \\
\multicolumn{1}{l|}{ Claude-3.7}             & 68.4                                     & 60.7                                      & 52.4                                     & \underline{49.6}                                      & 68.8                                     & \underline{89.2}                                     & 38.7                                     & 68.8                                      & \underline{65.8}                                     & \underline{21.2}                                     & \underline{40.2}                                     & \textbf{17.5}                                      & \underline{56.9}                                     & \textbf{88.4}                    & \textbf{15.7} & \textbf{72.8}                                     & \textbf{41.1}       & \textbf{62.3} & \textbf{50.2}                             \\ \hline
\rowcolor{gray!25}\multicolumn{20}{c}{\textit{Multilingual LLMs (open source)}} \\
\multicolumn{1}{l|}{ Deepseek-v3\space$_{\mathrm{671B}}$}        & 67.1                                     & 60.7                                      & 54.5        & \textbf{49.7} & 84.7 & 63.6 &\textbf{57.2} & \underline{76.3} & 24.0 & 12.4 & 19.5 & 7.6 & 42.4 & 76.8 & 33.9 & \underline{57.2} & \underline{32.3} & 57.6 & 40.6        \\
\multicolumn{1}{l|}{ Deepseek-r1\space$_{\mathrm{671B}}$}        & \textbf{72.5}  &  \underline{62.8} & \underline{57.2} & 48.3 & 69.9 & 46.4 & 36.3 & 47.9 & - & 0.8 & 1.1 & 0.7 & 0.8 & 1.4 & 98.6 & 1.4 & 0.7 & 42.2   & 13.9                                 \\
\multicolumn{1}{l|}{ Llama3.3-it.\space$_{\mathrm{70B}}$}        & 64.8 & 50.1 & 43.9 & 23.6 & \textbf{97.9} & 84.5 & 32.3 & 62.4 & 61.0 & 12.8 & 25.3 & 4.5 & 50.7 & \underline{81.2} & 25.0 & 50.1 & 17.2 & 52.9 & 42.4 \\
\multicolumn{1}{l|}{ Qwen3\space$_{\mathrm{32B}}$}          & 69.6 & \textbf{66.2} & \textbf{58.3} & 47.2 & 91.2 & 84.4 & 46.7 & 63.5 & 52.7 & 17.0 & 34.1 & \underline{15.4} & 26.8 & 49.3 & 58.3 & 40.0 & 22.9   & 56.9    & 41.2                              \\
\multicolumn{1}{l|}{ Gemma3-it\space$_{\mathrm{27B}}$}          & 64.1 & 49.5 & 54.1 & 26.0 & \underline{96.3} & 84.8 & 37.9 & 68.8 & 61.7 & 15.5 & 32.4 & 6.4 & \textbf{60.0} & 49.3 & \underline{22.1}& 36.1 & 15.6 & 51.5 & 43.3 \\
\multicolumn{1}{l|}{ Qwen3\space$_{\mathrm{14B}}$}          & 67.7   & \underline{62.8} & 56.2 & 44.3 & 63.9 & \underline{89.2}  & 36.6  &62.4  &13.0  &4.3  &12.2  &9.8  &16.9  &26.1  & 73.6  & 22.8  &17.1 & 51.6      & 24.0    \\
\multicolumn{1}{l|}{ Gemma3-it\space$_{\mathrm{12B}}$}          & 62.0  & 42.0  & 47.9  & 19.2  & 51.8  & 84.1  & 35.3  & 63.9  & 39.1  & 12.0  & 23.7  & 3.5  & 41.3  & 58.0  & 40.8  & 38.3  & 8.1 & 48.6 & 30.1\\
\multicolumn{1}{l|}{ GLM4\space$_{\mathrm{9B}}$}                & 66.0  & 54.9  & 54.0  & 32.4  & 56.6  & 82.2  & 40.4  & 52.2  & 12.7  & 2.3  & 8.0  & 1.5  & 25.1  & 39.1  & 75.0  & 24.8  & 10.1 & 47.9 & 22.4                                \\
\multicolumn{1}{l|}{ Qwen3\space$_{\mathrm{8B}}$}           & 66.2 & 57.1 & 55.0 & 36.9 & 74.2 & 72.7 & 17.6 & 22.8 & 9.9 & 5.8 & 14.2 & 7.2 & 16.3 & 26.1 & 67.4 & 19.4 & 16.7  & 42.8       & 21.9                        \\
\multicolumn{1}{l|}{ InternLM3-it.\space$_{\mathrm{8B}}$}       & 65.3 & 55.3 & 55.5 & 42.4 & 91.0 & 72.2 & 36.1 & 53.8 & 31.8 & 7.7 & 19.0 & 4.2 & 29.1 & 60.9 & 45.1 & 46.7 & 28.7  & 51.6         & 35.8                      \\
\multicolumn{1}{l|}{ Ministral-it.\space$_{\mathrm{8B}}$}    & 59.6 & 37.6 & 51.5 & 14.5 & 60.0 & 2.8 & 34.2 & 14.0 & 22.7 & 5.7 & 12.4 & 0.5 & 2.1 & 4.3 & 98.0 & 2.3 & 0.9   & 23.8                        & 16.9          \\
\multicolumn{1}{l|}{ Qwen2.5-it.\space$_{\mathrm{7B}}$}         & 67.2 & 54.9 & 52.6 & 31.3 & 89.5 & 81.4 & \underline{47.1} & 59.7 & 15.6 & 7.3 & 14.1 & 7.5 & 26.4 & 49.3 & 54.5 & 37.5 & 25.4    & 50.5       & 32.9                       \\
\multicolumn{1}{l|}{ Qwen3\space$_{\mathrm{4B}}$}           & 62.2 & 55.9 & 50.7 & 37.2 & 51.4 & 83.7 & 33.1 & 51.5 & 28.0 & 11.3 & 18.3 & 7.6 & 20.8 & 44.9 & 57.3 & 32.9 & 18.8    & 49.7             & 26.4                  \\ \hline
\rowcolor{gray!25}\multicolumn{20}{c}{\textit{Legal-specific LLMs}} \\
\multicolumn{1}{l|}{ LawLLM\space$_{\mathrm{13B}}$}               & 56.3 & 47.3 & 44.9 & 20.8 & 12.9 & 2.9 & 0.6 & - & 1.0 & - & 0.3 & - & 12.7 & 17.4 & 87.4 & 10.9 & 2.5   & 23.7        & 5.1                          \\
\multicolumn{1}{l|}{ Chatlaw2\space$_{\mathrm{7B}}$}             & 60.6 & 53.1 & 50.4 & 25.0 & 22.4 & 17.0 & 1.3 & 0.9 & - & - & - & - & - & - & - & - & - & 25.9 & 3.0
\\\hline
\end{tabular}
}
\caption{Performance among LLM-driven legal agents on {\textit{\textbf{J1-E}\scalebox{0.8}{\textbf{NVS}}}}, where the bolded score indicates the best performance, while the underlined score represents the second-best. Note that “-” denotes failure to accomplish the task, and lower \textbf{III-4} indicates higher performance. }
\label{tab:level}
\end{table*}

\section{Experiments}
\subsection{Experimental Settings}
\paragraph{Agent Models.}
We evaluate legal agents driven by various LLMs in two categories. (1) \textit{General multilingual LLMs}: GPT-4o\footnote{gpt-4o-2024-11-20} \cite{achiam2023gpt}, Claude-3.7\footnote{claude-3-7-sonnet-20250219} \cite{TheC3}, 
Deepseek-v3~\cite{liu2024deepseek}, Deepseek-r1~\cite{guo2025deepseek},
InternLM3-instruct 8B~\cite{cai2024internlm2},
Qwen2.5-instruct 7B~\cite{yang2024qwen2}, Qwen3 4B/8B/14B/32B~\cite{yang2025qwen3}, ChatGLM4-chat 9B~\cite{glm2024chatglm}, and Ministral-Instruct-2410 8B \cite{jiang2023mistral7b}. (2) \textit{Legal-specific LLMs}: LawLLM 13B \cite{yue2024lawllm}, and Chatlaw2 7B \cite{cui2023chatlaw}.

\paragraph{Implementation Detail.} 
We utilize arbitrary LLMs, GPT-4o~\cite{achiam2023gpt} or Qwen3-32B~\cite{yang2025qwen3}, to drive the environments.
Due to page limitations, details of evaluation settings are provided in the Appendix \ref{sec:appendix: setting_details}.

\subsection{Performance among Various Legal Agents}
In this setting, we leverage GPT-4o to drive our interactive environments.

\paragraph{Performance at Level I \& II.}
Table \ref{tab:level} reports our metrics across levels. 
(1) At \textit{Level I}, both general-purpose and legal-specific models achieve strong performance on {K\scalebox{0.8}{Q}}, reflecting solid legal knowledge. Performance declines on the more interactive {L\scalebox{0.8}{C}} task, which demands proactive engagement.  
(2) At \textit{Level II},
LLaMA-3.3 and Deepseek-v3 achieve the highest scores on \metricid{II-1} for {C\scalebox{0.8}{D}} and {D\scalebox{0.8}{D}} tasks, demonstrating strong template adherence in the long legal context.
Notably, performance in {D\scalebox{0.8}{D}} scenarios is lower than in {C\scalebox{0.8}{D}}, as {D\scalebox{0.8}{D}} settings require models to construct more complex and coherent counterarguments in response to an opposing party's complaint.
Detailed results are reported in Tables \ref{tab:DOC in Complaint Drafting} and \ref{tab:Defence} of the Appendix.

\paragraph{Performance at Level III.}
Civil and criminal courts involve distinct procedures and judgment forms. 
The \textbf{civil court} comprises five stages, yet most models struggle to complete the full process, particularly in legal-specific models, reasoning models, and smaller models. 
GPT-4o and Claude-3.7 perform well in
 \metricid{III-1}, \metricid{III-2} and \metricid{III-5}. 
The criminal court includes three stages and more participants, where 
Claude-3.7 achieves leading performance.
Additionally, Deepseek-r1’s poor performance in executing the court process highlights the limitations of reasoning models in complex environments. 
Overall, \textit{effective reasoning and adherence to procedural protocols within extended legal contexts are essential for achieving accurate legal judgments}.
Detailed judgment component results are reported in Tables \ref{tab:Civil Court} and \ref{tab: Criminal Court}.
\begin{figure}[t]
    \centering
\includegraphics[width=\linewidth]{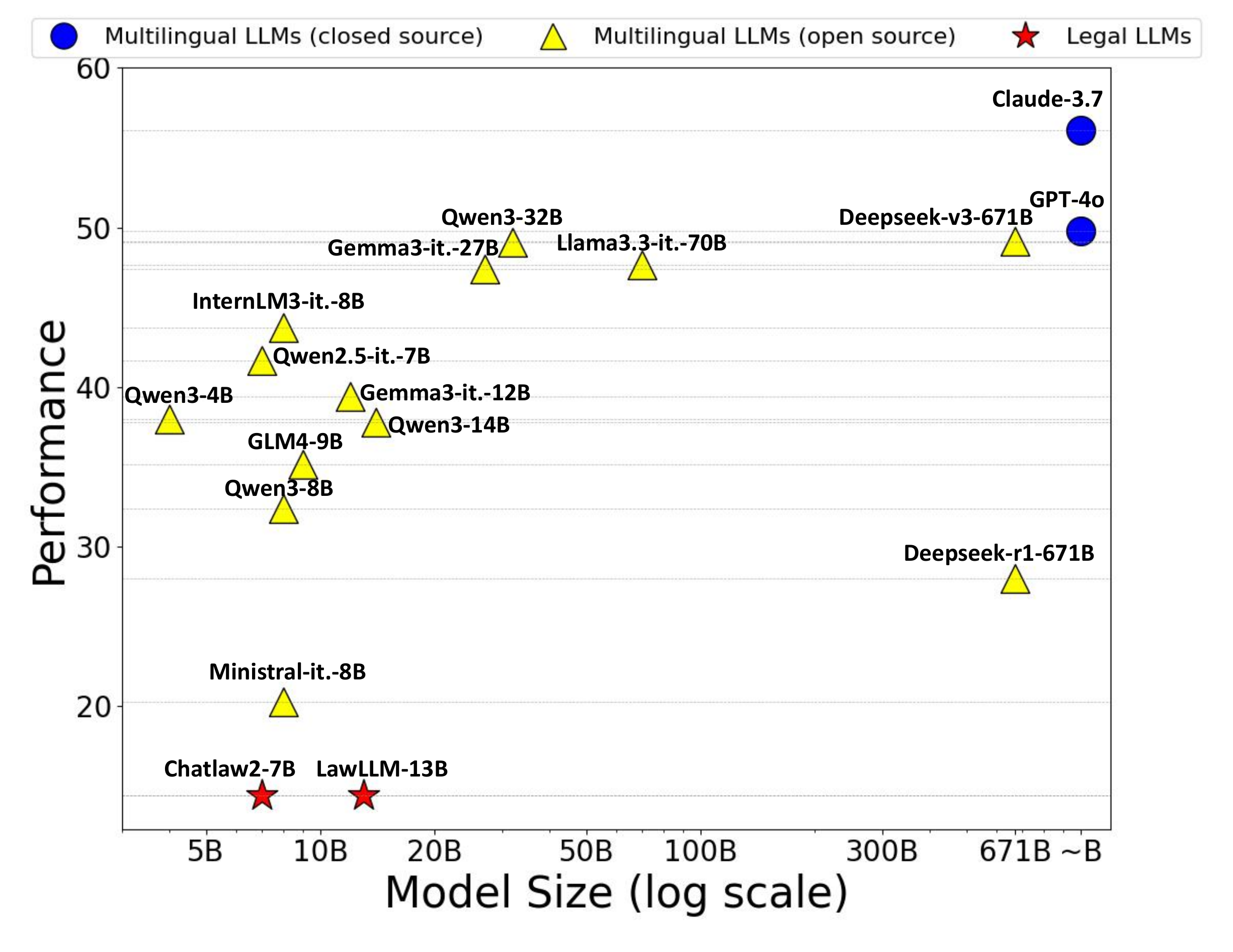}
    \caption{Overall performance ranking across different LLM agent sizes.}
    \label{fig:total_performance}
\end{figure}

\paragraph{Overall Performance.}
We compute the average scores across all environments except \metricid{III-4}, and rank the models by size. 
As shown in Figure \ref{fig:total_performance}, Claude-3.7 achieves the best performance, showing strong legal intelligence. 
Notably, although the legal-specific LLMs perform comparably to GPT series on existing legal benchmarks~\cite{fei2024lawbench}, they exhibit significantly weaker performance in our setting, falling behind even smaller models. This highlights a key limitation: \textbf{despite possessing legal knowledge, the absence of interactive capability hampers their effectiveness in dynamic, realistic environments}.

\begin{figure}[t]
    \centering
    \includegraphics[width=\linewidth]{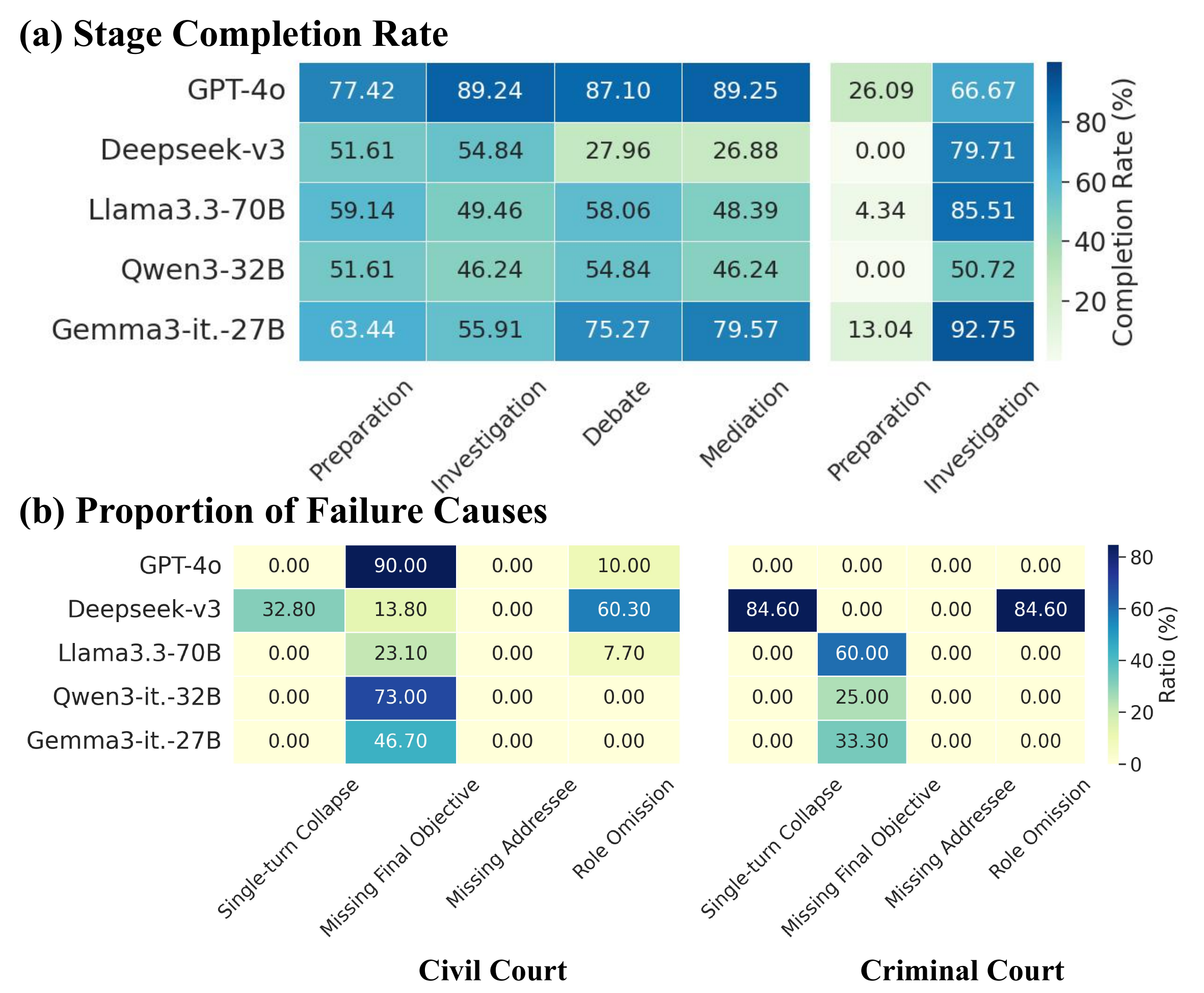}
    \caption{Procedural-following performance of legal agents in civil and criminal courts. (a) Completion rate of court stages in Civil \& Criminal Court. (b) Proportion of failed cases attributed to each cause.}
    \label{fig:completion rate}
\end{figure}

\subsection{Further Analysis}
\paragraph{Analysis from process- and outcome-oriented perspectives.}
Table \ref{tab:level}  presents the average performance of Outcome-oriented ({O\scalebox{0.85}{O}}) and Process-oriented ({P\scalebox{0.85}{O}}) metrics. Models with higher {P\scalebox{0.85}{O}} consistently achieve superior {O\scalebox{0.85}{O}}, and the {O\scalebox{0.85}{O}} scores significantly surpass the {P\scalebox{0.85}{O}} across all models. Based on these, we draw two conclusions:
(1) Processal adherence is not a mere formality but the bedrock of logical validity and factual completeness, making it a necessary condition for precise legal judgment.
(2) In the legal domain, processal justice is often prioritized over the outcome itself. The lag in  {P\scalebox{0.85}{O}} metrics reveals a \textbf{severe deficiency in current models' capabilities to understand, adhere to, and execute legal procedures}.

\paragraph{Analysis of court completeness.}
As shown in Fig. \ref{fig:completion rate}, the success rates of the agent in completing all required steps vary across different stages of civil (CI) and criminal (CR) courts. 
The results indicate that 
(1) Current legal agents encounter difficulties in the preliminary stages of court proceedings. While these stages are largely formalistic, they are indispensable for ensuring procedural justice in real-world practice.
(2) Navigating the criminal court poses greater challenges for legal agents, as the environment contains more roles that increase the complexity of the interaction.
In addition, we identify four causes of failure: single-turn collapse, missing final objective, missing addressee, and role omission. Definitions of causes are provided in Appendix \ref{sec:appendix:failed causes}. Figure~\ref{fig:completion rate} shows that missing the final objective accounts for most failures.
These findings underscore the need for enhanced \textbf{procedural learning for legal agents}, and the importance of \textbf{strengthening their communication and coordination with different roles}.

\begin{table}[t]
\centering
\small
\resizebox{0.92\linewidth}{!}{
\begin{tabular}{ccc}
\hline
\multicolumn{1}{c|}{Agent}            & Env.GPT-4o & Env.Qwen3\space$_{\mathrm{32B}}$ \\ \hline
\rowcolor{gray!25} \multicolumn{3}{c}{Evaluator(GPT-4o)}                                          \\
\multicolumn{1}{c|}{ GPT-4o}           & 94.22          & 95.49                 \\
\multicolumn{1}{c|}{ Deepseek-v3\space$_{\mathrm{671B}}$} & 90.97          & 94.71                 \\
\multicolumn{1}{c|}{ Qwen3\space$_{\mathrm{32B}}$}    & 89.97          & 94.99                 \\ \hline
\rowcolor{gray!25} \multicolumn{3}{c}{Evaluator(Human)}                                           \\
\multicolumn{1}{c|}{ GPT-4o} &  76.25               &   75.63                   \\
\multicolumn{1}{c|}{ Deepseek-v3\space$_{\mathrm{671B}}$} &  74.13              &   75.80                    \\
\multicolumn{1}{c|}{ Qwen3\space$_{\mathrm{32B}}$}    &    74.00           &     74.25                  \\ \hline
\end{tabular}}%

\caption{Behavior consistency of environment roles across different legal agents (GPT-4o, Deepseek-v3, Qwen 32B) under Human and LLM evaluation.}
\label{tab:human}
\end{table}

\subsection{Can LLMs drive \textit{\textbf{J1-E}\textbf{\small NVS}} and \textit{\textbf{J1-E}\textbf{\small VAL}} ?}
This section aims to validate the \textbf{stability} and \textbf{reliability} of our environment (P1, P2) and evaluation method (P3). we leverage GPT-4o and Qwen3-32B to drive the \textit{\textbf{J1-E}\scalebox{0.8}{\textbf{NVS}}} separately.
All human evaluation settings are detailed in the Appendix \ref{sec:appendix:human evaluation}.

\paragraph{P1: behavioral consistency analysis of environment roles.}
Table \ref{tab:human} presents average consistency scores (1–10) evaluated by GPT-4o (all samples) and humans (20 per scenario).
Across different evaluators, environment roles maintain high and stable scores, regardless of the underlying LLM. 
Furthermore, scores maintain stability across diverse environments for the same legal agent.
The result validates the reliability and effectiveness of our environment, laying a solid foundation for assessing agents in interactive legal scenarios. Detailed results are provided in Tables \ref{tab:Detailed GPT evaluation} and \ref{tab:Detailed Human Evaluation}.

\begin{figure}[t]
    \centering
    \includegraphics[width=\linewidth]{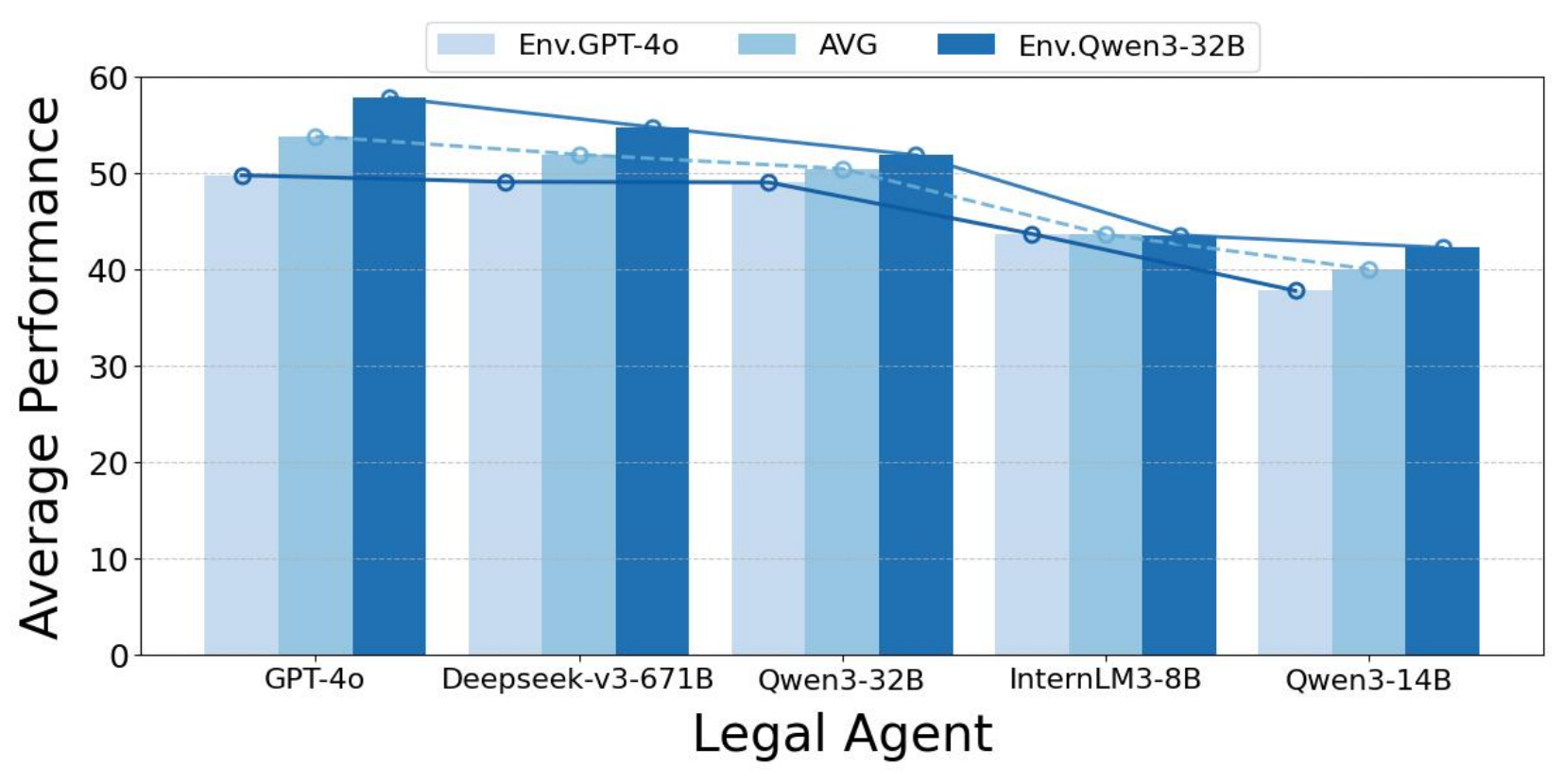}
    \caption{Overall performance of legal agents with {\textit{\textbf{J1-E}\scalebox{0.8}{\textbf{NVS}}}} driven by GPT-4o or Qwen3-Instruct-32B.}
    \label{fig:total_per_dJ1}
\end{figure}

\paragraph{P2: analysis of agent performance across different environments.}
Figure \ref{fig:total_per_dJ1} shows that all five agents successfully complete their tasks in the Qwen-based environment, with performance rankings consistent with those observed in the GPT-based environment. 
Although agent performances under the Qwen3-32B setting differ from those in GPT-based environments due to differences in model scale and capability, such variation is expected and acceptable. Importantly, achieving a balance between dynamics and controllability is more critical than strict performance equivalence across models.
In addition, Appendix Tables \ref{tab:confidence interval kq lc cd dd} and \ref{tab:confidence interval ci cr} demonstrate the stability of agent performance across multiple repeats.
Results validate the robustness of our framework and support its applicability to diverse deployment scenarios.

\begin{figure}[t]
    \centering
    \includegraphics[width=0.85\linewidth]{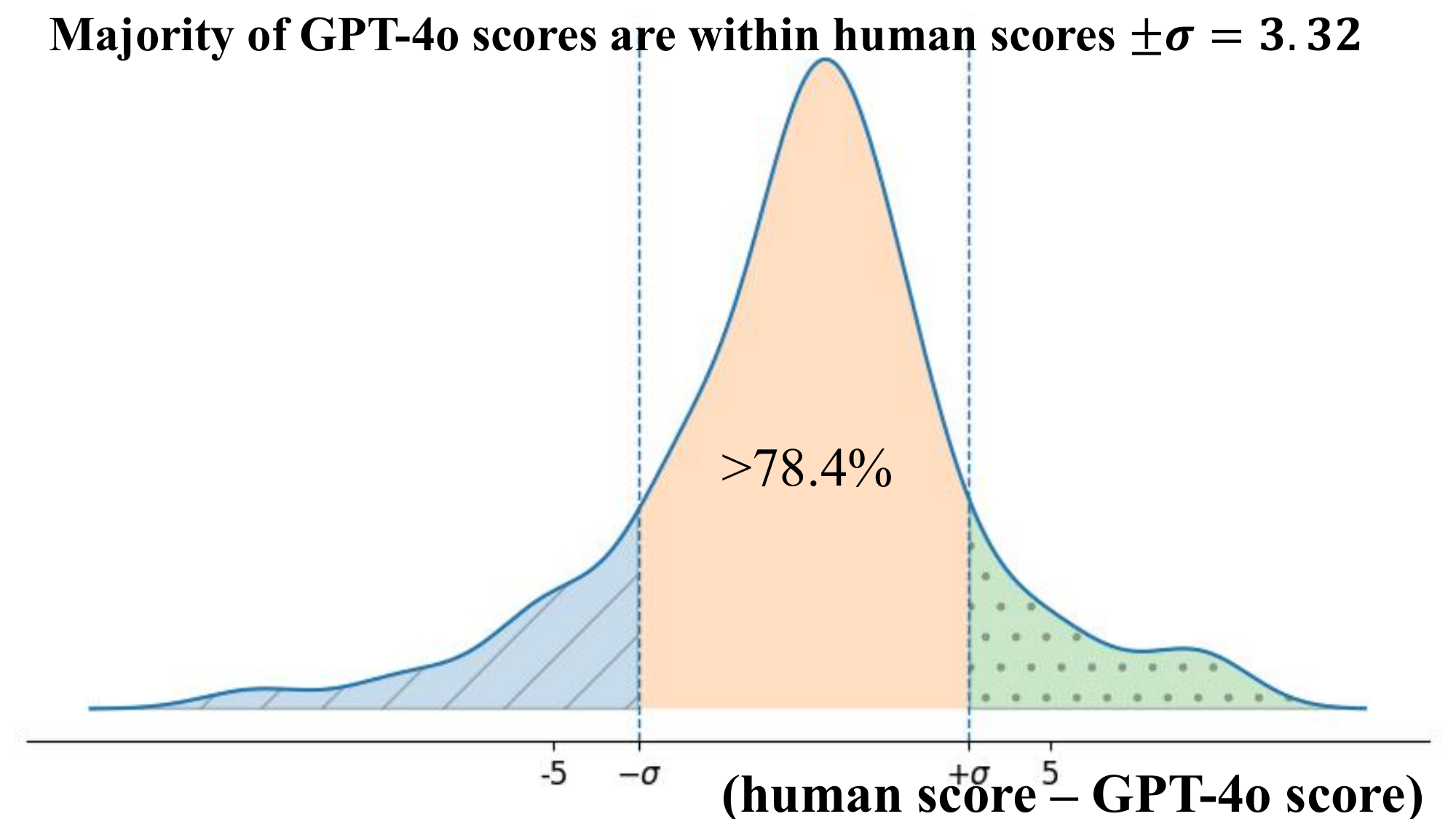}
    \caption{The distribution of the difference between GPT-4o scores and human evaluation}
    \label{fig:evaluation_evaluation}
\end{figure}

\paragraph{P3: analysis of the LLM evaluator in \textit{\textbf{J1-E}\scalebox{0.8}{\textbf{VAL}}}.}
We assess the reliability of LLM-based metrics by measuring agreement between human judgments and model outputs. Following the protocol in Appendix \ref{sec:appendix:human evaluation}, human annotators are instructed to score 20 cases per scenario.
The pairwise difference between human and model judgments is averaged with extremes removed.
As shown in Fig. \ref{fig:evaluation_evaluation}: 
(1) The average difference is 0.03, indicating \textbf{a high level of overall agreement between LLM-based and human assessments}.
(2) 78.4\% of the cases exhibit a score difference within one standard deviation($\leq 1 * \sigma = 3.32$), suggesting that \textbf{evaluation differences are statistically controlled}. Causes of the divergence are analyzed in Appendix \ref{sec:Analysis of differences between human evaluation and LLM-as-Judge}. 
Overall, these results support the reliability of LLM-based evaluation.

\section{Related Work}

\paragraph{Legal Benchmark.}
Existing legal evaluations of LLM are based on static, single-turn settings \cite{fei2024lawbench, li2024lexeval, guha2023legalbench, dai2023laiw, yue2024lawllm}. For example, LegalBench \cite{guha2023legalbench} and LawBench \cite{fei2024lawbench} assess legal capabilities by adapting existing NLP legal tasks to LLMs, while Law-Eval \cite{yue2024lawllm} evaluates models using multiple-choice questions from law exams. LexEval \cite{li2024lexeval}, inspired by Bloom’s taxonomy, organizes legal abilities into a cognitive hierarchy, and LAiW \cite{dai2023laiw} structures legal reasoning into three levels based on syllogistic logic.
However, these static, task-centric benchmarks fail to capture the dynamic complexity of real scenarios, rendering them inadequate for agent evaluation.
\paragraph{LLM-based Simulation.}
By emulating human perception and behavior, LLMs can be used to construct role agents with diverse personalities and environmental contexts, ranging from individuals \cite{shao2023character,argyle2023out} to demographic groups \cite{zhang2025socioverse,jiang2023evaluating}. These agents are capable of coordinating, collaborating, exchanging information, and even engaging in debates, thereby exhibiting social intelligence \cite{zhou2023sotopia,mou-etal-2025-agentsense}.
Building on such role agents, recent work has explored more complex simulated environments \cite{stade2024large,fan2025ai}. For example, Generative Agents \cite{park2023generative} model a small town with 25 agents, while AI Hospital \cite{fan2025ai} simulates dynamic medical interactions among doctors and NPCs.
Overall, the strong role-representation ability of LLMs provides a foundation for constructing complex community-level simulations.

\section{Conclusion}
We present the first dynamic benchmark for assessing legal intelligence, powered by \textit{\textbf{J1-E}\scalebox{0.8}{\textbf{NVS}}}, an environment system designed to simulate agent interactions within authentic legal scenarios, featuring 508 specific environments across three hierarchical levels and six distinct scenarios; and \textit{\textbf{J1-E}\scalebox{0.8}{\textbf{VAL}}}, which evaluates complex task performance based on both outcome rationality and procedural justice. Through extensive benchmarking of general-purpose and legal-oriented LLMs, we reveal the deficiencies of existing models in realistic, dynamic legal contexts. This work underscores the necessity for agents with robust interactive capabilities and procedural awareness, while positioning \textit{\textbf{J1-E}\scalebox{0.8}{\textbf{NVS}}} as a foundational infrastructure for subsequent data construction and agent training.


\section*{Limitations}
Our benchmark focuses primarily on the procedural flow of legal tasks, while in real-world applications, agents may require more complex capabilities to complete tasks effectively, such as statutory retrieval or precedent reasoning. The prerequisite for evaluating such capabilities is the establishment of a dynamic interactive environment. Therefore, we expect our pioneering effort to serve as a necessary starting point for future research on legal agents' integration of these advanced capabilities in dynamic environments.
In addition, we concentrate on developing dynamic environments within the Chinese legal system, presenting a comprehensive methodology for reconstructing interactive scenarios based on real legal data. The approach is readily adaptable to other legal systems for future work.


\section*{Ethics Statement}
All data used in our study were obtained from publicly accessible sources  with sensitive information properly anonymized. {\textit{\textbf{J1-E}\scalebox{0.8}{\textbf{NVS}}}} and \textit{\textbf{J1-E}\scalebox{0.8}{\textbf{VAL}}} are designed as simulation and evaluation tools to enhance legal professionals' capabilities and advance our understanding of legal AI systems. We acknowledge several important ethical considerations. First, while {\textit{\textbf{J1-E}\scalebox{0.8}{\textbf{NVS}}}} demonstrates promising results in simulating legal scenarios, it is not intended to replace human legal professionals or make actual legal decisions. The system should be used as a supplementary tool for legal training and research purposes only. Second, the legal knowledge and strategies learned by our system should not be misused for generating deceptive legal arguments. Furthermore, we emphasize that the system requires careful review and validation by qualified legal professionals before any practical application. The system's responses should not be considered as formal legal advice without proper human oversight.



\bibliography{custom}

\appendix
\section*{\centering Content of Appendix}

In this paper, we introduce \textit{\textbf{J1-E}\scalebox{0.8}{\textbf{NVS}}} and \textit{\textbf{J1-E}\scalebox{0.8}{\textbf{VAL}}} to encourage research on interactive legal intelligence. We showed that our framework can be used to evaluate interactive legal intelligence. In the appendix, we provide the following items for a detailed illustration:

\begin{itemize} 
    \item \ref{sec:related work} Additional Related work;
    \item \ref{sec:appendix:Role setting details} Role setting details;
    \item \ref{sec:appendix: Data statistics} Additional data statistics;
    \item \ref{sec:appendix:Evaluation Metric} Detailed illustration of evaluation metrics;
    \item \ref{sec:appendix: setting_details} Implementation details;
    \item \ref{sec:appendix:additional experiment} Additional experiment results;
    \item \ref{sec:appendix:human evaluation} Human evaluation;
    \item \ref{sec:appendix:prompts} Prompts.
\end{itemize}

\section{Additional Related Work}
\label{sec:related work}
\paragraph{Legal Intelligence.}
Legal AI applies artificial intelligence to enhance or automate various tasks in the legal domain, such as legal information extraction~\cite{yao2022leven,shen2024empowering}, judgment prediction~\cite{huang2021dependency}, and legal question answering~\cite{kien2020answering,zhong2020jec}.
Early approaches \cite{zhong2020jec,ji2021discrete}  rely on task-specific models and curated datasets for specific legal tasks.
With LLMs' strong generalizability~\cite{brown2020language}, they have shown notable capabilities in the legal domain. 
For example, Deng \textit{et al.} \cite{deng-etal-2023-syllogistic} introduced a structured prompting method to enhance judgment prediction. ChatLaw \cite{cui2023chatlaw} and
LawLLM \cite{yue2023disc,yue2024lawllm} enhance multitasking capabilities through training on legal knowledge and instructions.  
Recent explorations \cite{shengbinyue2025multi,chen2024agentcourt} have employed LLM-driven agents to simulate legal scenarios. However, a key challenge remains: the lack of a unified framework for evaluating agent performance and guiding future progress.

\begin{table*}[t]
\centering
\small
\resizebox{0.65\linewidth}{!}{
\begin{tabular}{lcccccc}
\toprule
\textbf{Level} 
& \multicolumn{2}{c}{\textbf{Level I}} 
& \multicolumn{2}{c}{\textbf{Level II}} 
& \multicolumn{2}{c}{\textbf{Level III}} \\
\hline

\textbf{Env.} 
& KQ & LC & CD & DD & CI & CR \\

\textbf{Count} 
& 98 & 62 & 93 & 93 & 93 & 69 \\
\midrule

\textbf{Source} 
& \multicolumn{2}{c}{ Legal Article} 
& \multicolumn{3}{c}{ Civil Documents} 
&  Criminal Documents \\
\bottomrule
\end{tabular}}
\caption{Data statistics of \textit{\textbf{J1-E}\scalebox{0.8}{\textbf{VAL}}}.}
\label{tab:env-data}
\end{table*}

\section{Role Setting Details}
\label{sec:appendix:Role setting details}
\subsection{Legal Sources Processing}
\label{sec:appendix:Legal Sources Processing}
We utilize two primary legal sources to construct role-specific settings.

\textbf{Chinese Judgment Documents} are official court records of judicial proceedings, outcomes, and factual details. We collect both civil and criminal judgment documents, and employ GPT-4o to extract legal elements.
For civil judgment documents, the following legal elements are extracted.
\begin{itemize}
    \item \textbf{Personal information} covers both individuals and legal entities. For individuals, we collect name, gender, ethnicity, birthdate, and residential address. For legal entities (corporations), personal information includes the corporate name, registered address, and the identity of the legal representative.

    \item \textbf{Case information} include plaintiff claims, case detail, defence strategies, and evidences.

    \item \textbf{Court Information} includes the final judgment, third-party findings, and legal provisions applied by the court in the case.
\end{itemize}
    
For criminal judgment documents, the following legal elements are extracted:
\begin{itemize}
    \item \textbf{Personal information} include the defendant's name, ethnicity, address, gender, birth date, occupation, level of education, and custody status.

    \item \textbf{Plea} captures the stated stance of the defendant and the lawyer towards the accusations.

    \item \textbf{Court information} contains the final verdict, court findings, and legal provisions cited by the judge.

    \item \textbf{Procurator information} comprises formal charges and procurator opinions.
\end{itemize}

\textbf{Chinese Legal Articles} are authored by legal experts to explain legal concepts, discuss controversial topics, and address issues in legal practice.
\begin{itemize}
    \item \textbf{\textit{Civil Trial Practice Q\&A} literature} is a curated collection of legal articles on civil trials practice, each offering an in-depth analysis of a specific legal issue. Leveraging GPT-4o, we identify lay person with relevant occupations, and construct a topic list in the format of <topic, answer> pair. The topics are then summarized into a shared theme.

    \item \textbf{\textit{HUALV} website} includes frontier legal articles, where each article is framed as a specific legal question accompanied by a detailed background description. We leverage GPT-4o to divide the article into two parts: the background part and the explanation part. The explanation part is utilized to construct <topic, answer> pairs.
\end{itemize}

To ensure diversity, we categorize topics into two categories: Binary(i.e., yes/no) topics and open-ended questions, where responses are expected to be either statutory provisions or legal phrases. Comprehensive statistics related to these topics are provided in Sec. \ref{sec:appendix: Data statistics}.

\subsection{Personality Theory Construction}
\label{sec:appendix:Personality Theory Construction}
Real-world clients exhibit behavioral styles according to occupation and experiences. Previous research\cite{yue2025synergistic, chen2024agentcourt} has demonstrated that LLM-driven agents could effectively follow predefined traits, which motivates our design. Therefore, we model the personality of lay person based on the Big Five Personality Traits theory, including five dimensions: extraversion, neuroticism, openness, agreeableness, and conscientiousness. Each dimension is divided into three levels (high, medium, and low). We leverage GPT-4o to assign a level based on the individual information of lay person. 
Specifically, in \textbf{(1) Knowledge Questioning}, occupations are primarily exploited as individual information. 
For \textbf{(2) Consultation}, background descriptions are leveraged to generate behavioral styles. 
For \textbf{(3) Complaint Drafting}, case details are utilized to construct behavioral styles.
In \textbf{(4) Defence Drafting}, personality modeling is based on the defendant's statement of defence.
To enhance the distinctness of the agents' portrayals, dimensions that are initially assigned to the medium level are randomly redistributed across the three levels following a 2:1:2 ratio.
Based on the five-dimensional traits, GPT-4o is leveraged to generate consistent behavioral styles.

\subsection{Role Profile Configuration}
\label{sec:appendix:Roles Profiles Configuration}
Based on the functional distinctions, we configure two types of roles, each associated with different sets of legal elements.

\textbf{\textit{Type I: Lay Person}} represent non-professional participants in legal scenarios, including general public in Level-I, plaintiff \& defendant in Level-II, and the defendant in Criminal Court:
\begin{itemize}
    \item \textbf{General Public} denotes clients of various professional and backgrounds seeking legal counseling on a particular theme(Env.1) or a specific case(Env.2). For Env.1, we assign each role agent a topic list together with a shared theme, and equip them with behavioral styles derived from the personality traits described in Sec.\ref{sec:appendix:Personality Theory Construction}. 
    For Env.2, detailed background information is additionally provided.

    \item \textbf{Specific Characters} denote the plaintiff and the defendant in Level-II document drafting scenarios. 
    We provide personal attributes (\textit{e.g.}, name and gender) as well as case-related information. 
    Specifically, plaintiffs are assigned claims and supporting evidence, while defendants are provided with statements of defence and corresponding evidence. 
    Behavioral traits are further incorporated to enhance the realism of simulation.

    \item \textbf{Defendant} denotes the accused in Criminal Court. The role is provided with case details and a stated stance.
\end{itemize}

\begin{algorithm}[t]
\caption{KQ/LC Scenario Simulation}
\label{alg:kq_lc}
\begin{algorithmic}[1]
\Require 
Legal agent $A_l$, General public agent $A_g$ with topic set $T$, 
Supervisor $S$, maximum dialogue turns $N$
\Ensure 
Dialogue history $H$

\State Initialize dialogue history $H \gets \emptyset$
\State Initialize unaddressed topic list $U \gets T$
\State Initialize turn counter $k \gets 0$

\While{$k < N$ \textbf{and} $U \neq \emptyset$}
    \State $k \gets k + 1$

    \State $q_k \gets A_g.\textsc{Ask}(U)$
    \State Append $q_k$ to $H$

    \State $a_k \gets A_l.\textsc{Answer}(q_k)$
    \State Append $a_k$ to $H$

    \State $S$ updates $U$ based on $(q_k, a_k)$
\EndWhile

\State \Return $H$
\end{algorithmic}
\end{algorithm}

\begin{algorithm}[!t]
\caption{CD/DD Scenario Simulation}
\label{alg:cd_dd}
\begin{algorithmic}[1]
\Require 
Lawyer agent $A_l$, Specific character agent $A_c$, 
legal document template $\mathcal{T}$, maximum dialogue turns $N$
\Ensure 
Generated legal document $D$

\State Provide template $\mathcal{T}$ to lawyer agent $A_l$
\State Initialize dialogue history $H \gets \emptyset$
\State Initialize turn counter $k \gets 0$

\While{$k < N$}
    \State $k \gets k + 1$

    \State $q_k \gets A_l.\textsc{Ask}(H, \mathcal{T})$
    \State Append $q_k$ to $H$

    \If{$q_k$ indicates end-of-dialogue}
        \State \textbf{break}
    \EndIf

    \State $a_k \gets A_c.\textsc{Answer}(q_k)$
    \State Append $a_k$ to $H$
\EndWhile

\State $D \gets A_l.\textsc{GenerateDocument}(\mathcal{T}, H)$
\State \Return $D$
\end{algorithmic}
\end{algorithm}

\begin{algorithm}[t]
\caption{CI/CR Scenario Simulation}
\label{alg:ci_cr}
\begin{algorithmic}[1]
\Require 
Judge agent $A_j$, participant agent set $\mathcal{A}$, 
court procedure $\mathcal{P}$, maximum dialogue turns $N$
\Ensure 
Dialogue history $H$, predicted judgment $V$

\State Provide court procedure $\mathcal{P}$ to judge agent $A_j$
\State Initialize dialogue history $H \gets \emptyset$
\State Initialize turn counter $k \gets 0$

\While{$k < N$}
    \State $k \gets k + 1$

    \State $u_k \gets A_j.\textsc{Speak}(\mathcal{P}, H)$
    \State Append $u_k$ to $H$

    \If{$u_k$ indicates end-of-trial}
        \State \textbf{break}
    \EndIf

    \State Identify addressee role $r_k$ from $u_k$
    \State Select agent $A_r \in \mathcal{A}$ with role $r_k$

    \State $v_k \gets A_r.\textsc{Respond}(u_k)$
    \State Append $v_k$ to $H$
\EndWhile

\State $V \gets A_j.\textsc{PredictJudgment}(H)$
\State \Return $(H, V)$
\end{algorithmic}
\end{algorithm}

\textbf{\textit{Type II: Legal Professionals}} represent trained practitioners in legal scenarios, including the lawyer agent in Level-II, the plaintiff's \& defendant's lawyer, the procurator, and the judge agent in Level-III,
\begin{itemize}
    \item \textbf{Lawyer agent} targets gathering information from the plaintiff/defendant, and drafting complaints/defences accordingly. The agent is provided with predefined legal document formats. 
    Additionally, the lawyer agent in Env.4 is supplemented with complaint information to support defence drafting.

    \item \textbf{Plaintiff's \& defendant's lawyers} are provided with client(Plaintiff \& defendant) information. The plaintiff's lawyer is equipped with the client's claims, while the defendant's lawyer is assigned defence strategies.

    \item \textbf{Procurator} is assigned charges and opinions.

    \item \textbf{Judge Agent} targets predicting judgments while following predefined court procedures. The agent is configured with procedural information and neutral third-party findings. 
\end{itemize}

\subsection{Pseudo-code}
\label{sec:pseudo-code}
The simulation pseudocodes are described in algorithm \ref{alg:kq_lc}, algorithm \ref{alg:cd_dd}, and algorithm \ref{alg:ci_cr}.

\section{Data Statistics}
\label{sec:appendix: Data statistics}
As shown in Table \ref{tab:env-data}, we meticulously select 322 quality data in \textit{\textbf{J1-E}\scalebox{0.8}{\textbf{val}}}, with 160 pieces of legal articles, 93 civil judgment documents(shared among CD, DD, and CI), and 69 criminal judgment documents, totaling 508 instances. Legal articles are then utilized for agent configurations in Knowledge Questioning and Legal Consultation, with 98 pieces for Knowledge Questioning and 62 pieces for Legal Consultation. Civil judgment documents are utilized for Complaint \& Defence Drafting, as well as the Civil Court. Criminal judgment documents are employed for Criminal Court setups. Therefore, we include 508 cases in \textit{\textbf{J1-E}\scalebox{0.8}{\textbf{val}}}.

In addition, for topic lists in Knowledge Questioning, 286 binary topics and 188 open-ended topics are designed, with an average of 4.84 topics for each case. For topic lists in Legal Consultation, we construct 116 binary topics and 107 open-ended topics, with an average of 3.60 topics for each of the cases. The balanced percentage and diversity of different topic types have ensured a comprehensive evaluation of the model-driven agent's performance.

\begin{table}[!t]
 \setlength{\tabcolsep}{2mm}
 \small
 \centering
 \resizebox{1\linewidth}{!}{
\begin{tabular}{l|lllll}
\hline
\multicolumn{1}{c|}{\multirow{3}{*}{Agent}} & \multicolumn{5}{c}{ Complaint Drafting}            \\ \cline{2-6} \multicolumn{1}{c|}{}     &   \multicolumn{5}{c}{II-2}                                                                                                                                                                              \\ \multicolumn{1}{c|}{} &  \multicolumn{1}{c}{\textbf{P}\scalebox{0.8}{\textbf{LA}}} & \multicolumn{1}{c}{\textbf{D}\scalebox{0.8}{\textbf{EF}}} & \multicolumn{1}{c}{\textbf{C}\scalebox{0.8}{\textbf{LA}}} & \multicolumn{1}{c}{\textbf{E}\scalebox{0.8}{\textbf{VI}}} & \multicolumn{1}{c}{\textbf{F}\scalebox{0.8}{\textbf{\&R}}}    \\ \hline
\rowcolor{gray!25}\multicolumn{6}{c}{Multilingual LLMs(close source)}                                                                                                                                                                                                                            \\
\multicolumn{1}{l|}{ GPT-4o}         & \underline{92.7}                                                    &  \textbf{96.3}                   & 97.3                   & 77.7                   & \underline{85.3}                    \\
\multicolumn{1}{l|}{ Claude-3.7}                                        &92.0                   & 93.1                   & 96.2                   & \textbf{85.2}                   & 79.6                                               \\ \hline
\rowcolor{gray!25}\multicolumn{6}{c}{Multilingual LLMs(open source)}                                                                                                                                                                                                                             \\
\multicolumn{1}{l|}{ Deepseek-v3\space$_{\mathrm{671B}}$}           & 71.4                   & 87.7                   & 67.3                   & 35.2                   & 56.5                                      \\
\multicolumn{1}{l|}{ Deepseek-r1\space$_{\mathrm{671B}}$}       & 75.3                   & 71.0                   & 44.3                   & 11.7                    & 29.8                                              \\
\multicolumn{1}{l|}{ Llama3.3-it.\space$_{\mathrm{70B}}$}  & 78.9                   & 93.1                   & 98.0                   & 66.7                   & \textbf{86.0}                               \\
\multicolumn{1}{l|}{ Qwen3\space$_{\mathrm{32B}}$}        & 74.0                   & 94.3                   & \textbf{98.5}                   & 76.0                   & 79.4                                    \\
\multicolumn{1}{l|}{ Gemma3-it.\space$_{\mathrm{27B}}$}       & 86.5                   & 94.4                   & \underline{98.4}                   & 74.7                   & 70.1                                                \\
\multicolumn{1}{l|}{ Qwen3\space$_{\mathrm{14B}}$}    & \textbf{94.0}                   & \underline{94.5}                   & 97.1                   & \underline{79.7}                   & 80.5                   \\
\multicolumn{1}{l|}{ Gemma3-it.\space$_{\mathrm{12B}}$}    & 80.4                   & 91.5                   & 96.9                   & 73.8                   & 78.2                              \\
\multicolumn{1}{l|}{ GLM4\space$_{\mathrm{9B}}$}         & 80.8                   & 83.9                   & 89.4                   & 73.6                   & 83.8                              \\
\multicolumn{1}{l|}{ Qwen3\space$_{\mathrm{8B}}$}     & 76.3                   & 77.1                   & 81.0                   & 60.9                   & 68.3                                    \\
\multicolumn{1}{l|}{ InternLM3-it.\space$_{\mathrm{8B}}$}    & 77.4                   & 52.1                   & 79.6                   & 77.0                   & 74.8                                     \\
\multicolumn{1}{l|}{ Ministral-it.\space$_{\mathrm{8B}}$} 
& -                       & 14.2                       & -                       & -                       & -                                                   \\
\multicolumn{1}{l|}{ Qwen2.5-it.\space$_{\mathrm{7B}}$}         & 84.2                   & 94.1                   & 91.8                   & 52.3                   & 84.7                                      \\
\multicolumn{1}{l|}{ Qwen3\space$_{\mathrm{4B}}$}          & 89.3                   & 91.1                   & 94.8                   & 76.1                   & 67.00                                  \\ \hline
\rowcolor{gray!25}\multicolumn{6}{c}{Legal-specific LLMs}                                                                                                                                                                                                                                                 \\
\multicolumn{1}{l|}{ LawLLM\space$_{\mathrm{13B}}$}              & 0.2                    & 2.1                      & 5.0                    & 4.0                      & 3.0                                         \\
\multicolumn{1}{l|}{ Chatlaw2\space$_{\mathrm{7B}}$}          & 6.9                    & 16.7                    & 23.3                    & 15.0                   & 23.0                           \\ \hline
\end{tabular}}%

\caption{Level II: \metricid{II-2} in Complaint Drafting.
where PLA, DEF, CLA, EVI and F\&R denote defendant information, claims, evidence, and fact \& reason respectively. }
\label{tab:DOC in Complaint Drafting}
\end{table}

\begin{table}[t]
\small
\centering
\setlength{\tabcolsep}{2mm}
\resizebox{0.73\linewidth}{!}{
\begin{tabular}{l|lll}
\hline
\multicolumn{1}{c|}{\multirow{3}{*}{Agent}} & \multicolumn{3}{c}{ Defence Drafting}                                                                                                                                      \\ \cline{2-4} \multicolumn{1}{c|}{} & \multicolumn{3}{c}{II-2}
\\ \multicolumn{1}{c|}{}            & \multicolumn{1}{c}{\textbf{R}\scalebox{0.8}{\textbf{ES}}} & \multicolumn{1}{c}{\textbf{D}\scalebox{0.8}{\textbf{EF}}} & \multicolumn{1}{c}{\textbf{E}\scalebox{0.8}{\textbf{VI}}}   \\ \hline
\rowcolor{gray!25}\multicolumn{4}{c}{Multilingual LLMs(close source)}                                                                                                                                                                     \\
\multicolumn{1}{l|}{ GPT-4o}                     & \textbf{92.5}                   & \textbf{86.6}                   & \underline{75.3}                            \\
\multicolumn{1}{l|}{ Claude-3.7}            & \underline{79.9}                   & \underline{85.3}                   & 41.1                                             \\ \hline
\rowcolor{gray!25}\multicolumn{4}{c}{Multilingual LLMs(open source)}                                                                                                                                                                      \\
\multicolumn{1}{l|}{ Deepseek-v3\space$_{\mathrm{671B}}$}         & 73.9                   & 74.3                   & \textbf{80.5}                   \\
\multicolumn{1}{l|}{ Deepseek-r1\space$_{\mathrm{671B}}$}         & 73.6                   & 53.7                   & 16.5                    \\
\multicolumn{1}{l|}{ Llama3.3-it.\space$_{\mathrm{70B}}$}  & 71.0                   & 71.9                   & 44.1                              \\
\multicolumn{1}{l|}{ Qwen3\space$_{\mathrm{32B}}$}         & 73.4                   & 79.3                   & 37.7                              \\
\multicolumn{1}{l|}{ Gemma3-it.\space$_{\mathrm{27B}}$}             & 78.9                   & 80.2                   & 47.3                        \\
\multicolumn{1}{l|}{ Qwen3\space$_{\mathrm{14B}}$}                              & 77.1                   & 76.2                   & 33.8                         \\
\multicolumn{1}{l|}{ Gemma3-it.\space$_{\mathrm{12B}}$}               & 68.3                   & 83.7                  & 39.6                          \\
\multicolumn{1}{l|}{ GLM4\space$_{\mathrm{9B}}$}            & 69.8                   & 61.6                  & 25.2                     \\
\multicolumn{1}{l|}{ Qwen3\space$_{\mathrm{8B}}$}     & 25.7                   & 27.9                   & 15.0                      \\
\multicolumn{1}{l|}{ InternLM3-it.\space$_{\mathrm{8B}}$}    & 47.4                   & 70.9                  & 43.2                          \\
\multicolumn{1}{l|}{ Ministral-it.\space$_{\mathrm{8B}}$}    & 14.1                   & 20.2                   & 7.6                     \\
\multicolumn{1}{l|}{ Qwen2.5-it.\space$_{\mathrm{7B}}$}      & 69.3                   & 73.6                 & 36.3                    \\
\multicolumn{1}{l|}{ Qwen3\space$_{\mathrm{4B}}$}        & 68.8                  & 56.1                   & 29.7                 \\ \hline
\rowcolor{gray!25}\multicolumn{4}{c}{Legal-specific LLMs}                                                                                                                                                                                          \\
\multicolumn{1}{l|}{ LawLLM\space$_{\mathrm{13B}}$}                                & -                       & -                       & -         \\
\multicolumn{1}{l|}{ Chatlaw2\space$_{\mathrm{7B}}$}           & 1.1                    & 1.6                    & -                   \\ \hline
\end{tabular}}
\caption{Level II: \metricid{II-2} in Defence Drafting,
where RES, DEF and EVI denote respondent, defence claims, evidence respectively.}
\label{tab:Defence}
\end{table}

\begin{table}[t]
\setlength{\tabcolsep}{2mm}
\centering
\small
\resizebox{0.74\linewidth}{!}{
\begin{tabular}{llll}
\hline
\multicolumn{1}{c|}{\multirow{3}{*}{Agent}} & \multicolumn{3}{c}{ Civil Court}                                                                                                                                                                                                        \\ \cline{2-4} \multicolumn{1}{c|}{} & \multicolumn{3}{c}{III-1} 
\\
\multicolumn{1}{c|}{}                       & \multicolumn{1}{c}{ \textbf{S}\scalebox{0.8}{\textbf{TA}}} & \multicolumn{1}{c}{ \textbf{A}\scalebox{0.8}{\textbf{CT}}} & \multicolumn{1}{c}{ \textbf{U}\scalebox{0.8}{\textbf{NI}}} \\ \hline
\rowcolor{gray!25}\multicolumn{4}{c}{ Multilingual LLMs(close source)}                                                                                                                                                                                                                                  \\
\multicolumn{1}{l|}{ GPT-4o}                 & \textbf{85.8}                   & \textbf{88.2}                   & \textbf{89.2}                \\
\multicolumn{1}{l|}{ Claude-3.7}             &         \underline{65.3}           &    69.1           &     68.8          \\
\rowcolor{gray!25}\multicolumn{4}{c}{ Multilingual LLMs(open source)}                                                                                                                                                                                                                                   \\ \hline
\multicolumn{1}{l|}{ Deepseek-v3\space$_{\mathrm{671B}}$}       & 36.6                   & 46.5   & 37.6                                    \\
\multicolumn{1}{l|}{ Deepseek-r1\space$_{\mathrm{671B}}$}       & -                     & 0.2                      & 1.1                                \\
\multicolumn{1}{l|}{ Llama3.3-it.\space$_{\mathrm{70B}}$}  & 53.8                   & \underline{78.9}                   & \underline{86.0}                   \\
\multicolumn{1}{l|}{ Qwen3\space$_{\mathrm{32B}}$}     & 49.7                   & 57.5                   & 60.2                     \\
\multicolumn{1}{l|}{ Gemma3-it.\space$_{\mathrm{27B}}$}          & 68.5                   & 75.4                   & 67.7                 \\
\multicolumn{1}{l|}{ Qwen3\space$_{\mathrm{14B}}$}     & 11.6                    & 19.6                   & 25.8                                               \\

\multicolumn{1}{l|}{ Gemma3-it.\space$_{\mathrm{12B}}$}          & 36.8                   & 52.2                   & 54.8                        \\
\multicolumn{1}{l|}{ GLM4\space$_{\mathrm{9B}}$}                & 11.0                       & 14.3                       & 18.3                           \\
\multicolumn{1}{l|}{ Qwen3\space$_{\mathrm{8B}}$}      & 9.1                    & 10.6                    & 28.0                   \\
\multicolumn{1}{l|}{ InternLM3-it.\space$_{\mathrm{8B}}$}  & 25.5                    & 38.1                    & 47.3                          \\
\multicolumn{1}{l|}{ Ministral-it.\space$_{\mathrm{8B}}$}  & 18.8                    & 26.6                    & 32.3                       \\
\multicolumn{1}{l|}{ Qwen2.5-it.\space$_{\mathrm{7B}}$}    & 11.0                   & 22.7                   & 32.3                 \\
\multicolumn{1}{l|}{ Qwen3\space$_{\mathrm{4B}}$}      & 26.6                   & 33.4                   & 38.7                 \\ \hline
\rowcolor{gray!25}\multicolumn{4}{c}{ Legal LLMs}                                                                                                                                                                                                                                                       \\
\multicolumn{1}{l|}{ LawLLM\space$_{\mathrm{13B}}$}             & 0.5                       & 1.5                       & 4.3                    \\
\multicolumn{1}{l|}{ Chatlaw2\space$_{\mathrm{7B}}$}            & 0.8                       & 2.2                       & -                  \\ \hline
\end{tabular}}

\caption{Level III: \metricid{III-1} in Civil Court, where STA and ACT denote stages and actions respectively, and UNI is the binary indicator.}
\label{tab:Civil Court}
\end{table}

\begin{table}[t]
        
\setlength{\tabcolsep}{2mm}
\small
\centering
\resizebox{\linewidth}{!}{
\begin{tabular}{l|ccccc}
\hline
\multirow{3}{*}{\hspace{2.0em}Agent} 
 & \multicolumn{5}{c}{ Criminal Court} \\
\cline{2-6}
& \multicolumn{3}{c}{III-1} & \multicolumn{2}{c}{\textcolor{red}{III-4}} \\
& \textbf{S\scriptsize{TA}} & \textbf{A\scriptsize{CT}} & \textbf{U\scriptsize{NI}} & \textcolor{red}{\textbf{S\scriptsize{EN}}} & \textcolor{red}{\textbf{F\scriptsize{INE}}} \\
\hline
\rowcolor{gray!25}\multicolumn{6}{c}{Multilingual LLMs (closed source)} \\
 GPT-4o & 46.4 & 52.2 & 66.7 & 38.5 & 38.5 \\
Claude-3.7 & \underline{47.8} & \underline{59.6} & \textbf{95.7} & \textbf{21.6} & \textbf{9.8} \\
\rowcolor{gray!25}\multicolumn{6}{c}{Multilingual LLMs (open source)} \\
 Deepseek-v3\space$_{\mathrm{671B}}$ & 39.1 & 45.6 & 78.3 & 34.9 & 32.9 \\
 Deepseek-r1\space$_{\mathrm{671B}}$ & 2.2 & 2.6 & 1.4 & 98.6 & 98.6 \\
 Llama3.3-it.\space$_{\mathrm{70B}}$ & 44.9 & 56.5 & 85.5 & \textbf{21.6} & 28.4 \\
 Qwen3\space$_{\mathrm{32B}}$ & 25.4 & 28.2 & 50.7 & 60.4 & 56.3 \\
 Gemma3-it.\space$_{\mathrm{27B}}$ & \textbf{52.9} & \textbf{67.1} & \underline{92.8} & \underline{21.9} & \underline{22.3} \\
 Qwen3\space$_{\mathrm{14B}}$ & 15.2 & 18.7 & 29.0 & 73.4 & 73.7 \\
 Gemma3-it.\space$_{\mathrm{12B}}$ & 37.0 & 45.7 & 69.6 & 41.1 & 40.4 \\
 GLM4\space$_{\mathrm{9B}}$ & 21.7 & 28.5 & 39.1 & 76.2 & 73.8 \\
 Qwen3\space$_{\mathrm{8B}}$ & 18.8 & 14.8 & 37.7 & 67.8 & 67.0 \\
 InternLM3-it.\space$_{\mathrm{8B}}$ & 27.5 & 30.8 & 60.9 & 45.4 & 44.9 \\
 Ministral-it.\space$_{\mathrm{8B}}$ & 2.2 & 1.9 & 4.3 & 97.7 & 98.3 \\
 Qwen2.5-it.\space$_{\mathrm{7B}}$ & 21.0 & 31.7 & 52.2 & 56.0 & 53.0 \\
 Qwen3\space$_{\mathrm{4B}}$ & 23.2 & 18.5 & 49.3 & 59.1 & 55.5 \\
\rowcolor{gray!25}\multicolumn{6}{c}{Legal LLMs} \\
 LawLLM\space$_{\mathrm{13B}}$ & 11.6 & 16.1 & 21.7 & 84.3 & 90.6 \\
 Chatlaw2\space$_{\mathrm{7B}}$ & - & - & - & - & - \\
\hline
\end{tabular}}
\caption{Level III: \metricid{III-2} and \metricid{III-4} in Criminal Court, where STA, ACT, SEN, and FINE denote stages, actions, sentences, and fines respectively, and UNI is the binary indicator.}
\label{tab: Criminal Court}
\end{table}

\section{Evaluation Metric}
\label{sec:appendix:Evaluation Metric}
We construct a comprehensive evaluation framework that assesses the performance of the legal agent from multiple perspectives with a set of metrics.

\subsection{Level-I}
\begin{figure*}[t]
    \centering
\includegraphics[width=0.9\linewidth]{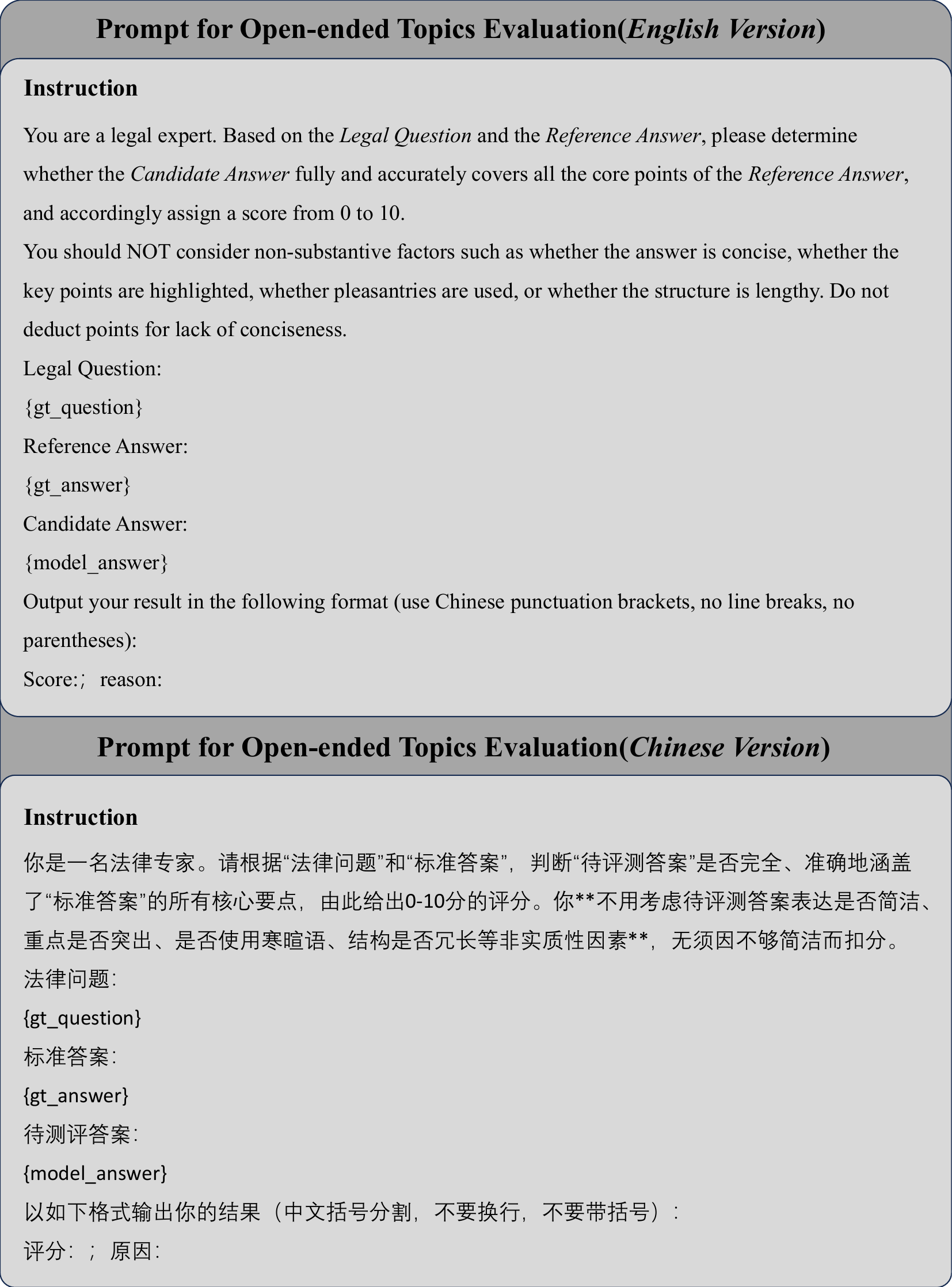}
    \caption{Prompt for open-ended topics evaluation}
    \label{fig:prompt of open-ended topics evaluation in chinese}
\end{figure*}

We leverage an LLM-based check mechanism to label each interaction round with its corresponding topic.
As illustrated in Sec.\ref{sec:appendix:Legal Sources Processing}, the topics are divided into binary type and open-ended type, which are evaluated for \textbf{Binary accuracy} and \textbf{Non-binary score} respectively. To be specific:

\begin{itemize}
    \item \textbf{Binary accuracy} measures accuracy in binary questions via exact matching. By comparing the summarized yes/no answer produced by the checking mechanism with the ground truth, the score is assigned as 1 for correct responses, 0 for incorrect responses, and 0.5 for conditionally acceptable answers.

    \item \textbf{Non-binary score} evaluates open-ended responses of the legal agent. When the ground truth consists of statutory provisions, we apply a matching-based evaluation, assigning a score of 1 if the relevant provisions are correctly cited and 0 otherwise. 
    For ground truths involving legal phrases, we leverage GPT-4o to produce a score ranging from 0 to 10, which is normalized to the interval [0,1]. The evaluation prompt is shown in Figure \ref{fig:prompt of open-ended topics evaluation in chinese}.
\end{itemize}

\subsection{Level-II}
We evaluate the legal document drafted by the legal agent from two perspectives.

\begin{figure*}[t]
    \centering
\includegraphics[width=0.8\linewidth]{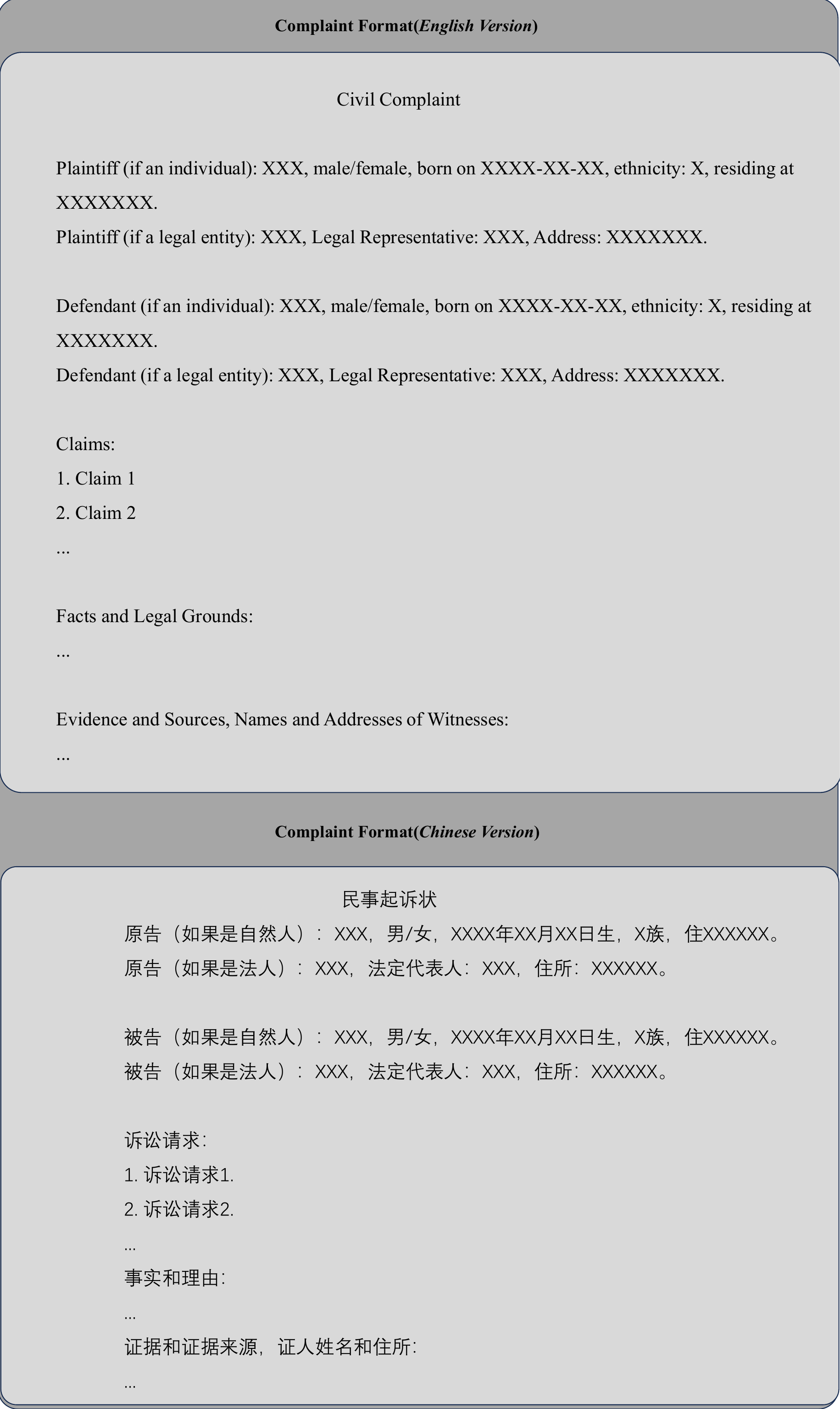}
    \caption{Complaint Format}
    \label{fig:complaint format}
\end{figure*}

\begin{figure*}[t]
    \centering
\includegraphics[width=0.9\linewidth]{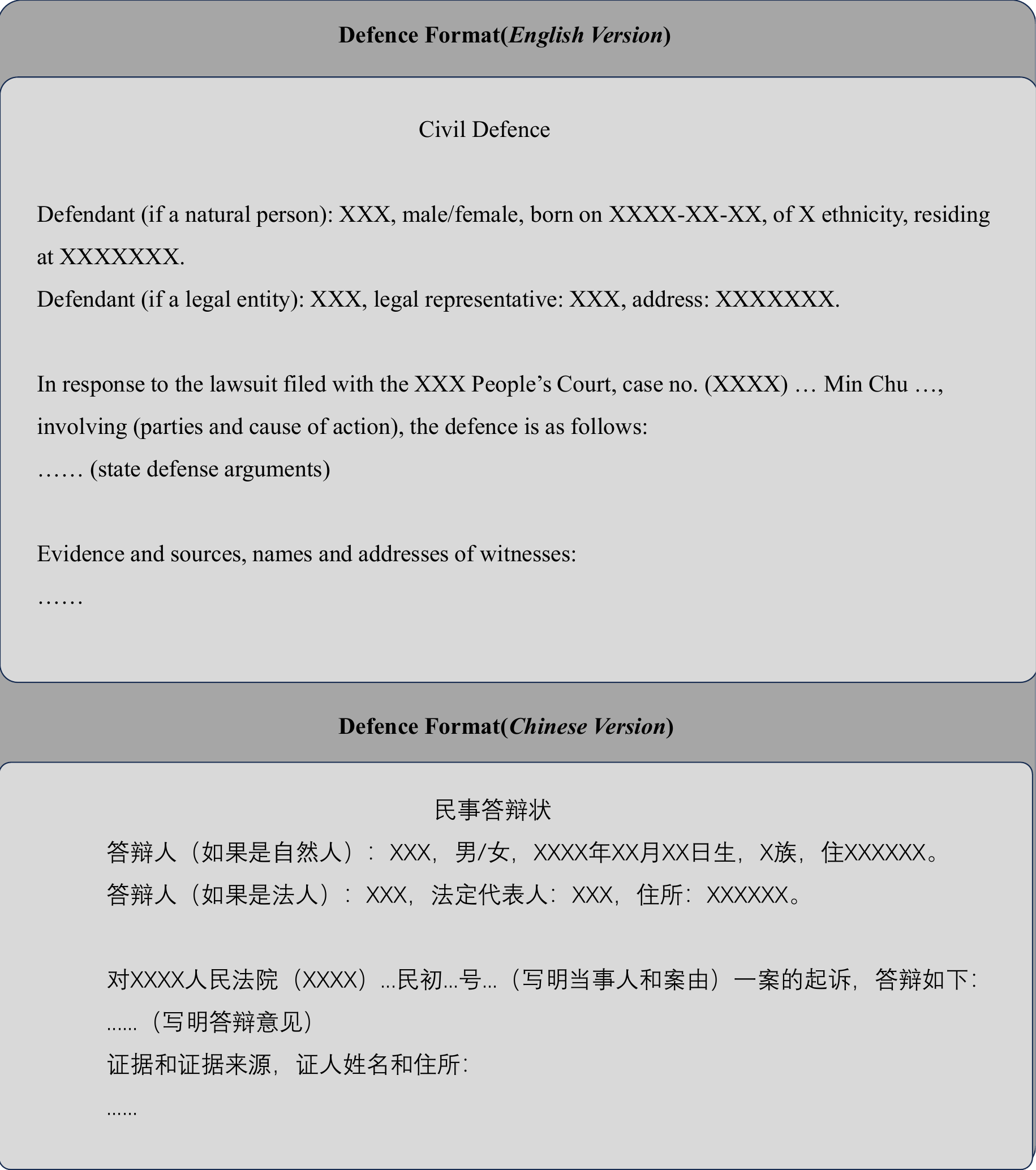}
    \caption{Defence Format}
    \label{fig:defence format}
\end{figure*}

\textbf{(1) Format-following score} measures the alignment with the given format. We break down the format into a set of labels $L=[L_1,L_2,...,L_k]$, which are organized in a specific sequential order $S$. As shown in Figure \ref{fig:complaint format}, the complaint format includes labels of \textit{Plaintiff}, \textit{Defendant}, \textit{Claims}, \textit{Facts and Legal Grounds}, and \textit{evidence and Sources, Names and Addresses of Witness}. As shown in Figure \ref{fig:defence format}, the defence format includes labels of \textit{Defendant}, \textit{defence arguments}, \textit{case id}, \textit{parties and cause of action}, \textit{evidence and Sources, names and addresses of witnesses}. Through matching, we check whether the generated sequential order $S_{agent}$ follows the given order $S$, and score the accuracy of labels by comparing the generated labels $L_{agent}$ with $L$. The final result is calculated with the following equation.
\begin{equation}
    FOR = Seq * Label
\end{equation}
Where: $FOR$ is \metricid{II-1}; $Seq$ is 1 if $S_{agent}$ is identical with $S$, and 0 if not; $Label$ is the accuracy of labels $S_{agent}$.

\textbf{(2) Document score} evaluates the accuracy of the information in the drafted documents. For the plaintiff and the defendant(\textit{i.e., name, ethnicity, address, birthdate, gender}), we apply the exact match method(1 for correct and 0 for incorrect).
For other components(i.e., claims, evidences, statement of defence, and case details), we apply the LLM-as-Judge method. 
The evaluation prompt is provided in Figure \ref{fig:prompt of comparative evaluation chinese}.

\subsection{Level-III}
We evaluate the legal agents' performance in Civil Court and Criminal Court from four dimensions.

\begin{figure*}[t]
    \centering
\includegraphics[width=0.9\linewidth]{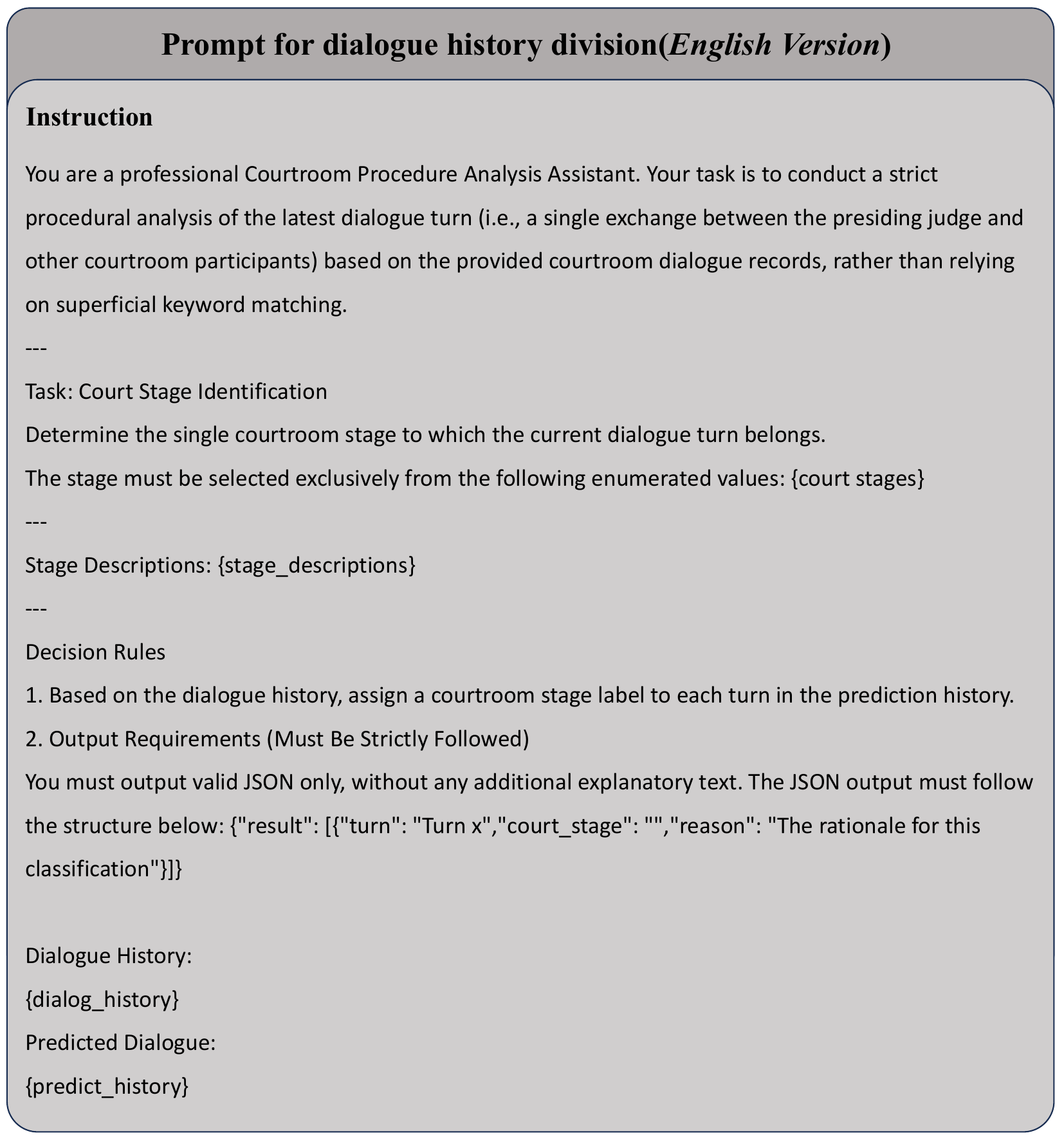}
    \caption{Prompt for Dialogue History Division(\textit{English Version})}
    \label{fig:prompt of dialogue history division english}
\end{figure*}

\begin{figure*}[t]
    \centering
\includegraphics[width=0.9\linewidth]{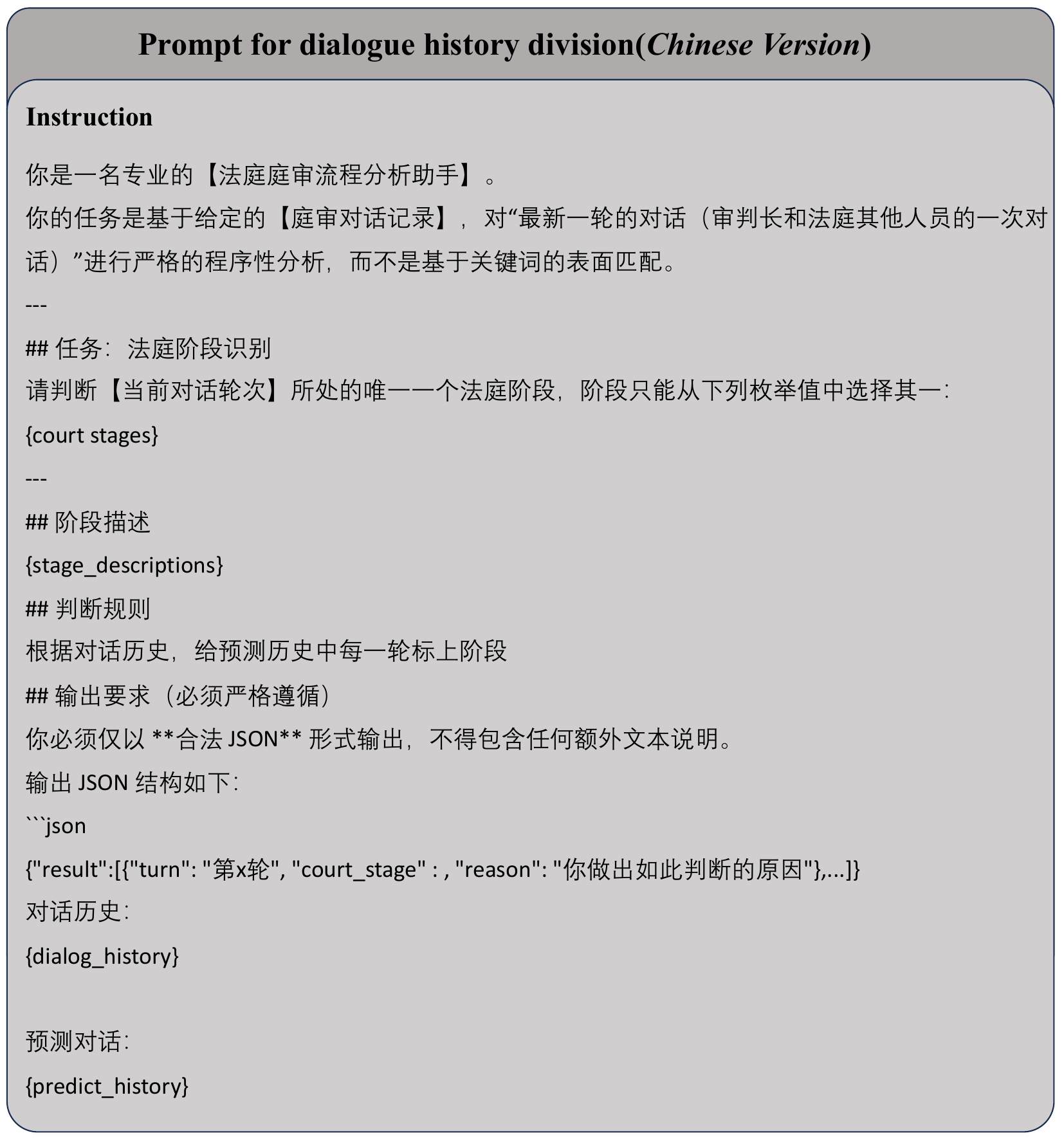}
    \caption{Prompt for Dialogue History Division(\textit{Chinese Version})}
    \label{fig:prompt of dialogue history division chinese}
\end{figure*}

\begin{figure*}[t]
    \centering
\includegraphics[width=0.8\linewidth]{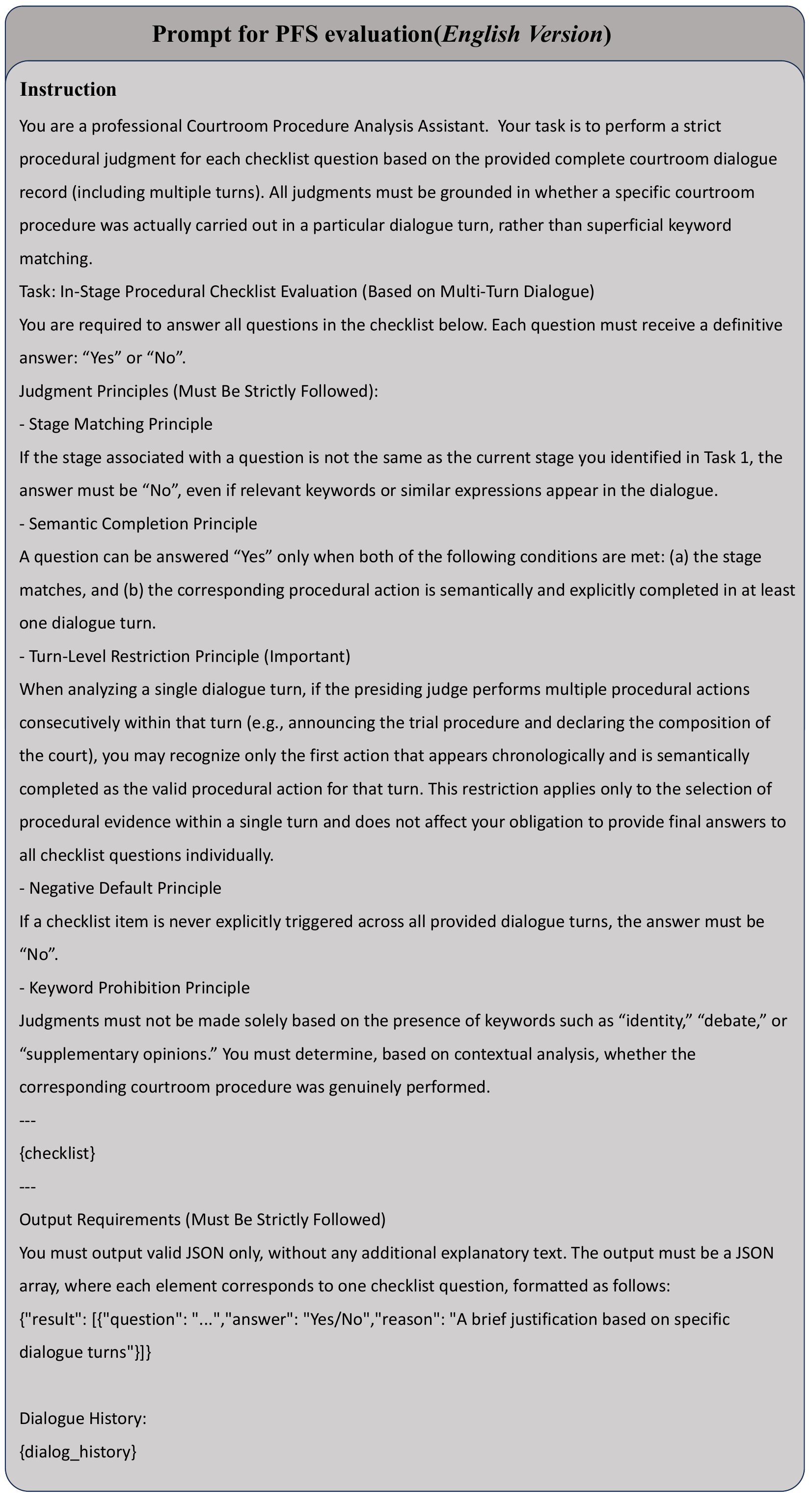}
    \caption{Prompt for \metricid{III-1} evaluation(\textit{English Version})}
    \label{fig:prompt of pfs evaluation english}
\end{figure*}

\begin{figure*}[t]
    \centering
\includegraphics[width=0.8\linewidth]{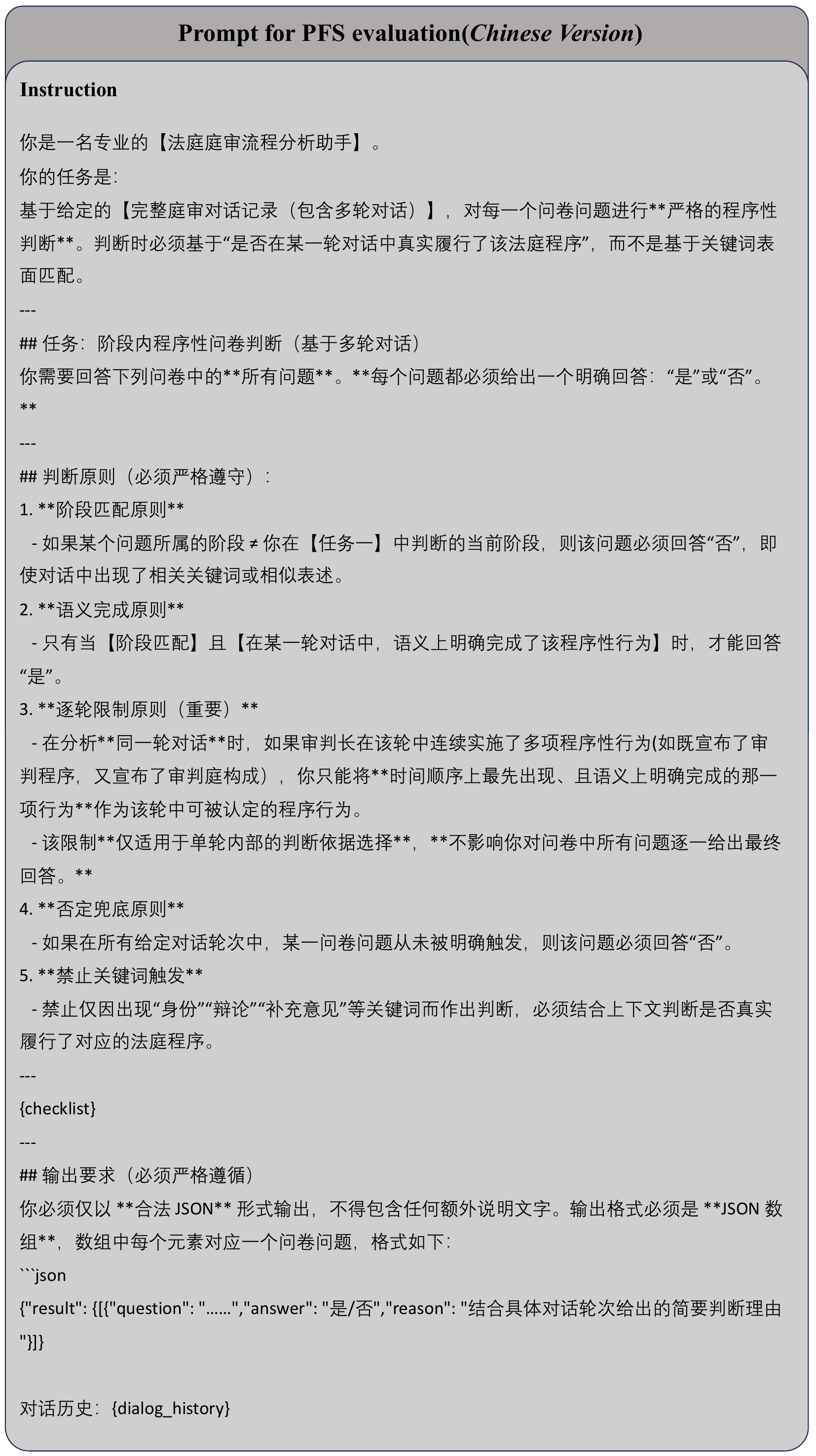}
    \caption{Prompt for \metricid{III-1} evaluation(\textit{Chinese Version})}
    \label{fig:prompt of pfs evaluation chinese}
\end{figure*}

\begin{figure*}[t]
    \centering
\includegraphics[width=0.8\linewidth]{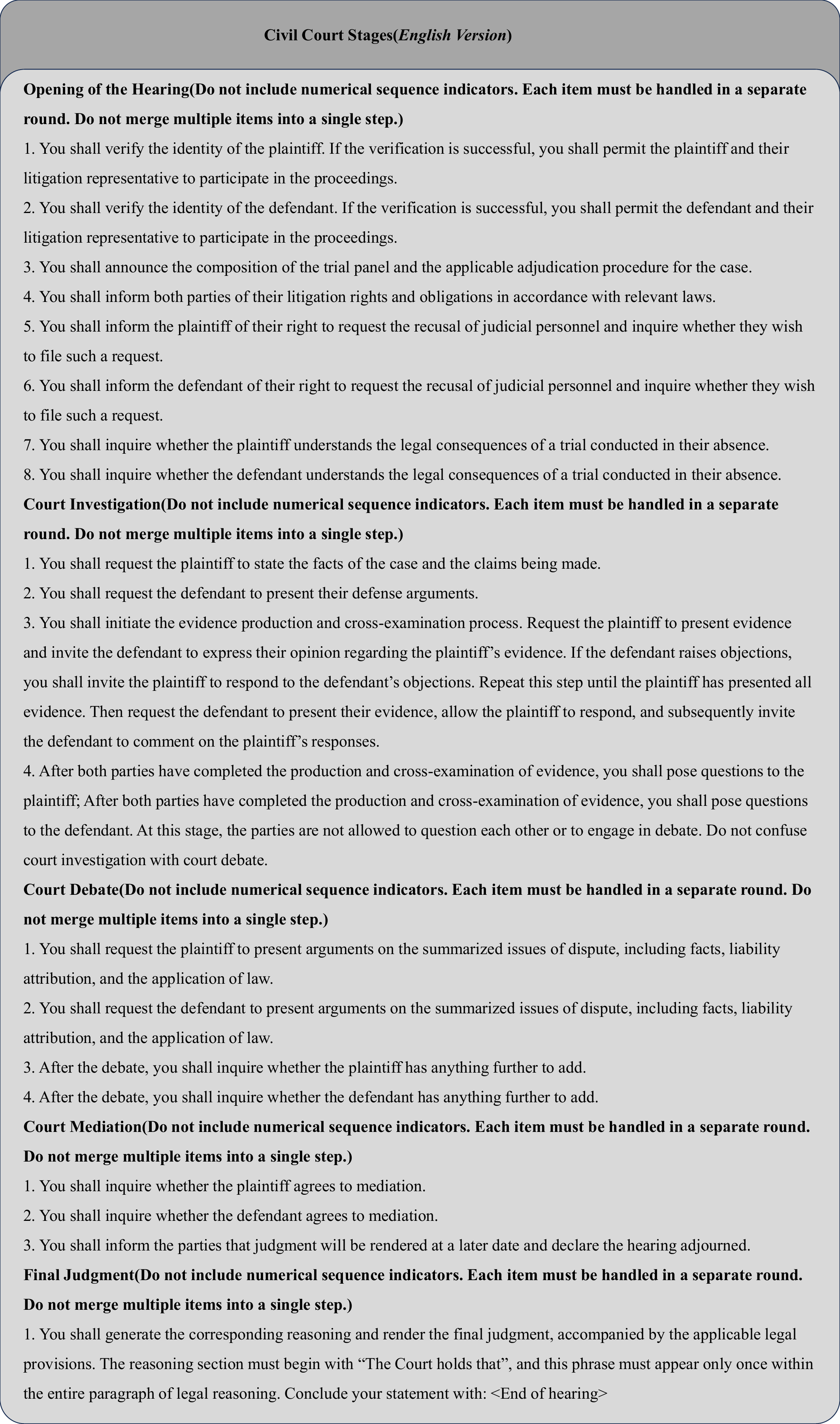}
    \caption{Civil Court Stages and Actions(\textit{English Version})}
    \label{fig:civil court stages and actions english}
\end{figure*}

\begin{figure*}[t]
    \centering
\includegraphics[width=0.9\linewidth]{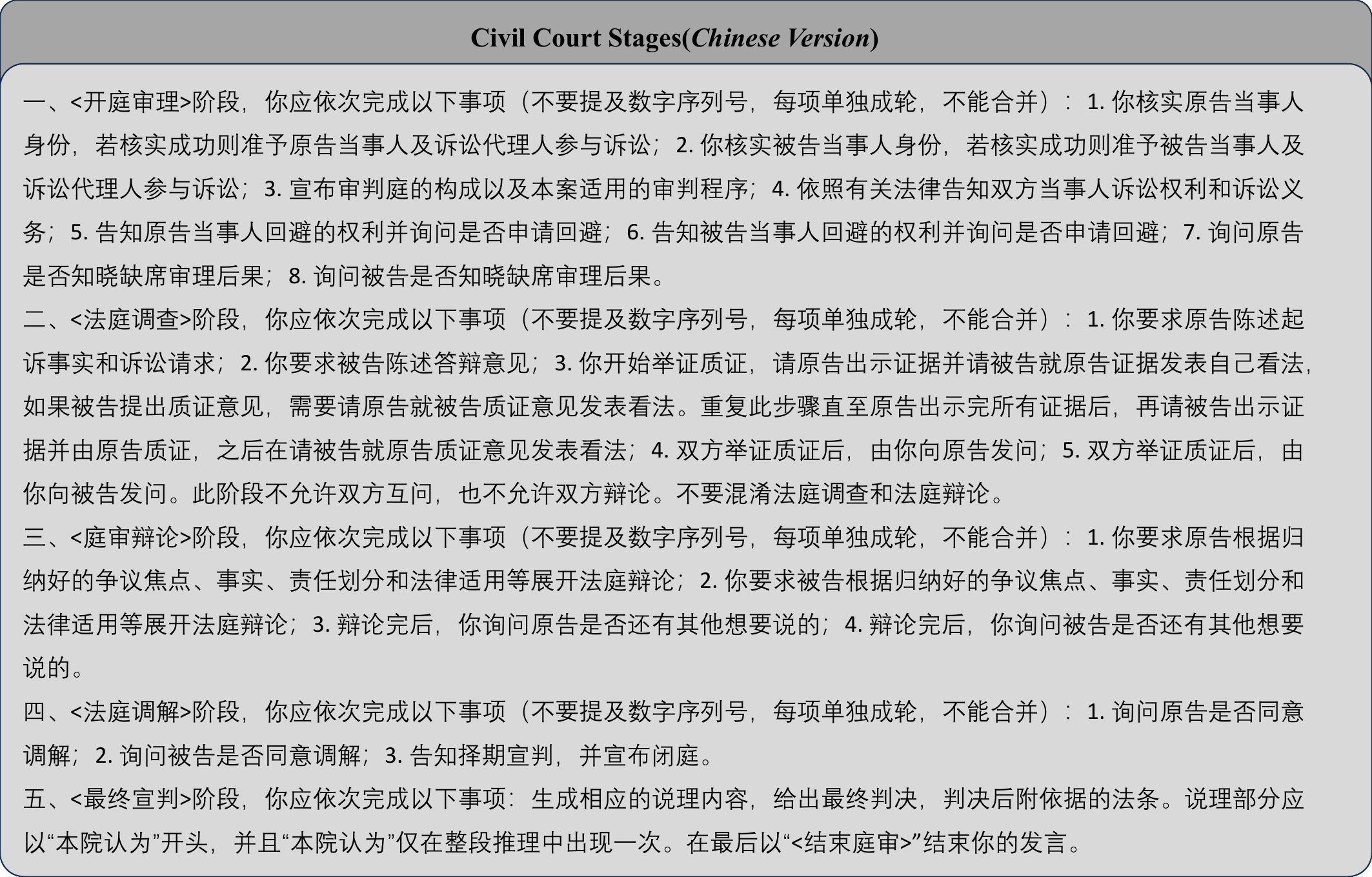}
    \caption{Civil Court Stages and Actions(\textit{Chinese Version})}
    \label{fig:civil court stages and actions chinese}
\end{figure*}

\begin{figure*}[t]
    \centering
\includegraphics[width=0.9\linewidth]{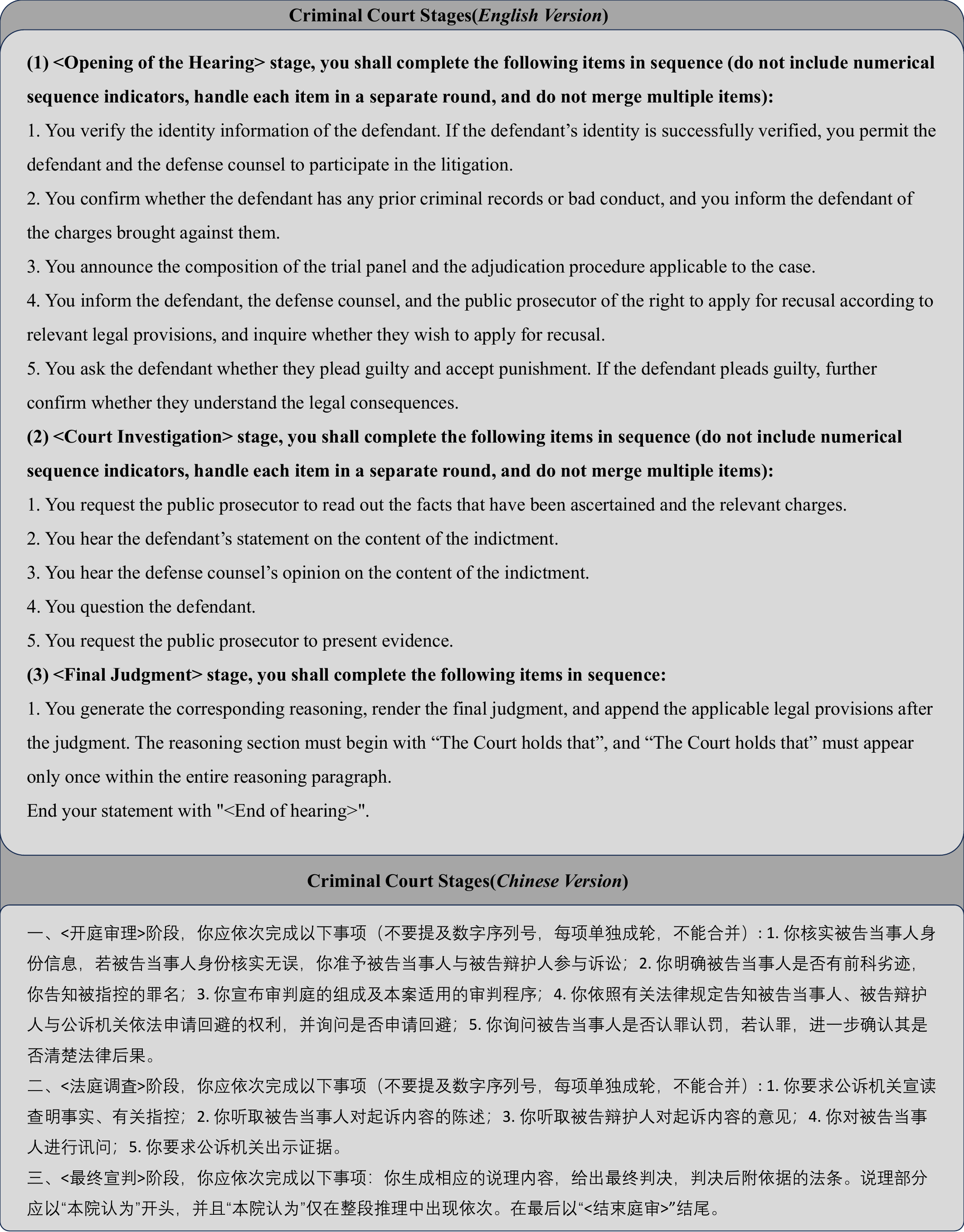}
    \caption{Criminal Court Stages and Actions}
    \label{fig:criminal court stages and actions}
\end{figure*}

\begin{figure*}[t]
    \centering
\includegraphics[width=0.8\linewidth]{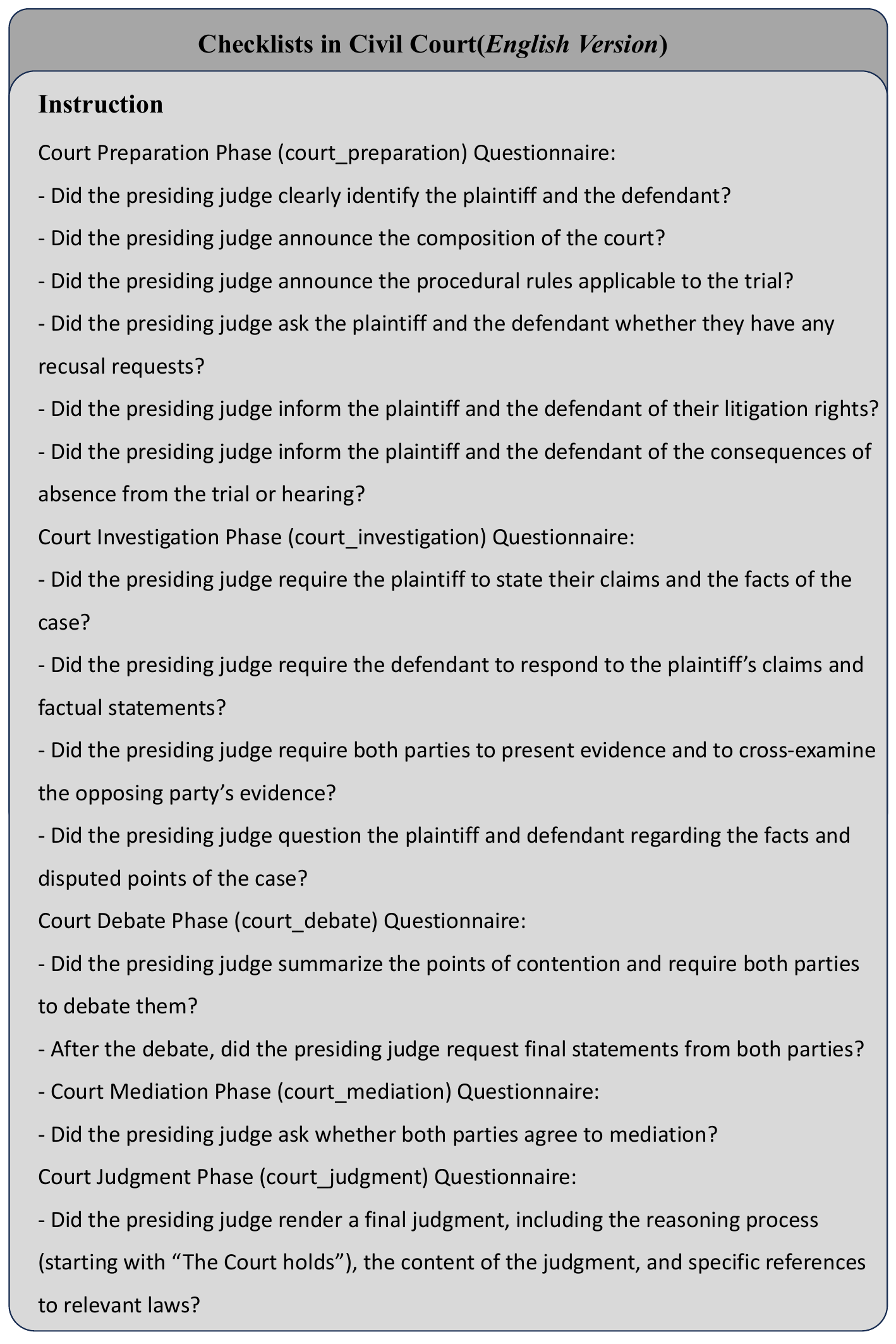}
    \caption{Civil Court Checklists(\textit{English Version})}
    \label{fig:civil checklist english}
\end{figure*}

\begin{figure*}[t]
    \centering
\includegraphics[width=0.9\linewidth]{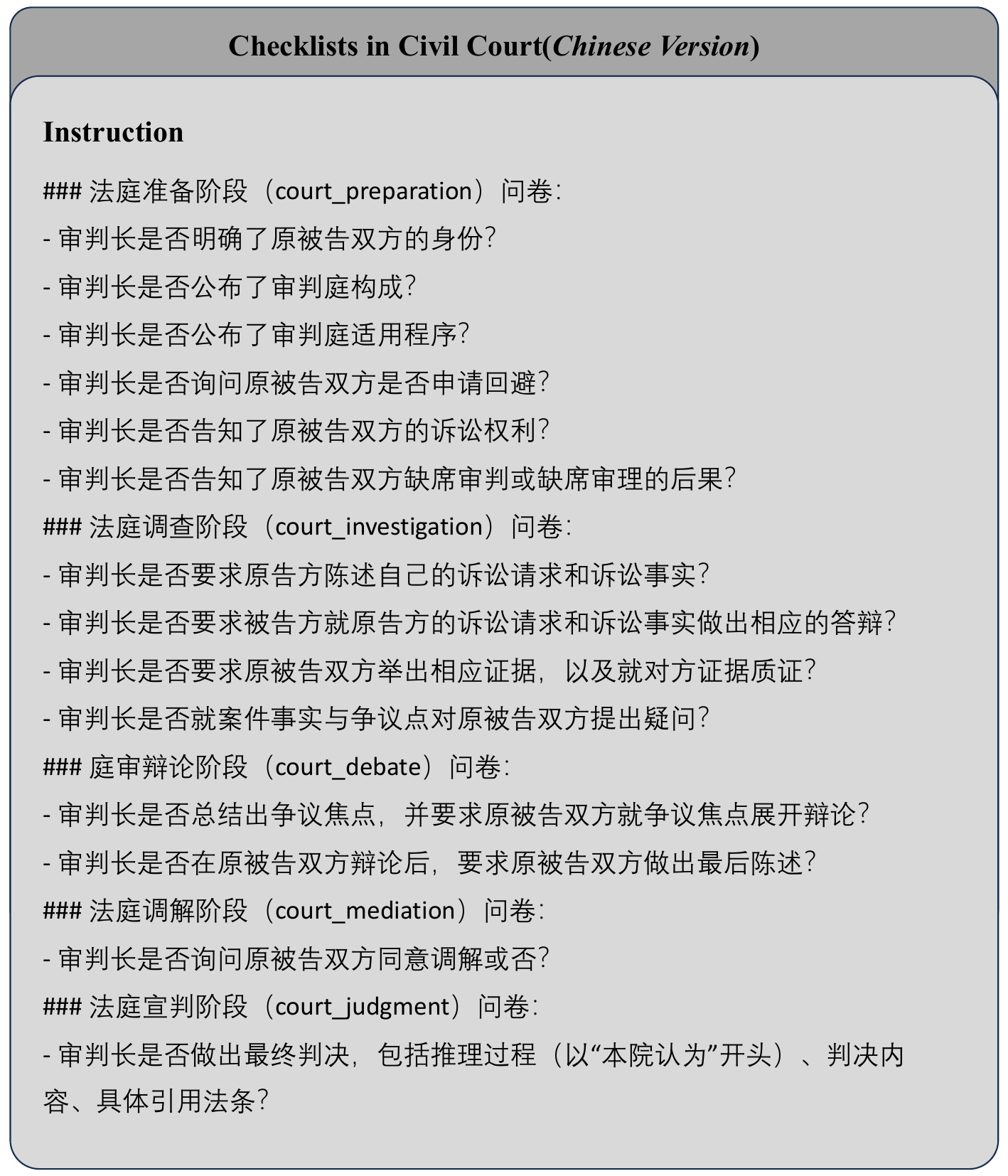}
    \caption{Civil Court Checklists(\textit{Chinese Version})}
    \label{fig:civil checklist chinese}
\end{figure*}

\begin{figure*}[t]
    \centering
\includegraphics[width=0.8\linewidth]{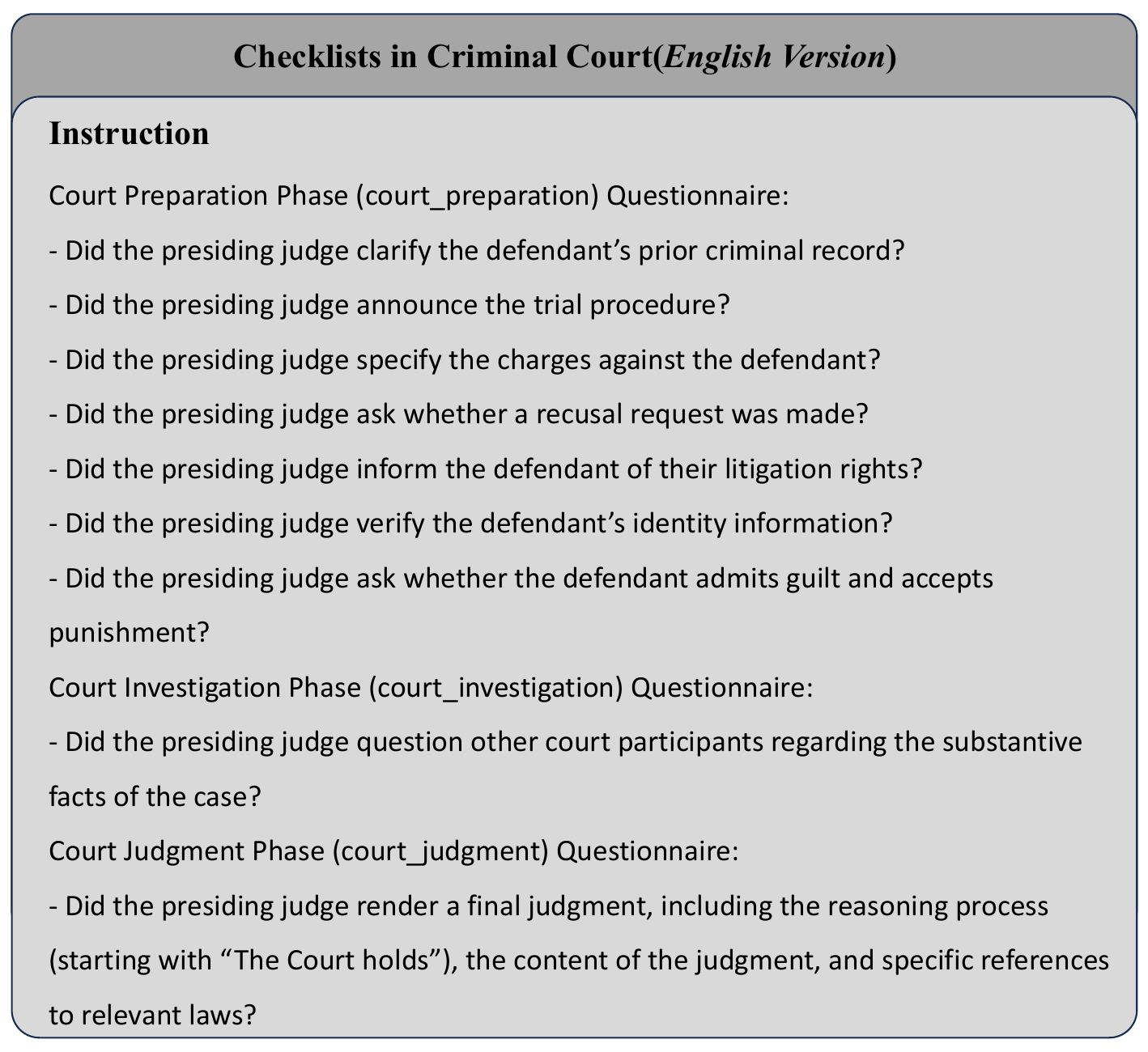}
    \caption{Criminal Court Checklists(\textit{English Version})}
    \label{fig:criminal court checklist english}
\end{figure*}

\begin{figure*}[!tbp]
    \centering
\includegraphics[width=0.8\linewidth]{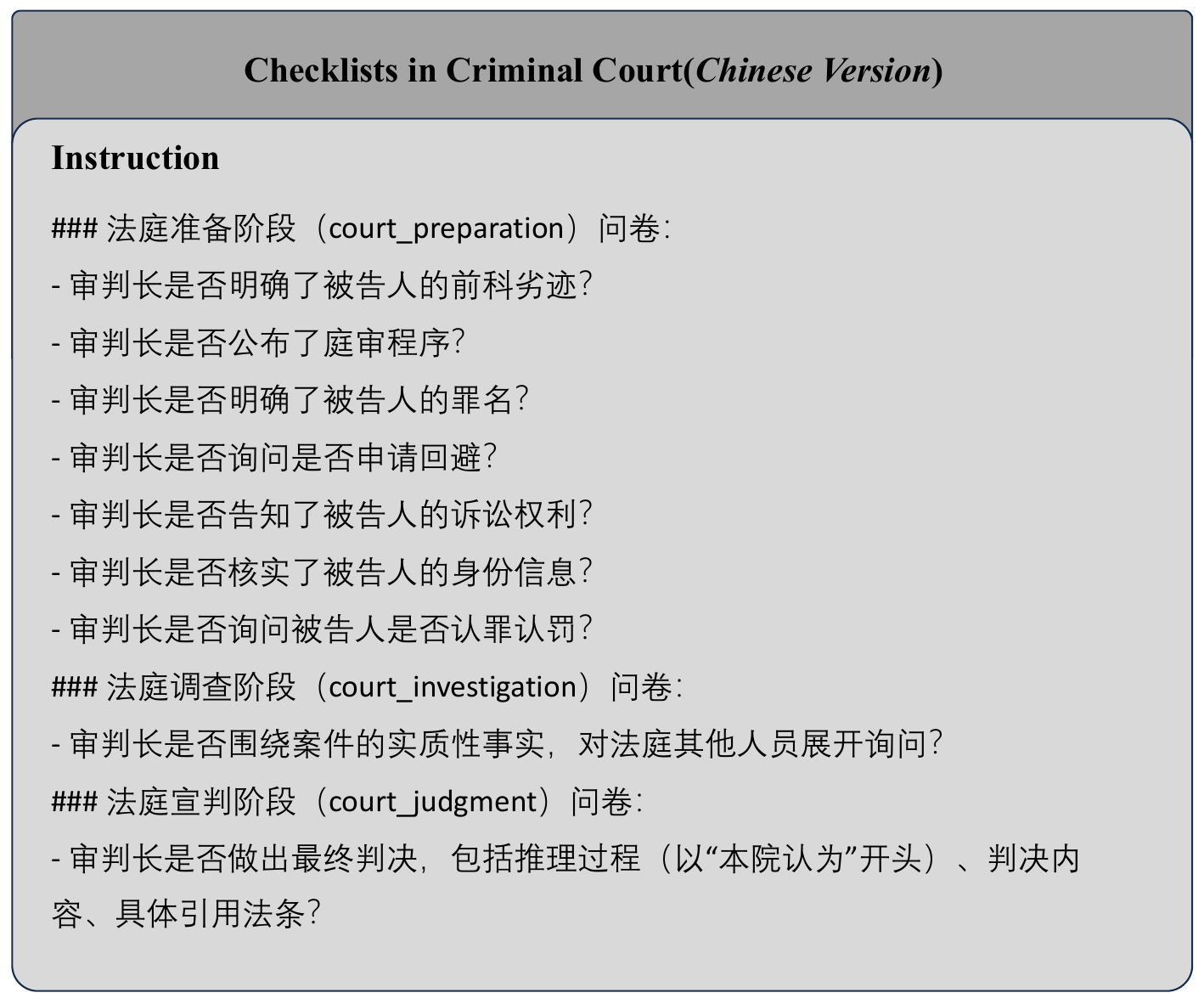}
    \caption{Criminal Court Checklists(\textit{Chinese Version})}
    \label{fig:criminal court checklist chinese}
\end{figure*}

\textbf{(1) Procedural-following score} evaluates the completeness of the procedural stages, each with several mandatory actions. A stage is complete ONLY when all mandatory actions are executed. 
In the civil court setting, proceedings are divided into five stages(\textit{i.e.}, court preparation, court investigation, court debate, court mitigation, and court decision). Descriptions of stages \& actions are provided in Figure \ref{fig:civil court stages and actions english} and Figure \ref{fig:civil court stages and actions chinese}.
In the criminal court setting, proceedings are divided into three stages(\textit{i.e.}, court preparation, court investigation, and court decision). Descriptions of stages \& actions are provided in Figure \ref{fig:criminal court stages and actions}.
If a legal agent delivers unilateral speech that executes all procedural stages within a single turn, the performance is regarded as a failure under our evaluation protocol.
The completeness of procedural stages and mandatory actions is evaluated using the LLM-as-Judge method with a checklist-style framework. Specifically, we first segment the dialogue history into procedural units using the prompts shown in Figure \ref{fig:prompt of dialogue history division english} and Figure \ref{fig:prompt of dialogue history division chinese}. We then assess each unit with a checklist of binary(Yes/No) questions. To mitigate the risk of keyword hacking, we introduce a general requirement instructing the model to evaluate cases according to their overall semantic content. The evaluation prompts and checklists are provided in Fig. \ref{fig:prompt of pfs evaluation english} and \ref{fig:prompt of pfs evaluation chinese}, Fig. \ref{fig:civil checklist english} to \ref{fig:criminal court checklist chinese}.

\begin{figure*}[t]
    \centering
\includegraphics[width=0.9\linewidth]{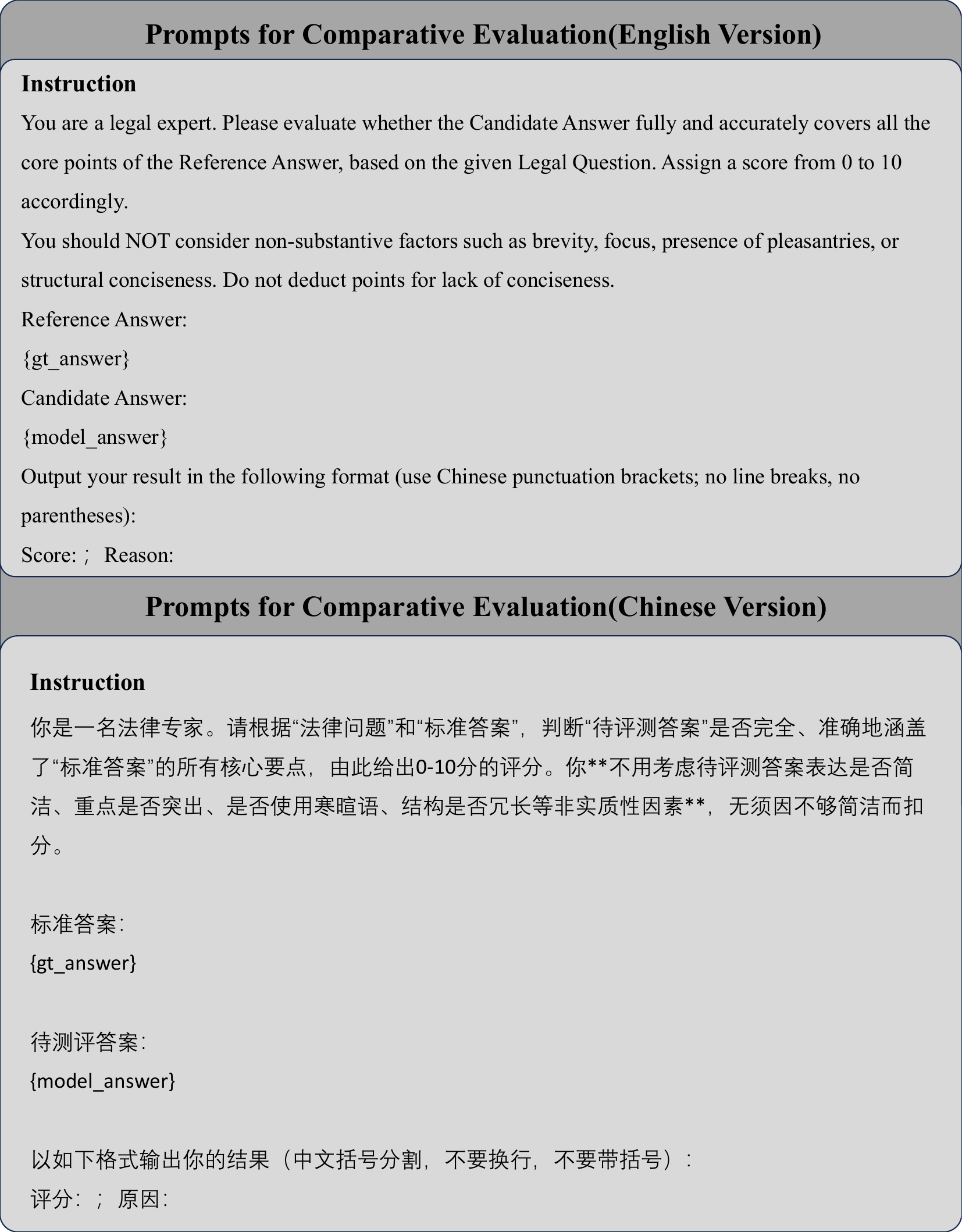}
    \caption{Prompt for comparative evaluation}
    \label{fig:prompt of comparative evaluation chinese}
\end{figure*}

\textbf{(2) Judgment score} evaluates the quality of the legal agent's judgment. In civil court, we evaluate the judgment based on the semantic alignment with the ground truth, and the prompt is provided in Figure \ref{fig:prompt of comparative evaluation chinese}. In criminal court, we evaluate the judgment(predicted fines \& sentences) with normalized logarithmic deviation. The average of the fine deviation and the sentence deviation is calculated as the final \metricid{III-4}.

\textbf{(3) Reasoning score} evaluates the quality of the legal agent's reasoning process based on the semantic alignment with the ground truth with the prompt in Figure \ref{fig:prompt of comparative evaluation chinese}.

\textbf{(4) Law score} evaluates the accuracy of cited legal provisions with exact matching.

\begin{table}[!t]
\centering
\small
\setlength{\tabcolsep}{2mm}
\resizebox{\linewidth}{!}{
\begin{tabular}{lllllll}
\hline
\multicolumn{1}{c|}{\textbf{Model}}              & \multicolumn{1}{c}{\textbf{KQ}}   & \multicolumn{1}{c}{\textbf{LC}}   & \multicolumn{1}{c}{\textbf{CD}}    & \multicolumn{1}{c}{\textbf{DD}}    & \multicolumn{1}{c}{\textbf{CI}}    & \multicolumn{1}{c}{\textbf{CR}}     \\ \hline
\rowcolor{gray!25}\multicolumn{7}{c}{\textit{Multilingual LLMs (close source)}}                                    \\
\multicolumn{1}{l|}{ GPT-4o}             & 5.9 & 4.9 & 6.7  & 5.2  & 37.6 & 15.3  \\
\multicolumn{1}{l|}{ Claude-3.7}         & 5.9 & 5.0 & 12.8 & 9.2  & 37.5 & 11.6  \\ \hline
\rowcolor{gray!25}\multicolumn{7}{c}{\textit{Multilingual LLMs (open source)}}                                     \\
\multicolumn{1}{l|}{ Deepseek-v3\space$_{\mathrm{671B}}$}   & 6.3 & 4.9 & 13.0 & 4.9  & 11.6 & 7.2   \\
\multicolumn{1}{l|}{ Deepseek-r1\space$_{\mathrm{671B}}$}   & 6.0 & 5.3 & 10.8 & 4.8  &  1.2  & 1.3 \\
\multicolumn{1}{l|}{ Llama3.3-it.\space$_{\mathrm{70B}}$} & 5.9 & 5.2 & 7.0  & 5.0  & 17.5  & 8.9 \\
\multicolumn{1}{l|}{ Qwen3\space$_{\mathrm{32B}}$}    & 5.9 & 4.9 & 9.1  & 7.6  & 33.3 & 21.0  \\
\multicolumn{1}{l|}{ Gemma3-it.\space$_{\mathrm{27B}}$}      & 5.9 & 5.1 & 7.7  & 6.1  & 31.5 & 11.4   \\
\multicolumn{1}{l|}{ Qwen3\space$_{\mathrm{14B}}$}    & 6.0 & 5.0 & 6.1  & 6.7  & 32.9  & 16.9   \\
\multicolumn{1}{l|}{ Gemma3-it.\space$_{\mathrm{12B}}$}      & 5.9 & 5.2 & 8.7  & 7.1  & 26.5 &  14.5  \\
\multicolumn{1}{l|}{ GLM4\space$_{\mathrm{9B}}$}            & 6.0 & 5.2 & 13.5 & 5.0  &  5.5 &  7.3  \\
\multicolumn{1}{l|}{ Qwen3\space$_{\mathrm{8B}}$}     & 6.1 & 4.8 & 14.3 & 14.1 & 32.3 & 13.1   \\
\multicolumn{1}{l|}{ InternLM3-It.\space$_{\mathrm{8B}}$} & 5.9 & 5.6 & 2.7  & 4.3  & 1.1  & 1.4   \\
\multicolumn{1}{l|}{ Ministral-it.\space$_{\mathrm{8B}}$} & 6.2 & 6.1 & 1.0  & 2.4  & 29.4  & 27.1   \\
\multicolumn{1}{l|}{ Qwen2.5-it.\space$_{\mathrm{7B}}$}   & 5.9 & 4.9 & 5.6  & 5.2  & 32.4 & 20.0 \\
\multicolumn{1}{l|}{ Qwen3\space$_{\mathrm{4B}}$}     & 6.0 & 5.2 & 8.1  & 8.2  & 24.2  & 15.5   \\ \hline
\rowcolor{gray!25}\multicolumn{7}{c}{\textit{Legal-specific LLMs}}                                                         \\
\multicolumn{1}{l|}{ LawLLM\space$_{\mathrm{13B}}$}         & 6.1 & 5.4 & 20.0 & 15.0 & 2.5 & 25.3  \\
\multicolumn{1}{l|}{ Chatlaw2\space$_{\mathrm{7B}}$}        & 7.2 & 7.9 & 6.7  & 6.3  & 21.8  & 15.2 \\ \hline
\end{tabular}}%
\caption{Average turn of interaction among LLM-driven legal agents on six environments.}
\label{tab:Average turn}
\end{table}

\begin{figure}[t]
    \centering
    \includegraphics[width=\linewidth]{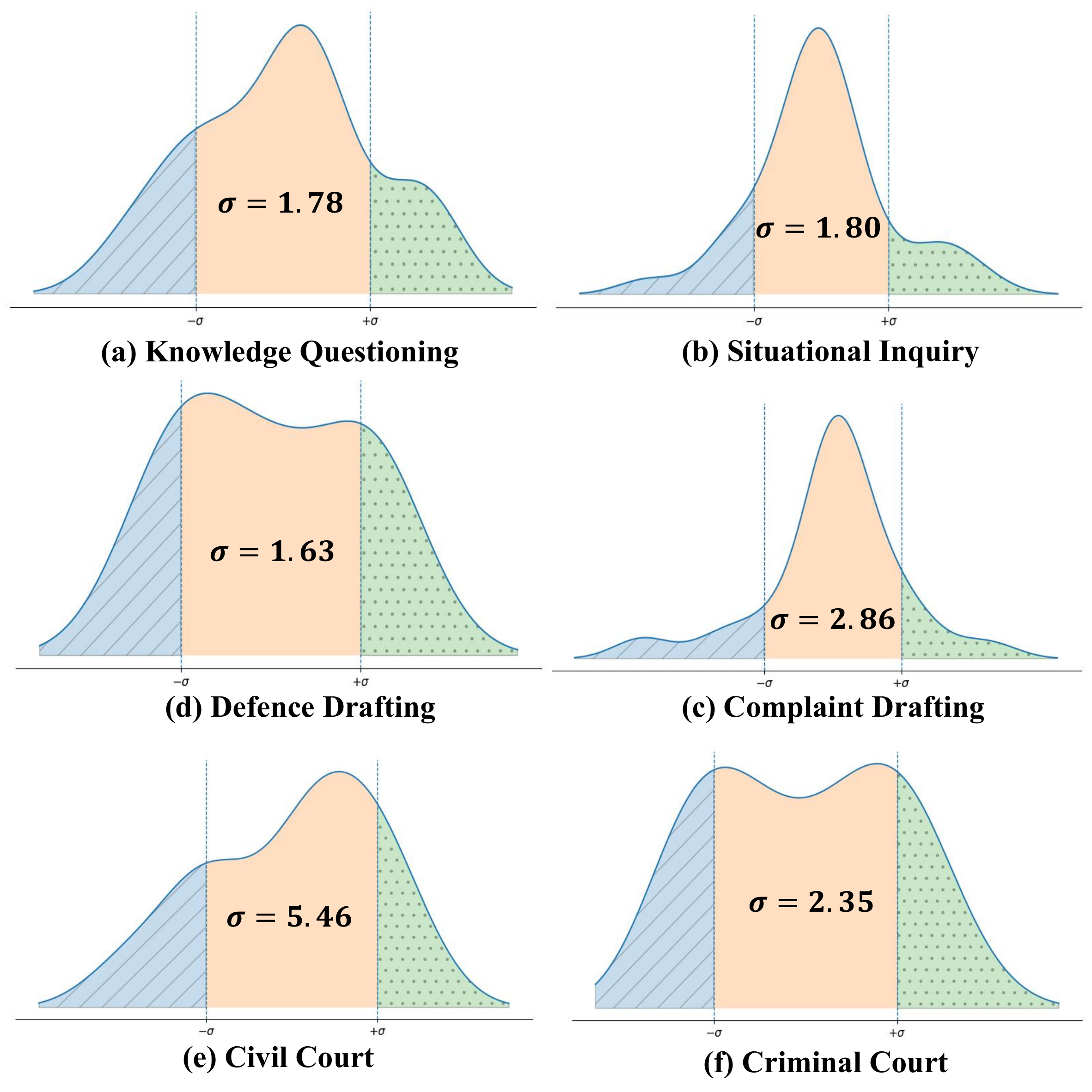}
    \caption{The distribution of the difference between GPT-4o scores and human evaluation across six scenarios}
    \label{fig:evaluation evaluation analysis}
\end{figure}

\begin{table*}[t]
 \setlength{\tabcolsep}{3mm}
\resizebox{\linewidth}{!}{
\begin{tabular}{l|ll|ll|ll|ll}
\hline
\multicolumn{1}{c|}{\multirow{2}{*}{Agent}} & \multicolumn{2}{c|}{KQ}                             & \multicolumn{2}{c|}{LC}                             & \multicolumn{2}{c|}{CD}                            & \multicolumn{2}{c}{DD}                                                                                      \\
\multicolumn{1}{c|}{}                       & \multicolumn{1}{c}{I-1} & \multicolumn{1}{c|}{I-2} & \multicolumn{1}{c}{I-1} & \multicolumn{1}{c|}{I-2} & \multicolumn{1}{c}{II-1} & \multicolumn{1}{c|}{II-2} & \multicolumn{1}{c}{II-1} & \multicolumn{1}{c}{II-2}  \\ \hline
 GPT-4o                                      & 65.7±1.9                   & 77.3±0.4                    & 48.5±1.3                   & 47.8±0.1                    & 59.4±0.2                   & 84.9±0.9                   & 82.1±1.1                   & 66.5±0.8                         \\
 Deepseek-v3\space$_{\mathrm{671B}}$                                 & 64.5±1.0 & 77.2±0.1                    & 51.7±0.9 & 48.0±0.1 & 98.2±0.7 & 81.7±1.0                   & 74.3±0.5 & 51.7±0.5         \\
 Qwen3\space$_{\mathrm{32B}}$                                  & 68.6±1.9 & 76.7±0.5 & 56.6±2.2 & 47.7±0.2                    & 85.3±0.9 & 87.2±0.2 & 89.5±1.3 & 57.1±0.8                          \\
 Qwen3\space$_{\mathrm{14B}}$                                   & 64.5±2.5 & 77.8±0.3 & 54.6±2.0 & 48.0±0.2                    & 71.9±1.3 & 61.1±0.4 & 71.9±1.3 & 61.1±0.4                      \\
 InternLM3-it.\space$_{\mathrm{8B}}$                                    & 63.0±0.2 & 76.8±0.1 & 51.4±1.1 & 45.6±0.1                    & 98.0±0.2 & 58.5±0.1 & 57.1±1.1 & 47.6±0.4                               \\ \hline
\end{tabular}}
\caption{Performance comparison of legal agents across multiple simulations(KQ, LC, CD, \& DD) in \textit{\textbf{J1-E}\scalebox{0.8}{\textbf{NVS}}} driven by Qwen3-32B. Scores are reported as average ± standard deviation.}
\label{tab:confidence interval kq lc cd dd}
\end{table*}

\begin{table*}[t]
 \setlength{\tabcolsep}{3mm}
\resizebox{\linewidth}{!}{
\begin{tabular}{l|llll|lllll}
\hline
\multicolumn{1}{c|}{\multirow{2}{*}{Agent}}            & \multicolumn{4}{c|}{CI}                                                                                & \multicolumn{5}{c}{CR}                                                                                        \\
\multicolumn{1}{c|}{}                     & \multicolumn{1}{c}{III-1} & \multicolumn{1}{c}{III-2} & \multicolumn{1}{c}{III-5} & \multicolumn{1}{c|}{III-6} & \multicolumn{1}{c}{III-1} & \multicolumn{1}{c}{III-3} & \multicolumn{1}{c}{\textcolor{red}{III-4}}   & \multicolumn{1}{c}{III-5} & \multicolumn{1}{c}{III-6} \\ \hline
 GPT-4o                                            & 80.9±1.3                   & 27.3±0.9                   & 47.2±1.7                   & 20.9±0.9                   & 21.1±0.9                   & 28.5±1.8                   & 74.2±2.8 & 24.1±1.4                   & 16.6±1.0                   \\
 Deepseek-v3\space$_{\mathrm{671B}}$                            & 21.7±1.0 & 10.8±0.8 & 13.7±1.5 & 8.0±1.8                   & 45.4±0.3 & 78.3±0.0 & 30.4±3.7 & 62.1±2.0 & 36.6±1.5                   \\
 Qwen3\space$_{\mathrm{32B}}$                & 45.9±0.2 & 13.1±1.4 & 29.3±0.9 & 16.2±0.5 &33.7±0.6 & 53.6±1.2 & 49.7±0.4 & 44.3±1.1 & 26.1±0.7                   \\
 Qwen3\space$_{\mathrm{14B}}$                        & 17.8±1.6 & 6.6±1.6 & 14.9±2.8 & 8.8±1.7                    & 23.2±1.0 & 41.1±0.8 & 67.0±2.7 & 31.3±1.3 & 21.9±0.8                   \\
 InternLM3-it.\space$_{\mathrm{8B}}$                             & 24.1±0.2 & 8.4±1.3 & 18.0±0.7 & 9.2±1.1 & 23.5±0.5 & 51.7±0.7 & 56.2±0.7 & 37.8±0.5 & 26.2±0.2                    \\ \hline
\end{tabular}}
\caption{Performance comparison of legal agents across multiple simulations(CI \& CR) in \textit{\textbf{J1-E}\scalebox{0.8}{\textbf{NVS}}} driven by Qwen3-32B. Scores are reported as average ± standard deviation.}
\label{tab:confidence interval ci cr}
\end{table*}

\section{Implementation Details.}
\label{sec:appendix: setting_details}
For adapted LLMs, we speed up inference using vllm \cite{kwon2023efficient}. Greedy decoding was used across the evaluations. 
For models smaller than 32B, we run evaluations using 8 * RTX 4090 GPUs with 24GB memory each. For models larger than 32B, we run evaluations using 8 * A100 GPUs with 80GB memory.
For some metrics calculating in the evaluation, we use GPT-4o \footnote{gpt-4o-2024-11-20}. To prevent infinite interaction loops caused by limited model capabilities, we set maximum interaction rounds for each scenario: 15 for knowledge 1
questioning, 10 for legal consultation, 20 for complaint drafting, 15 for defence drafting, 50 for civil court, and 35 for criminal court.

\section{Additional Experiment}
\label{sec:appendix:additional experiment}
\subsection{Additional overall performance}

We provide additional results of the overall performance among LLM-driven legal agents in {\textit{\textbf{J1-E}\scalebox{0.8}{\textbf{NVS}}}}.
\begin{itemize}

    \item \textbf{Complaint \& Defence Drafting.} Table \ref{tab:DOC in Complaint Drafting} and Table \ref{tab:Defence} report additional \metricid{II-2} results in Complaint Drafting and Defence Drafting. Both general and legal-specific models perform relatively well on five sub-metrics of \metricid{II-2}, indicating powerful information collection ability through the multi-turn conversation. However, agents driven by smaller or legal-specific models, such as LawLLM, encounter difficulty in accomplishing the assigned tasks, requiring further training.

    \item \textbf{Civil Court.} Table \ref{tab:Civil Court} reports additional \metricid{III-1} results in the Civil Court. Claude-3.7 and Gemma3-Instruct-27B demonstrate remarkable performance in \textbf{STA} and \textbf{ACT}, indicating strong procedural-following ability of these models, while legal-specific models, the inference model, and smaller models are unable to follow civil court procedures.

    \item \textbf{Criminal Court.} Table \ref{tab: Criminal Court} reports additional \metricid{III-1} and \metricid{II-2} results in the Criminal Court.Claude-3.7 score 95.7 in \textbf{UNI}. In contrast, legal-specific models fail to navigate the criminal court, resulting lower \textbf{UNI} and poorer performance.
\end{itemize}


\subsection{Analysis of interaction turn.}
Table \ref{tab:Average turn} presents the average number of interactions observed across the six environments.
Conversely, agents that fail to complete the task tend to have few turns or directly reach the maximum turn limit. For instance, Deepseek-r1, ChatGLM, InternLM3, and LawLLM achieve only 1.2, 5.5, 1.1,  and 2.5 turns on average at {C\scalebox{0.8}{I}}. 
The results indicate that weak interactive abilities prevent the models from sustaining focused, purposeful dialogues, often leading to off-topic, incomplete, or procedurally flawed interactions.

\subsection{Analysis of confidence intervals.}
To validate the robustness of our framework across repeated trials, we conduct simulations three times in environments driven by Qwen3-32B, and analyze the confidence intervals of outcomes.
As shown in Table \ref{tab:confidence interval kq lc cd dd} and Table \ref{tab:confidence interval ci cr}, scores across multiple simulations are generally stable, with only minor variations reflected in standard deviations. This indicates that our evaluation framework produces robust and consistent measurements of legal agent performance across repeated runs, validating the reliability of the metrics in \textit{\textbf{J1-E}\scalebox{0.8}{\textbf{NVS}}}.

\subsection{Definition of failed causes}
\label{sec:appendix:failed causes}
\begin{itemize}
    \item \textbf{Single-turn Collapse}. The judge agent completes all stages \& actions within a single turn, which is inconsistent with the purpose of dynamic simulation.

    \item \textbf{Missing Final Objective}. The judge agent fails to generate judgments at the end of the dialogue,

    \item \textbf{Missing Addressee}. The judge agent fails to explicitly contain addressee in an utterance.

    \item \textbf{Role Omission}. The judge agent overlooks one or more environmental roles during the simulation.
\end{itemize}

\section{Human Evaluation}
\label{sec:appendix:human evaluation}

\subsection{Implementation Details}
\begin{figure*}[t]
    \centering
\includegraphics[width=0.9\linewidth]{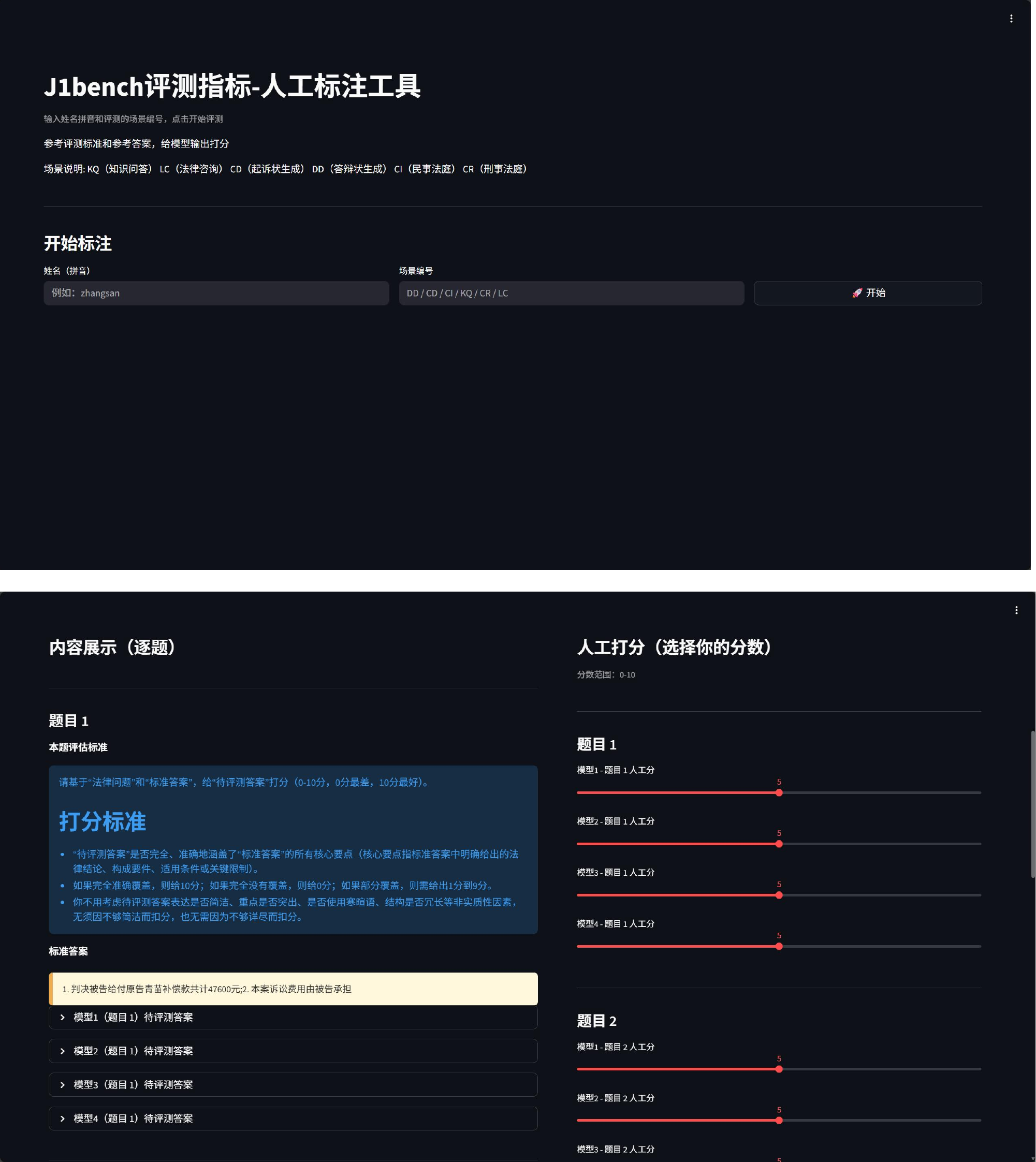}
    \caption{Screen shots of the human annotation system}
    \label{fig:annotation}
\end{figure*}

We invite a law professor to assist in recruiting seven evaluators. These evaluators have received systematic legal training and hold master’s or higher degrees in law. The human evaluation process is conducted with a dedicated annotation system shown in Fig. \ref{fig:annotation}. To reduce potential bias, all evaluators received identical task instructions and scoring guidelines prior to annotation and completed the evaluation independently. Model identities and experimental settings were concealed, and evaluators made judgments solely based on the content. We randomly sample 20 cases per scenario, and each evaluator independently scored all samples, resulting in a total of 840 human evaluation records. To mitigate the impact of outliers, we discard the maximum and minimum human ratings for each case and compute the average of the remaining scores as the final human evaluation score, which is compared against the GPT score.

\subsection{Analysis of behavioral consistency}
\begin{figure*}[t]
    \centering
\includegraphics[width=0.9\linewidth]{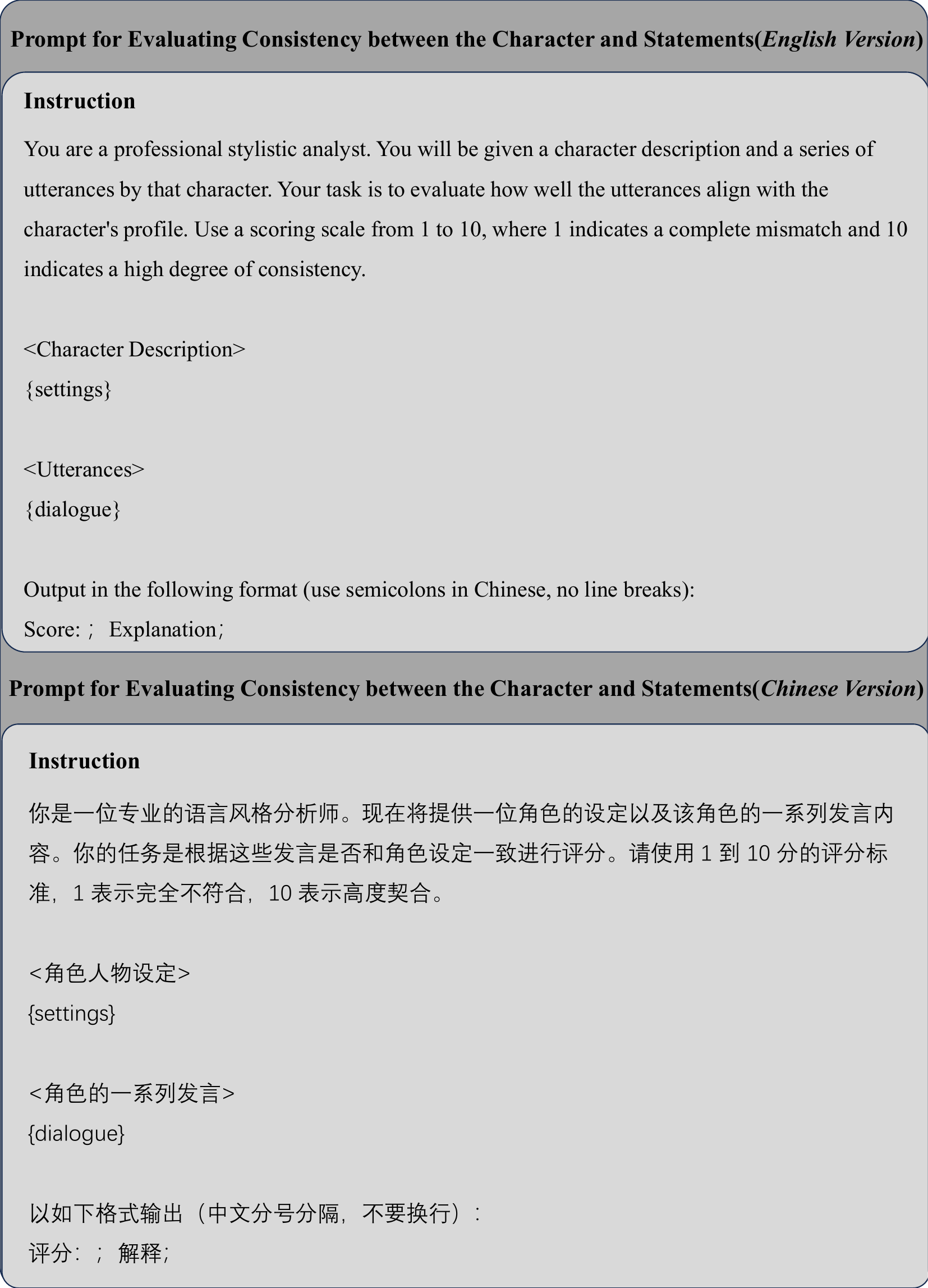}
    \caption{Prompt for evaluating consistency between the Character and Statements}
    \label{fig:prompt of consistency evaluation}
\end{figure*}

We perform a comprehensive analysis of the behavioral consistency of environment roles during interactions with three legal agents, leveraging GPT-4o and human evaluators to rate the alignment on a 10-point scale. GPT-4o assessment is performed across all samples with the prompt shown in Figure \ref{fig:prompt of consistency evaluation}, while human evaluators assess 20 samples per scenario. As shown in Table \ref{tab:Detailed GPT evaluation} and Table \ref{tab:Detailed Human Evaluation}, in interactions with various legal agents(GPT-4o, Deepseek-v3, and Qwen3-14B), the performance of all environmental roles remains consistently high and stable. The experiment demonstrates the reliability and effectiveness of environmental roles, establishing a solid basis for the subsequent assessment of the legal agent's performance.

\begin{table}[!t]
\centering
\setlength{\tabcolsep}{1mm}
\resizebox{\linewidth}{!}{
\begin{tabular}{lcccc}
\hline
\multicolumn{1}{c|}{\multirow{2}{*}{Agent}} & \multicolumn{1}{c}{KQ}     & \multicolumn{1}{c}{LC}     & \multicolumn{1}{c}{CD}        & \multicolumn{1}{c}{DD}                                                                         \\
\multicolumn{1}{c|}{}                       & \multicolumn{1}{c}{General Public} & \multicolumn{1}{c}{G{eneral Public}} & \multicolumn{1}{c}{Plaintiff} & \multicolumn{1}{c}{Defendant}  \\ \hline
\rowcolor{gray!25}\multicolumn{5}{c}{Env.GPT-4o}                                                                                                                                                                                                                                                                                                     \\
\multicolumn{1}{l|}{ GPT-4o}                 & 86.8                      & 80.5                      & 90.1                         & 86.7                                                \\
\multicolumn{1}{l|}{ Deepseek-v3\space$_{\mathrm{671B}}$}            & 87.6                      & 85.7                      & 66.2                         & 82.0                                 \\
\multicolumn{1}{l|}{ Qwen3\space$_{\mathrm{32B}}$}              & 85.1                      & 81.5                      & 85.2                         & 85.2                              \\ \hline
\rowcolor{gray!25}\multicolumn{5}{c}{Env.Qwen3\space$_{\mathrm{32B}}$}                                                                                                                                                                                                                                                                                                  \\
\multicolumn{1}{l|}{ GPT-4o}                 & 92.6                      & 90.3                      & 92.8                         & 89.1                                             \\
\multicolumn{1}{l|}{ Deepseek-v3\space$_{\mathrm{671B}}$}            & 94.4                      & 89.4                      & 86.2                         & 88.9                           \\
\multicolumn{1}{l|}{ Qwen3\space$_{\mathrm{32B}}$}              & 92.9                      & 90.8                      & 89.4                         & 89.4                              \\ \hline
\end{tabular}}
\caption{Behavior consistency of various environment roles under GPT evaluation.}
\label{tab:Detailed GPT evaluation}
\end{table}

\subsection{Analysis of differences between human evaluation and LLM-as-Judge}
\label{sec:Analysis of differences between human evaluation and LLM-as-Judge}
We analyze the discrepancy between human scores and GPT-4o scores. As shown in Figure \ref{fig:evaluation evaluation analysis}, the standard deviations in five scenarios (\textit{i.e.,} Knowledge Questioning, Consultation, Complaint Drafting, Defence Drafting, Criminal Court) are all below 3.0, while the civil court scenario exhibits a higher standard deviation of 5.46, resulting in higher the overall difference. This larger variance is due to the civil court scenario containing more evaluation metrics and longer textual content, which increases the likelihood of divergence between human judgments and GPT-4o’s assessments.

\subsection{Ethics Statement}
This work involves human annotators who manually score the outputs of legal agents. We carefully considered ethical issues related to human participation, data usage, and potential harm.
All annotators were recruited voluntarily and were informed of the purpose of the study, the nature of the annotation tasks, and the expected time commitment in advance. They participated with informed consent and were allowed to withdraw at any time. The annotation tasks involved evaluating model-generated legal outputs and did not require annotators to provide personal information or disclose sensitive data.
Annotators were fairly compensated for their work in accordance with local standards. No vulnerable populations were involved in the study.

\begin{table}[!t]
\centering
\setlength{\tabcolsep}{1.0mm}
\resizebox{\linewidth}{!}{
\begin{tabular}{lcccc}
\hline
\multicolumn{1}{c|}{\multirow{2}{*}{Agent}}   & \multicolumn{1}{c}{KQ}     & \multicolumn{1}{c}{LC}  & \multicolumn{1}{c}{CD}        & \multicolumn{1}{c}{DD}                                                                      \\
 \multicolumn{1}{c|}{}  & \multicolumn{1}{c}{General Public} & \multicolumn{1}{c}{General Public}     & \multicolumn{1}{c}{Plaintiff} & \multicolumn{1}{c}{Defendant}  \\ \hline
\rowcolor{gray!25}\multicolumn{5}{c}{Env.GPT-4o}                                                                                                                                                                                                                                                                                      \\
\multicolumn{1}{l|}{ GPT-4o}       & 77.0                      &    73.0                             & 78.5                                  & 76.5                               \\
\multicolumn{1}{l|}{ Deepseek-v3\space$_{\mathrm{671B}}$}      & 77.0                      & 73.0                & 72.5                         & 75.0                                   \\
\multicolumn{1}{l|}{ Qwen3\space$_{\mathrm{32B}}$}      & 75.0                      & 72.0                & 73.5                         & 75.5                                                 \\ \hline
\rowcolor{gray!25}\multicolumn{5}{c}{Env.Qwen3\space$_{\mathrm{32B}}$}                                                                       \\
\multicolumn{1}{l|}{ GPT-4o}     & 76.0                      & 75.0              & 74.0                         & 77.0                                           \\
\multicolumn{1}{l|}{ Deepseek-v3\space$_{\mathrm{671B}}$}   & 75.6                      & 75.0      & 74.5                         & 79.0                                                         \\
\multicolumn{1}{l|}{ Qwen3\space$_{\mathrm{32B}}$}     & 76.5                      & 77.0     & 72.0                         & 71.5                                                               \\ \hline
\end{tabular}}
\caption{Behavior consistency of various environment roles under human evaluation.}
\label{tab:Detailed Human Evaluation}
\end{table}

\begin{table*}[!t]
\setlength{\tabcolsep}{2mm}
\small
\centering
\resizebox{0.9\textwidth}{!}{
\begin{tabular}{lll}
\hline
\multicolumn{1}{c}{Prompt} & \multicolumn{1}{c}{Environment} & \multicolumn{1}{c}{Function} \\ \hline
Figure \ref{fig:prompt of client in knowledge questioning english} \& \ref{fig:prompt of client in knowledge questioning chinese}      &  KQ    &    Lay Person Prompt in Knowledge Questioning      \\
Figure \ref{fig:prompt of client in legal consultation english} \& \ref{fig:prompt of client in legal consultation chinese}      &  LC      &    Lay Person Prompt in Consultation      \\
Figure \ref{fig:prompt of client in complaint drafting english} \& \ref{fig:prompt of client in complaint drafting chinese}      &  CD     &    Lay Person Prompt in Complaint Drafting      \\
Figure \ref{fig:prompt of client in defence drafting english} \& \ref{fig:prompt of client in defence drafting chinese}      &  DD      &  Lay Person Prompt in Defence Drafting        \\ 
Figure \ref{fig:prompt of plaintiff in civil court english}, \ref{fig:prompt of plaintiff in civil court chinese}, \ref{fig:prompt of defendant in civil court english} and \ref{fig:prompt of defendant in civil court chinese} & CI & Plaintiff's \& Defendant's Lawyers Prompt in Civil Prediction \\ 
Figure \ref{fig:prompt of lawyer in criminal court english}, \ref{fig:prompt of lawyer in criminal court chinese}, \ref{fig:prompt of procurator in criminal court english}, \ref{fig:prompt of procurator in criminal court chinese}, and \ref{fig:prompt of defendant in criminal court} & CR & Character Prompt in Criminal Court \\
Figure \ref{fig:prompt of consistency evaluation} & KQ, LC, CD, \& DD & GPT Prompt for Consistency Evaluation \\
Figure \ref{fig:prompt of open-ended topics evaluation in chinese} & KQ \& LC & GPT Prompt for Open-ended Topics Evaluation \\
Figure \ref{fig:prompt of comparative evaluation chinese} & CD, DD, CI, \& CR & GPT Prompt for Comparative Evaluation \\

\hline
\end{tabular}}
\caption{Prompts and functions.}
\label{tab:prompts}
\end{table*}

\section{Case Study}
To present the performance of legal agents, we conduct a qualitative study in J1-Envs and J1-Eval, with GPT-4o acting as the legal agent.

\begin{figure*}[t]
    \centering
\includegraphics[width=\linewidth]{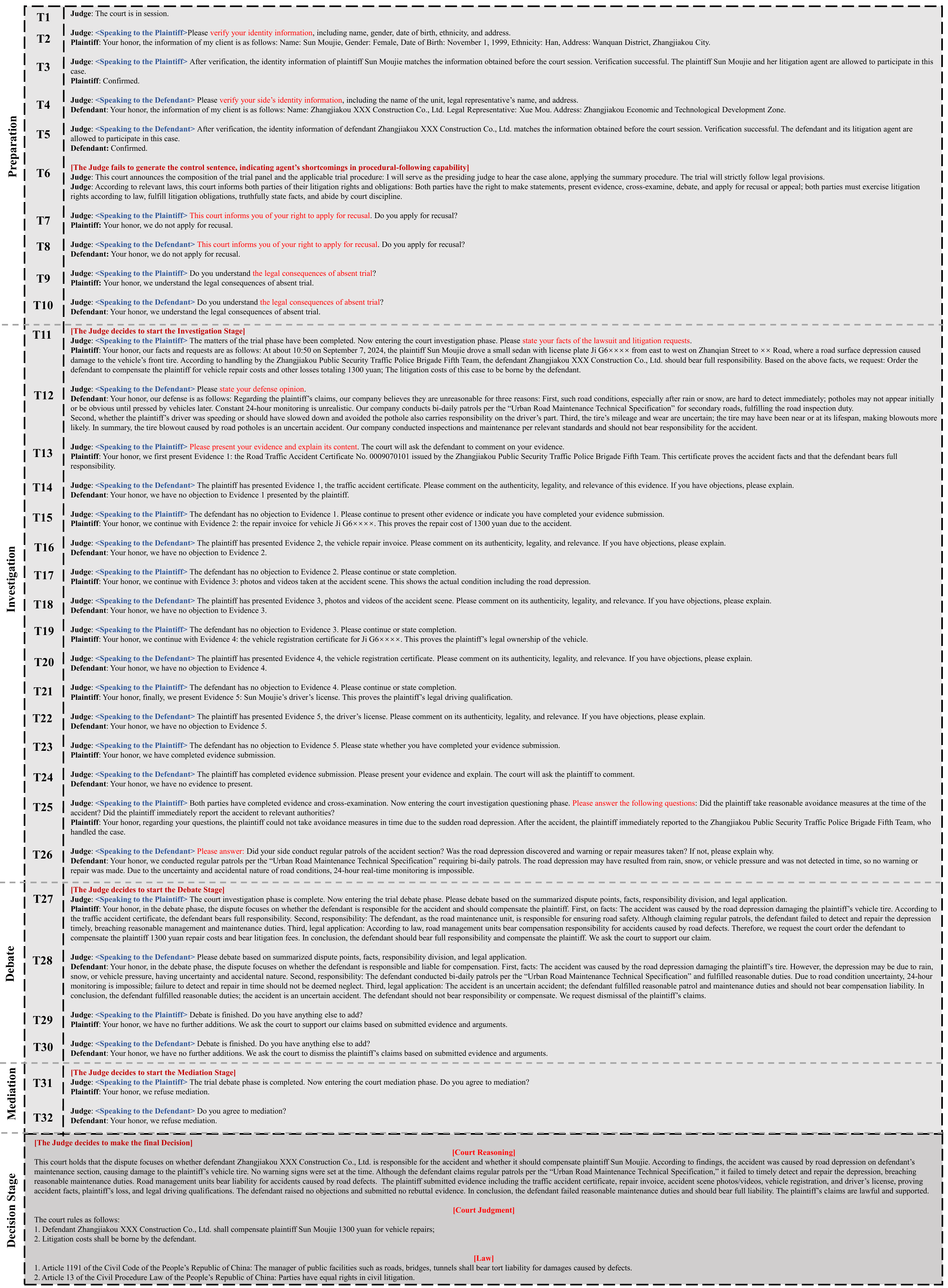}
    \caption{Qualitative result of GPT-4o as the judge in Civil Court. T$i$ denotes the $i$-th interaction turn. Blue denotes the control sentence utilized by the Judge to denote the target, while red denotes preset procedures}
    \label{fig:case_study_full_dialog_CI}
\end{figure*}

\begin{figure*}
    \centering
    \includegraphics[width=\linewidth]{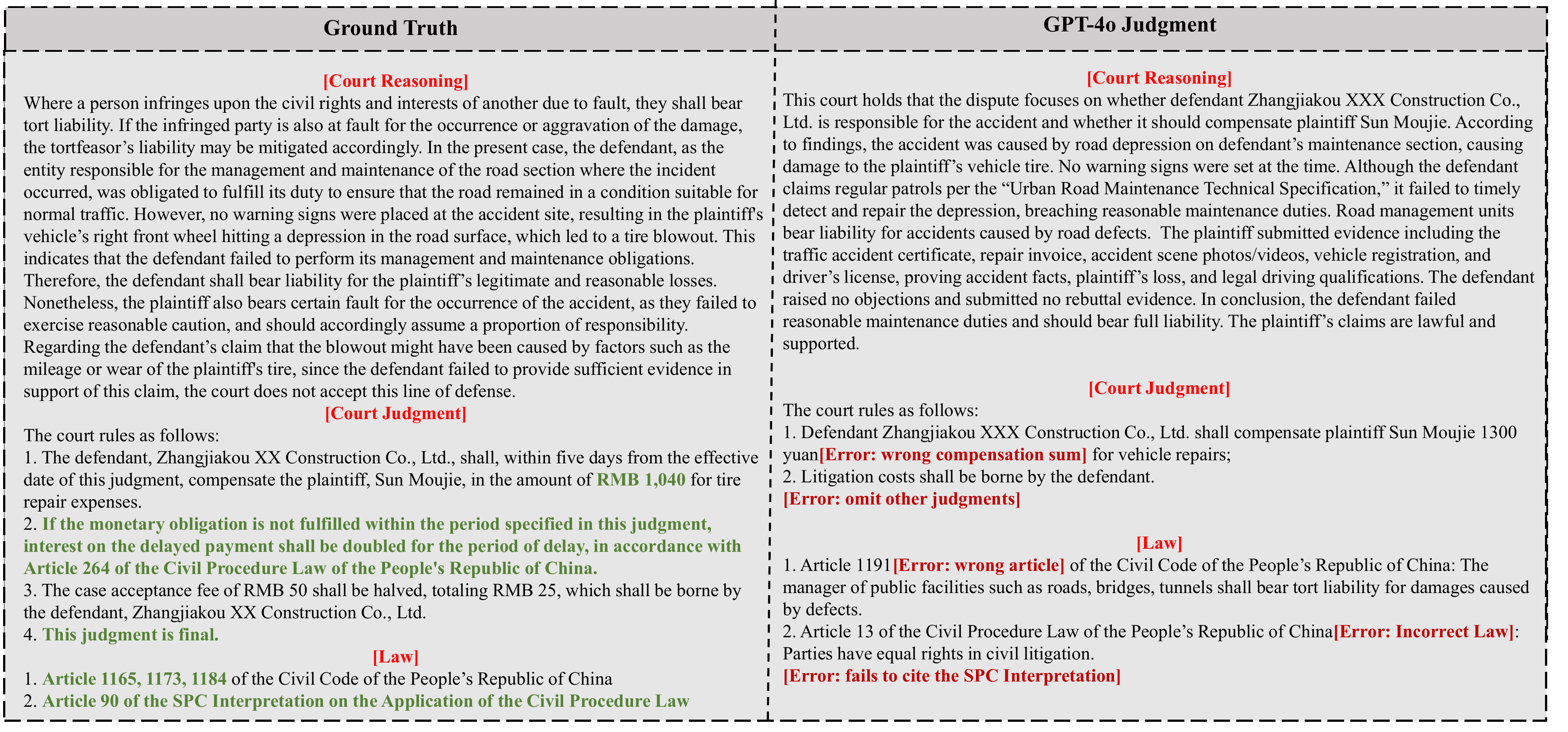}
    \caption{Qualitative result of GPT-4o judgment in comparison with the ground truth, where errors are highlighted in red, and corresponding correct answers are highlighted in green}
    \label{fig:case_study_eval_CI}
\end{figure*}

\begin{figure*}
    \centering
    \includegraphics[width=\linewidth]{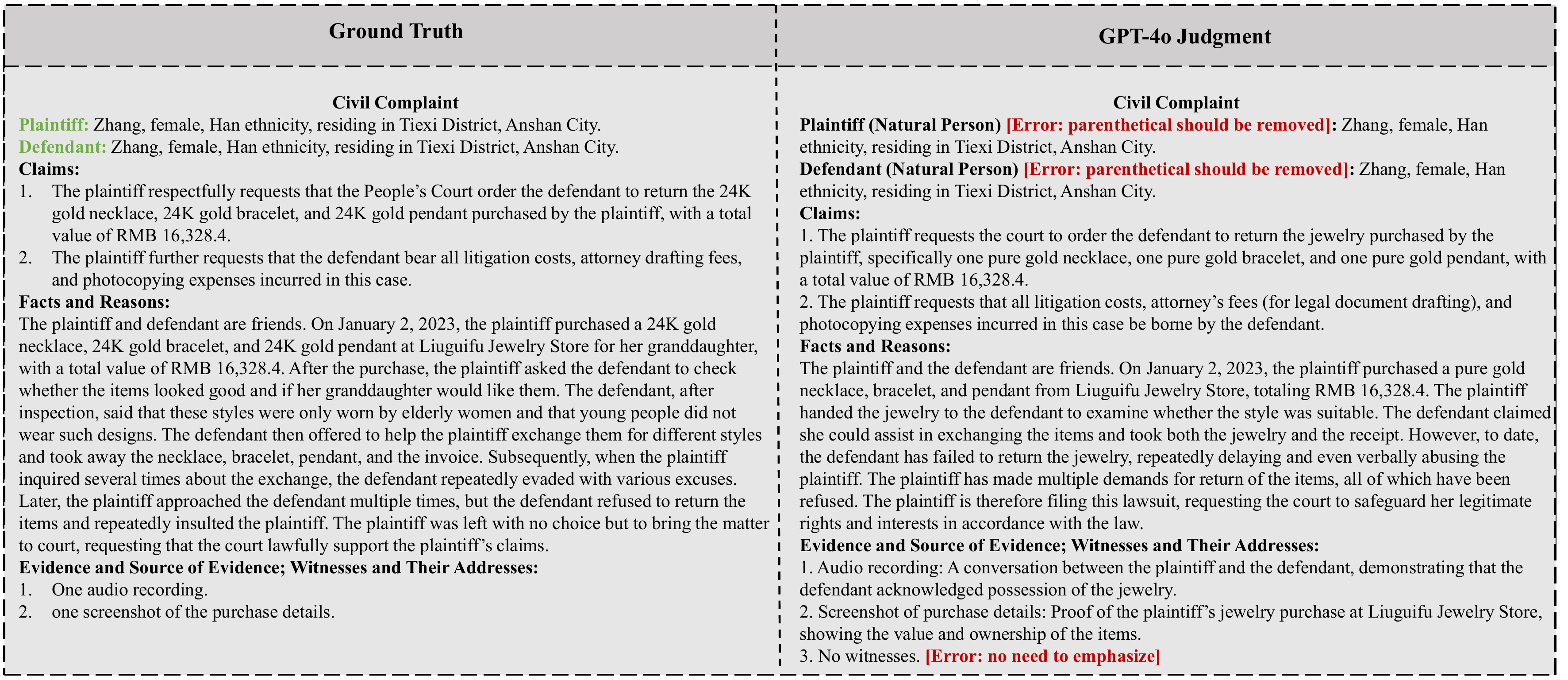}
    \caption{Qualitative result of GPT-4o drafted complaint in CD, where highlighted bold red boxes denote format errors, and green denotes corresponding correct answers}
    \label{fig:case_study_eval_CD}
\end{figure*}

\begin{itemize}
    \item \textbf{An example of CI}.  As illustrated in Figure \ref{fig:case_study_full_dialog_CI}, GPT-4o utilizes the control sentence(highlighted in blue) to designate the target of each dialogue turn while executing procedural steps (highlighted in red). After 32 turns, the legal agent concludes the court session and deliver the final decision. Despite having accomplished most stages and actions, GPT-4o occasionally fails to generate a proper control sentence(T6), resulting in a dialogue loop. The issue highlights the need for enhanced multi-agent coordination and collaboration capabilities.

    \item \textbf{Evaluation of the model judgment}. We further evaluate the quality of the model judgment in Figure \ref{fig:case_study_full_dialog_CI}. As shown in Figure \ref{fig:case_study_eval_CI}, compared to the ground truth, GPT-4o cites only basic provisions of the Civil Code while omitting relevant interpretive clauses issued by the Supreme Court, which offer detailed guidance, leading to misjudgments.

    \item \textbf{Evaluation of the model-drafted complaint}. As shown in Figure \ref{fig:case_study_eval_CD}, GPT-4o often rigidly reproduces format labels without proper modification. In 31.18\% of Complaint Drafting cases, GPT-4o preserve the content within parenthese. In 6.45\% of cases, GPT-4o incorrectly split the format label "Evidence and Source of Evidence; Witness and Their Addresses" into two distinct components.
\end{itemize}

\section{Prompts}

\label{sec:appendix:prompts}
We list prompts used in different environments in Table \ref{tab:prompts}. In each prompt, {xx} needs to fill with corresponding external inputs. We meticulously design prompts for data construction and agent settings to ensure clarity and functionality. These prompts are adaptable to most LLMs, enabling the reliable configuration of environment characters.

\begin{figure*}[t]
    \centering
\includegraphics[width=0.9\linewidth]{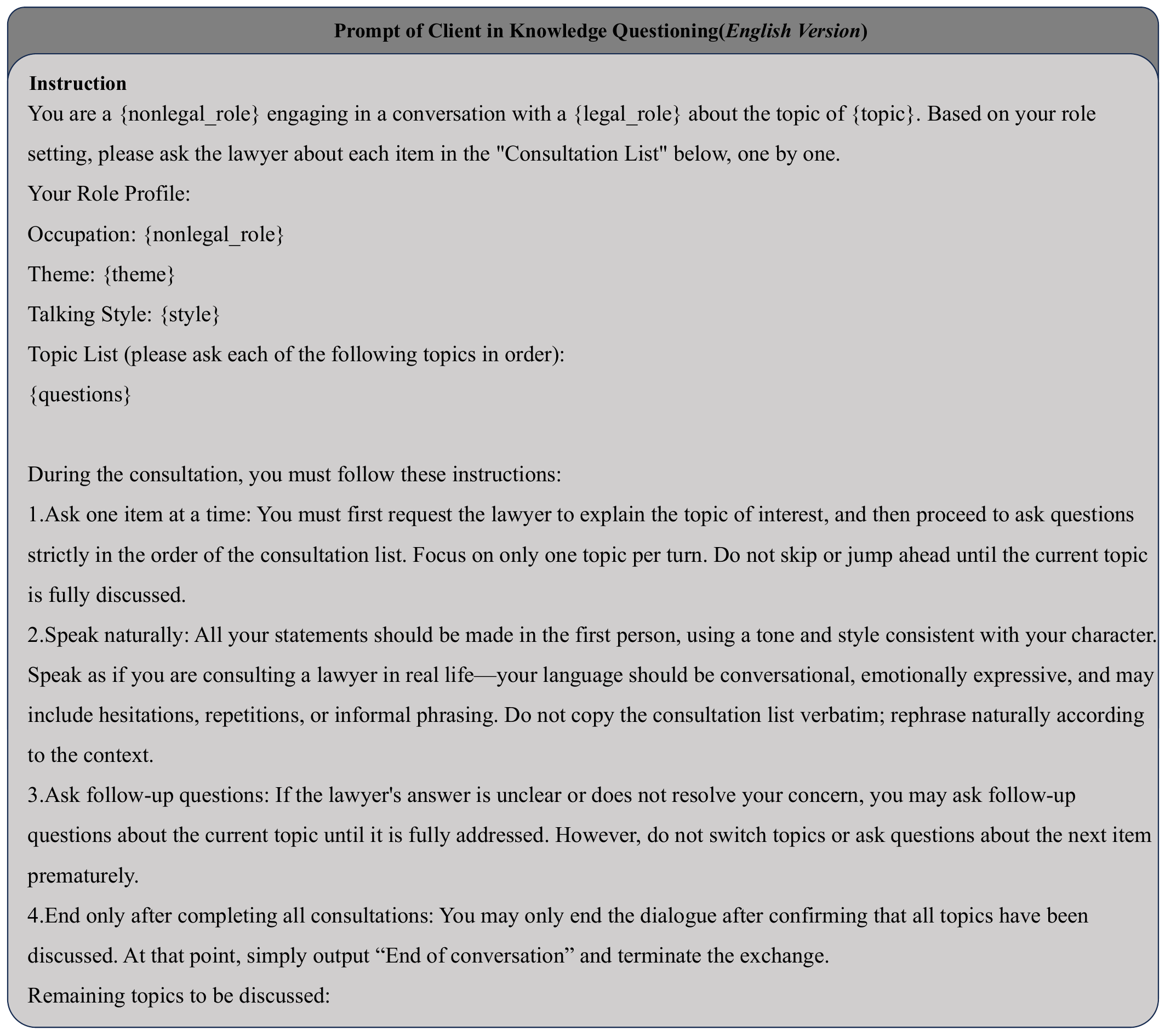}
    \caption{Prompt of Client in Knowledge Questioning(\textit{English Version})}
    \label{fig:prompt of client in knowledge questioning english}
\end{figure*}

\begin{figure*}[t]
    \centering
\includegraphics[width=0.9\linewidth]{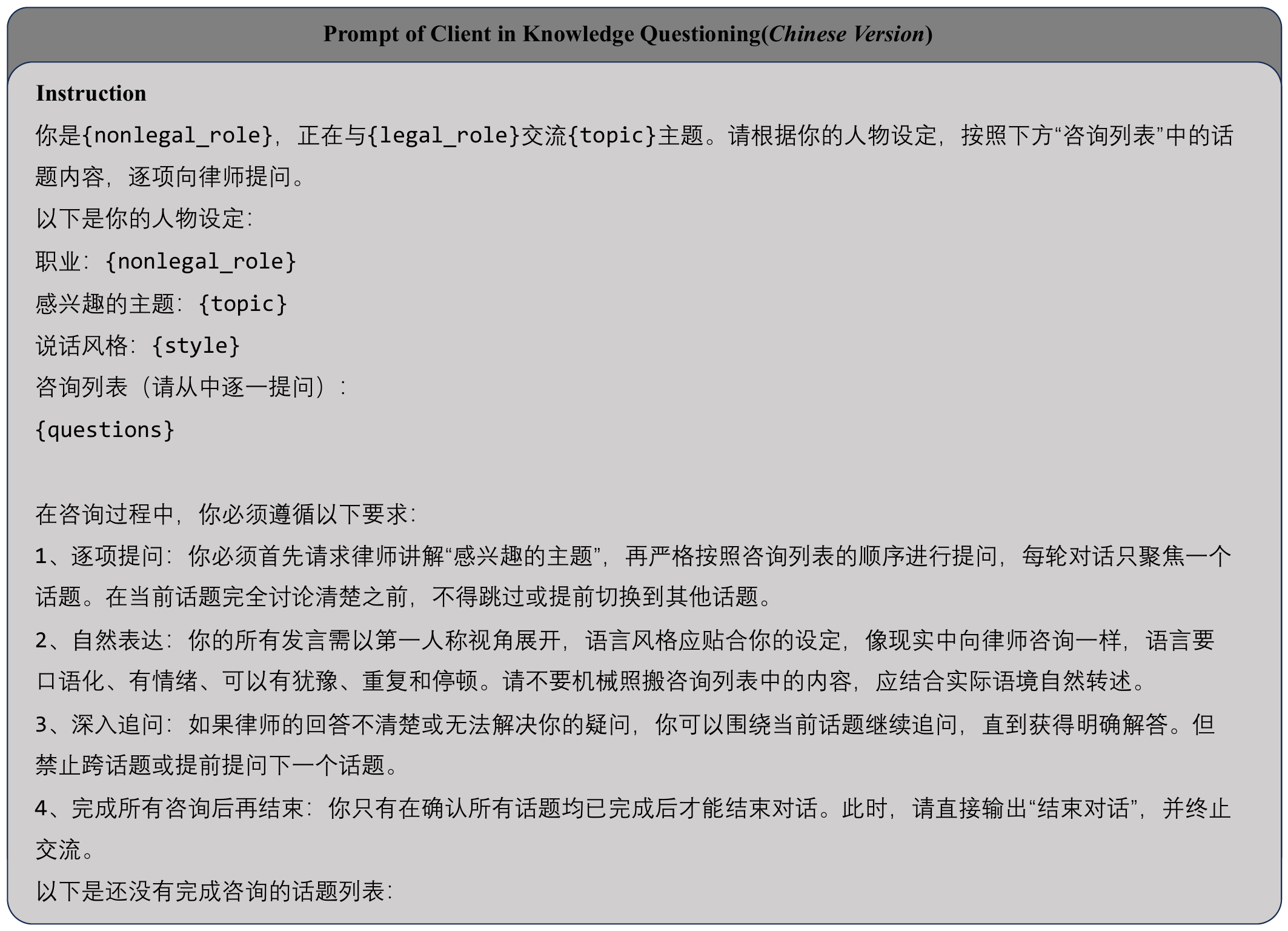}
    \caption{Prompt of Client in Knowledge Questioning(\textit{Chinese Version})}
    \label{fig:prompt of client in knowledge questioning chinese}
\end{figure*}

\begin{figure*}[t]
    \centering
\includegraphics[width=0.9\linewidth]{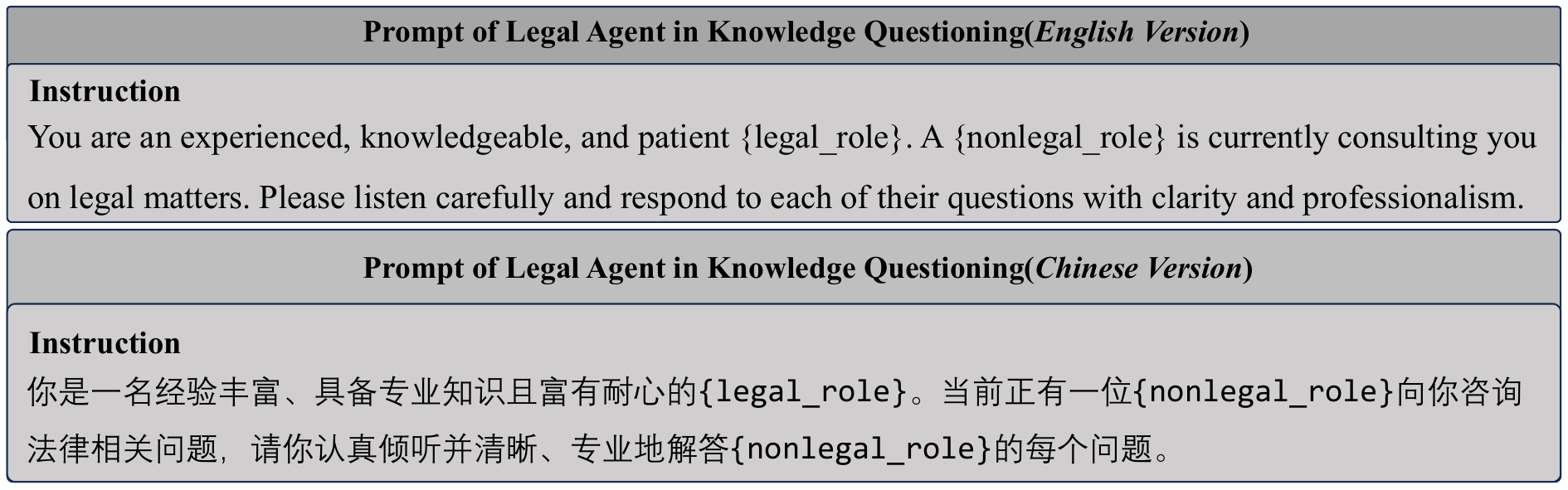}
    \caption{Prompt of Trainee in Knowledge Questioning}
    \label{fig:prompt of trainee in knowledge questioning}
\end{figure*}

\begin{figure*}[t]
    \centering
\includegraphics[width=0.9\linewidth]{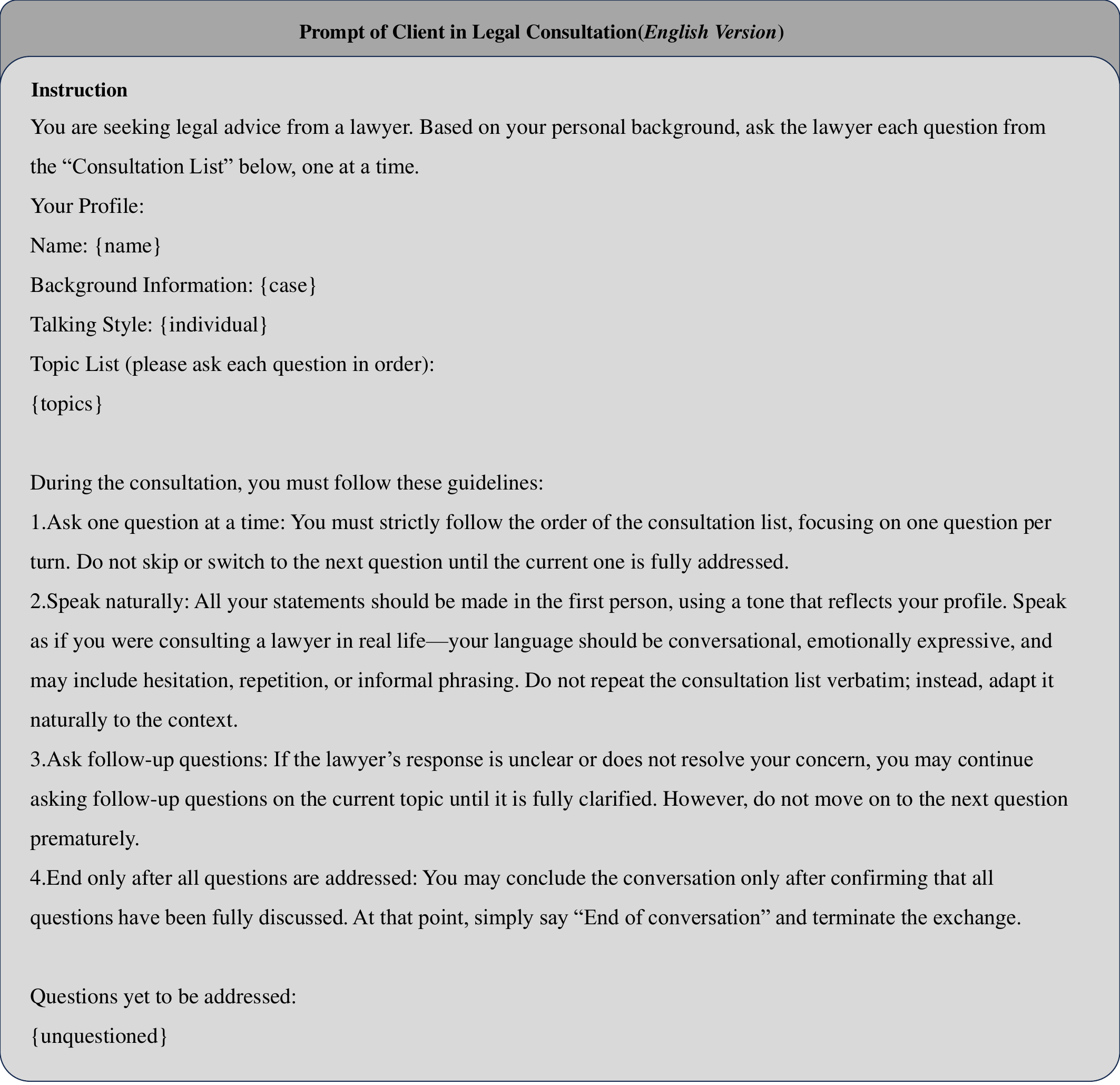}
    \caption{Prompt of Client in Legal Consultation(\textit{English Version})}
    \label{fig:prompt of client in legal consultation english}
\end{figure*}

\begin{figure*}[t]
    \centering
\includegraphics[width=0.9\linewidth]{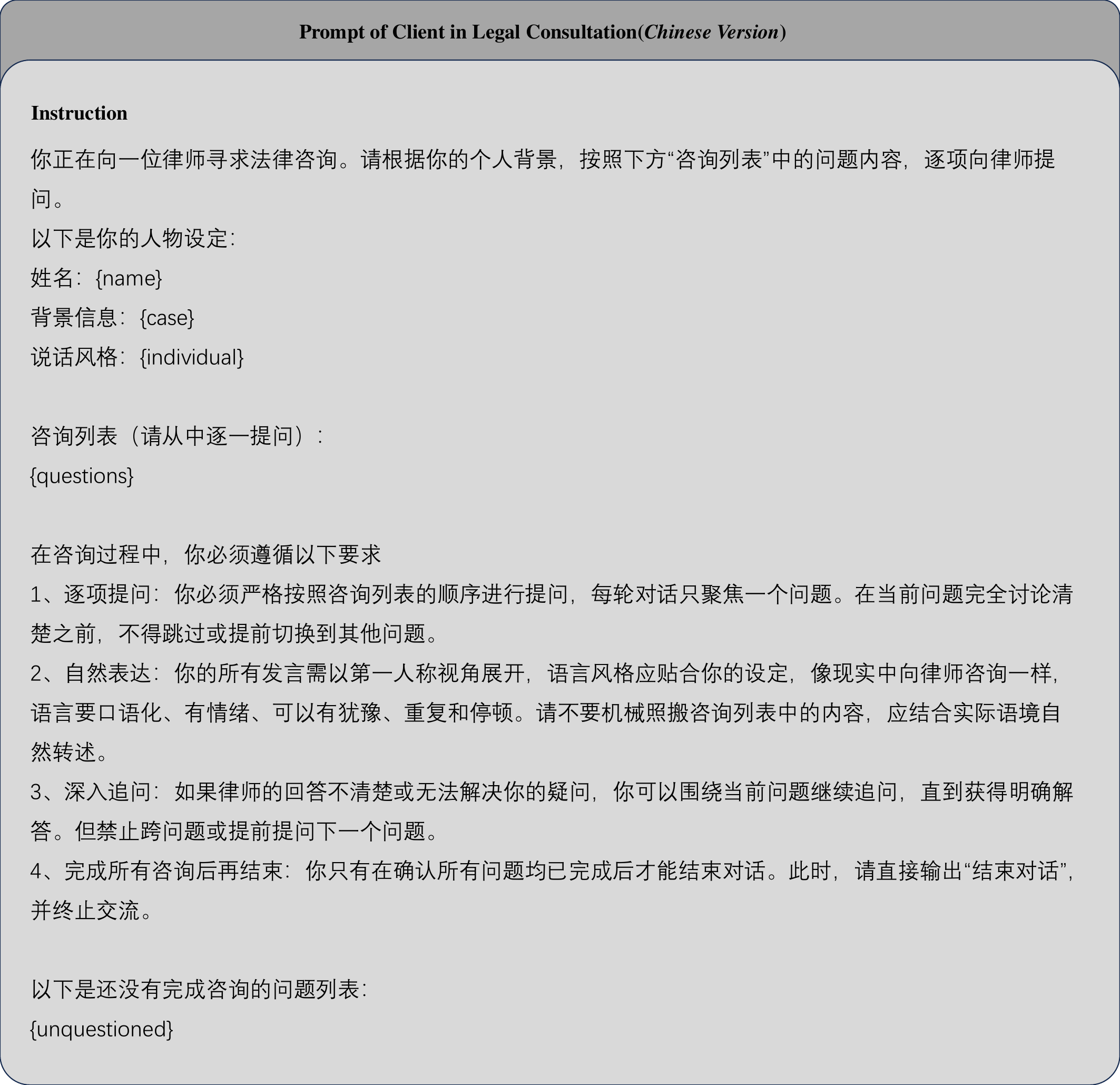}
    \caption{Prompt of Client in Legal Consultation(\textit{Chinese Version})}
    \label{fig:prompt of client in legal consultation chinese}
\end{figure*}

\begin{figure*}[t]
    \centering
\includegraphics[width=0.9\linewidth]{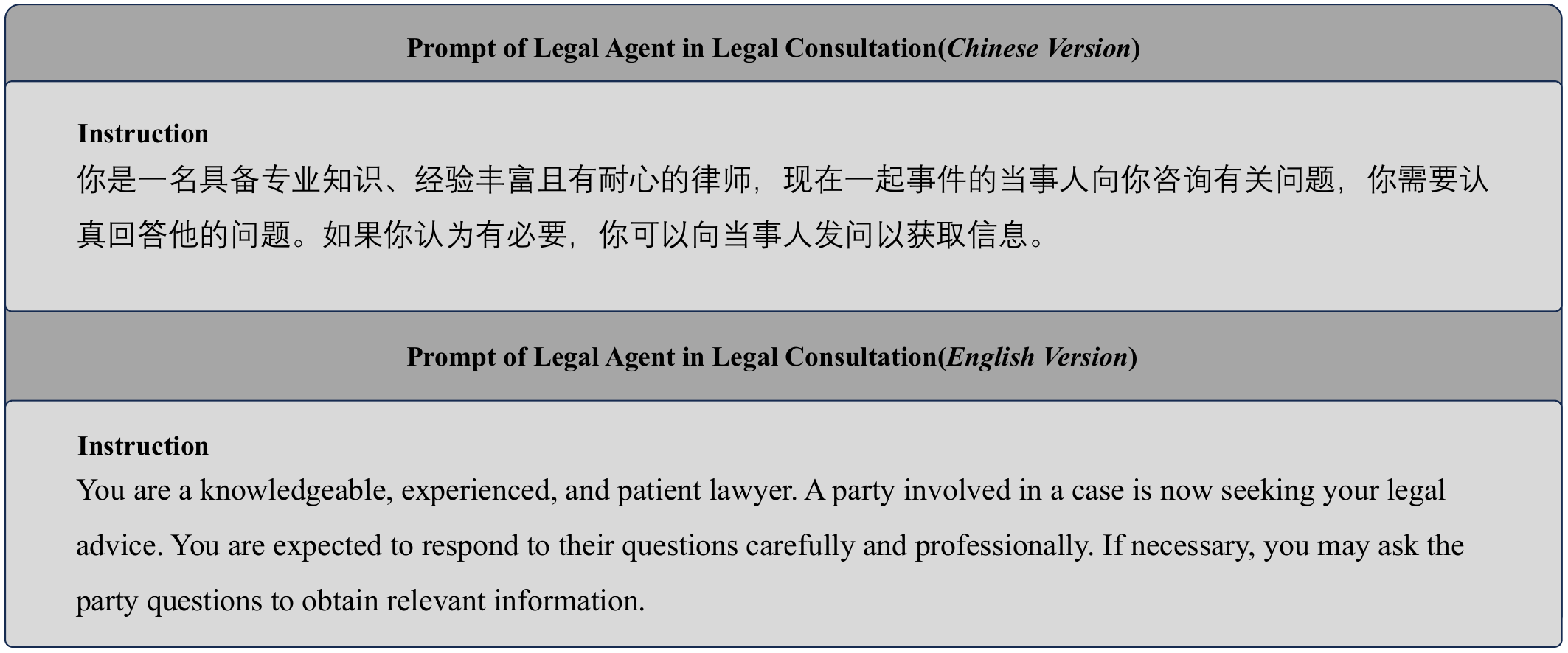}
    \caption{Prompt of Trainee in Legal Consultation}
    \label{fig:prompt of lawyer in legal consultation}
\end{figure*}

\begin{figure*}[t]
    \centering
\includegraphics[width=0.9\linewidth]{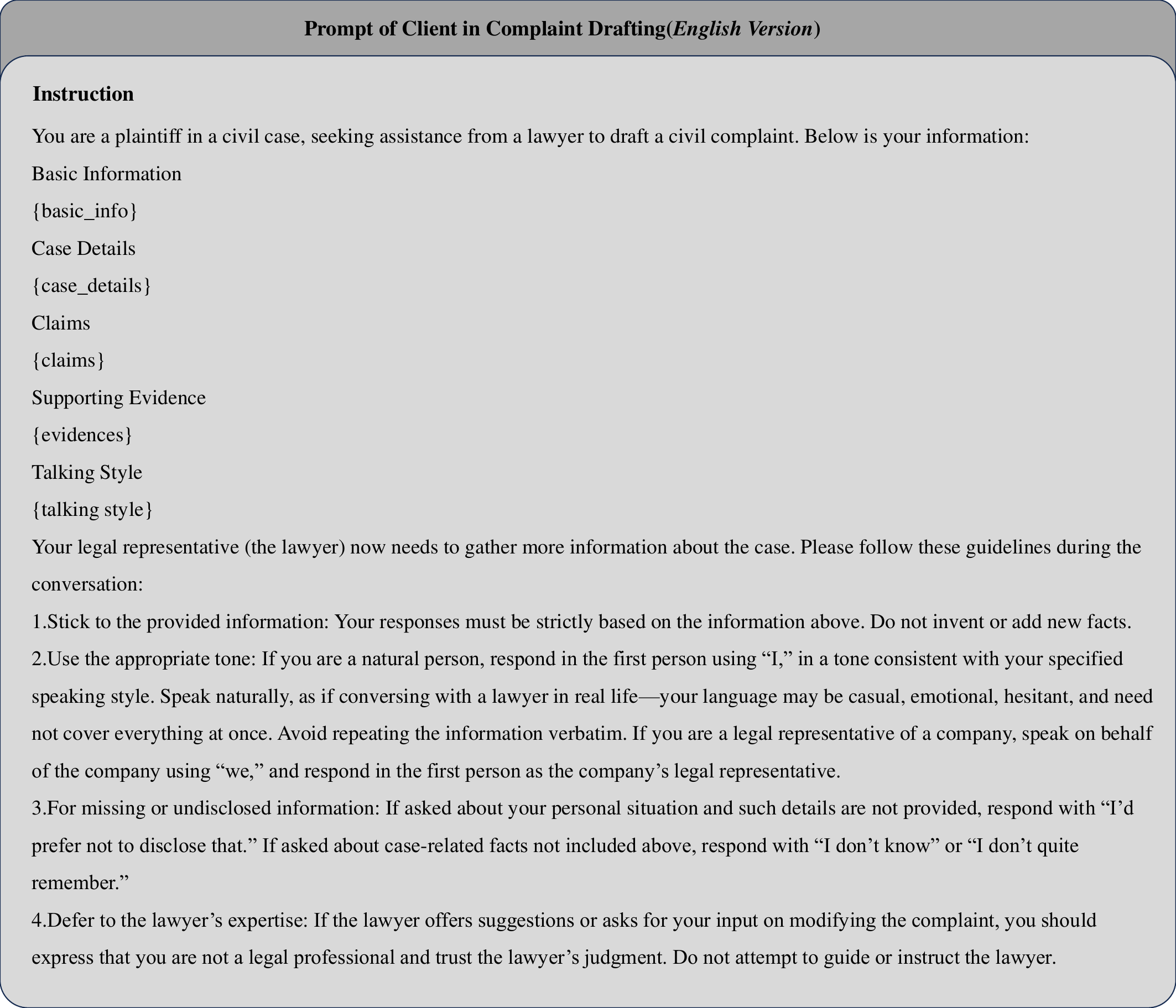}
    \caption{Prompt of Client in Complaint Drafting(\textit{English Version})}
    \label{fig:prompt of client in complaint drafting english}
\end{figure*}

\begin{figure*}[t]
    \centering
\includegraphics[width=0.9\linewidth]{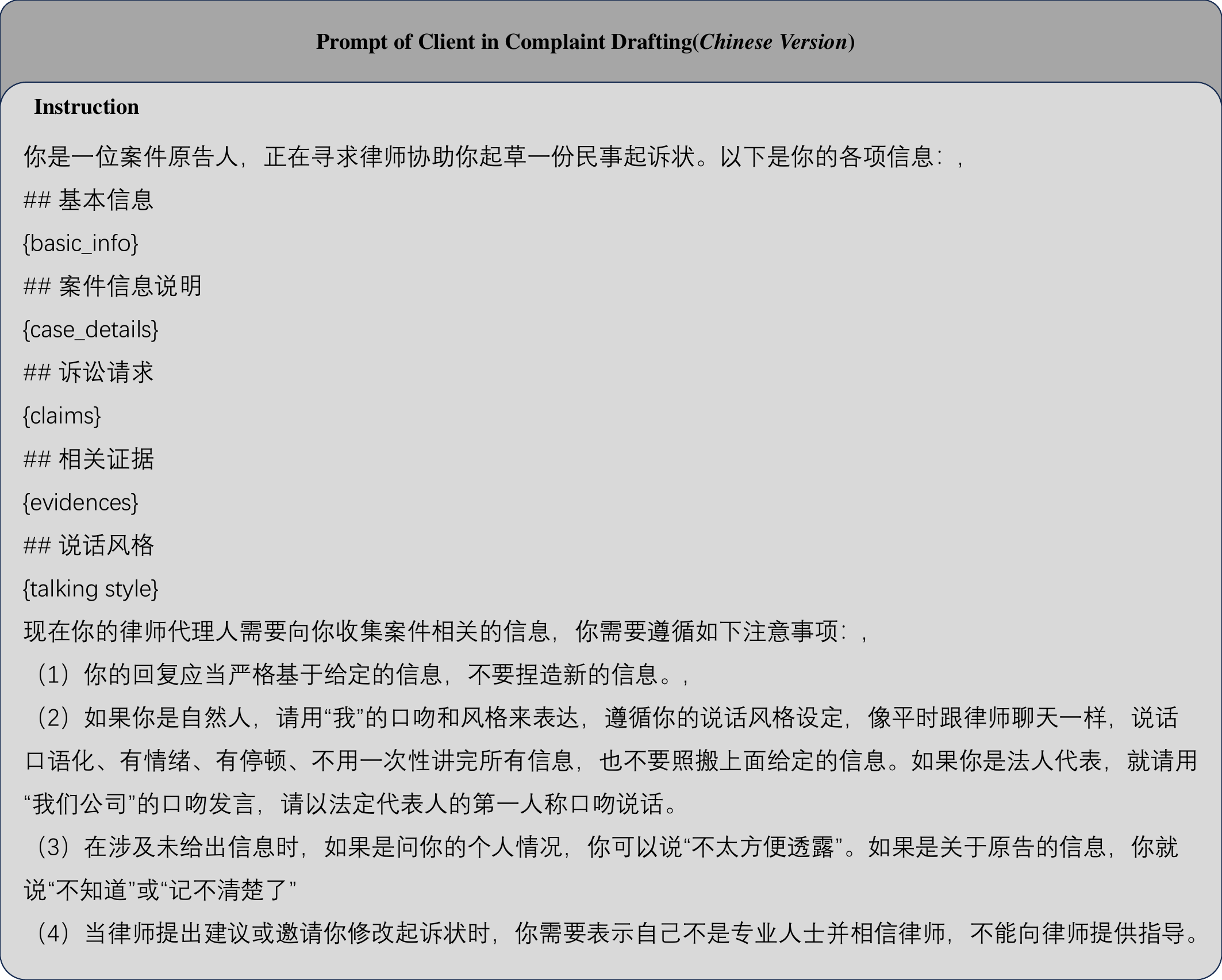}
    \caption{Prompt of Client in Complaint Drafting(\textit{Chinese Version})}
    \label{fig:prompt of client in complaint drafting chinese}
\end{figure*}

\begin{figure*}[t]
    \centering
\includegraphics[width=0.9\linewidth]{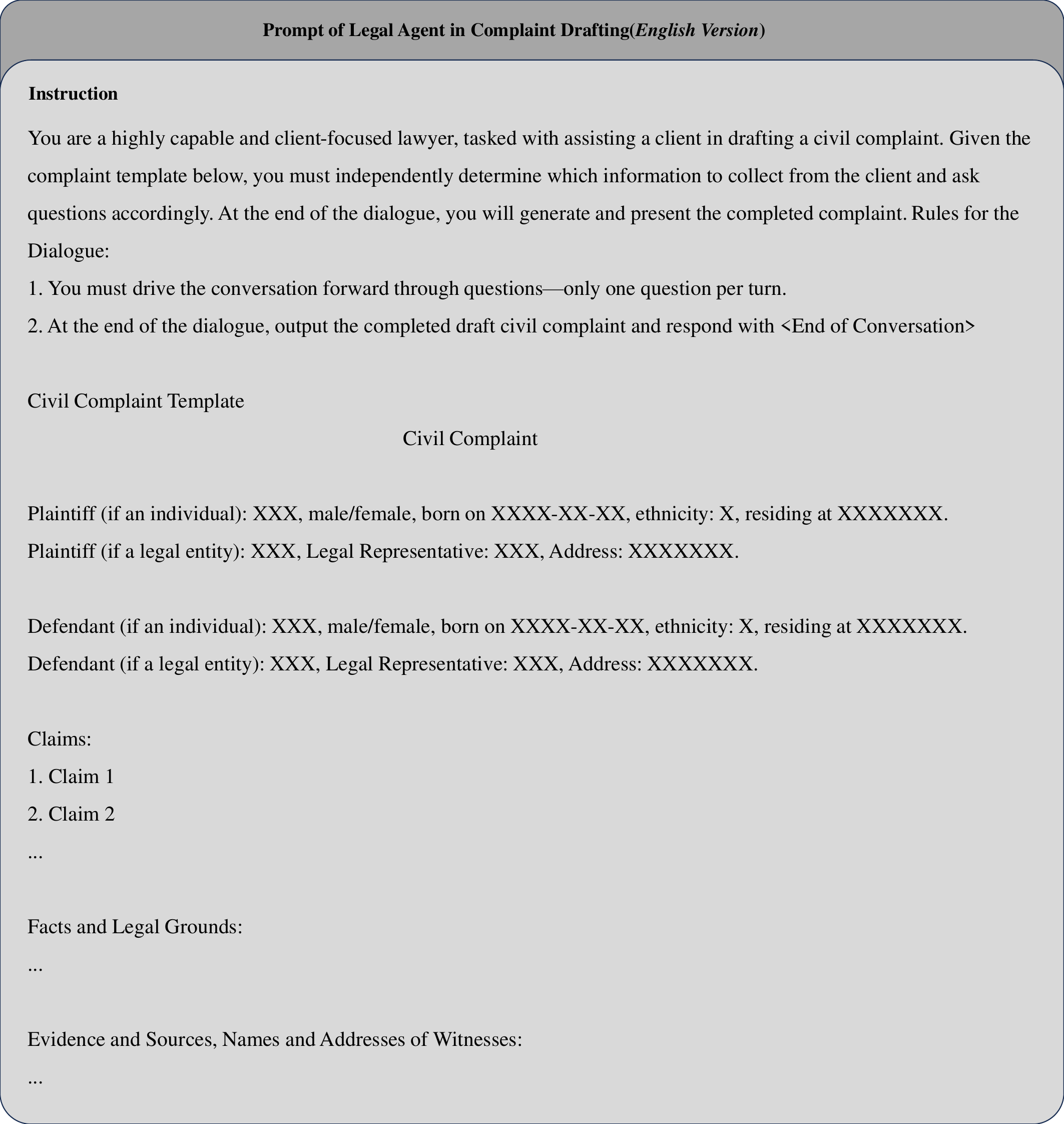}
    \caption{Prompt of Lawyer in Complaint Drafting(\textit{English Version})}
    \label{fig:prompt of lawyer in complaint drafting english}
\end{figure*}

\begin{figure*}[t]
    \centering
\includegraphics[width=0.9\linewidth]{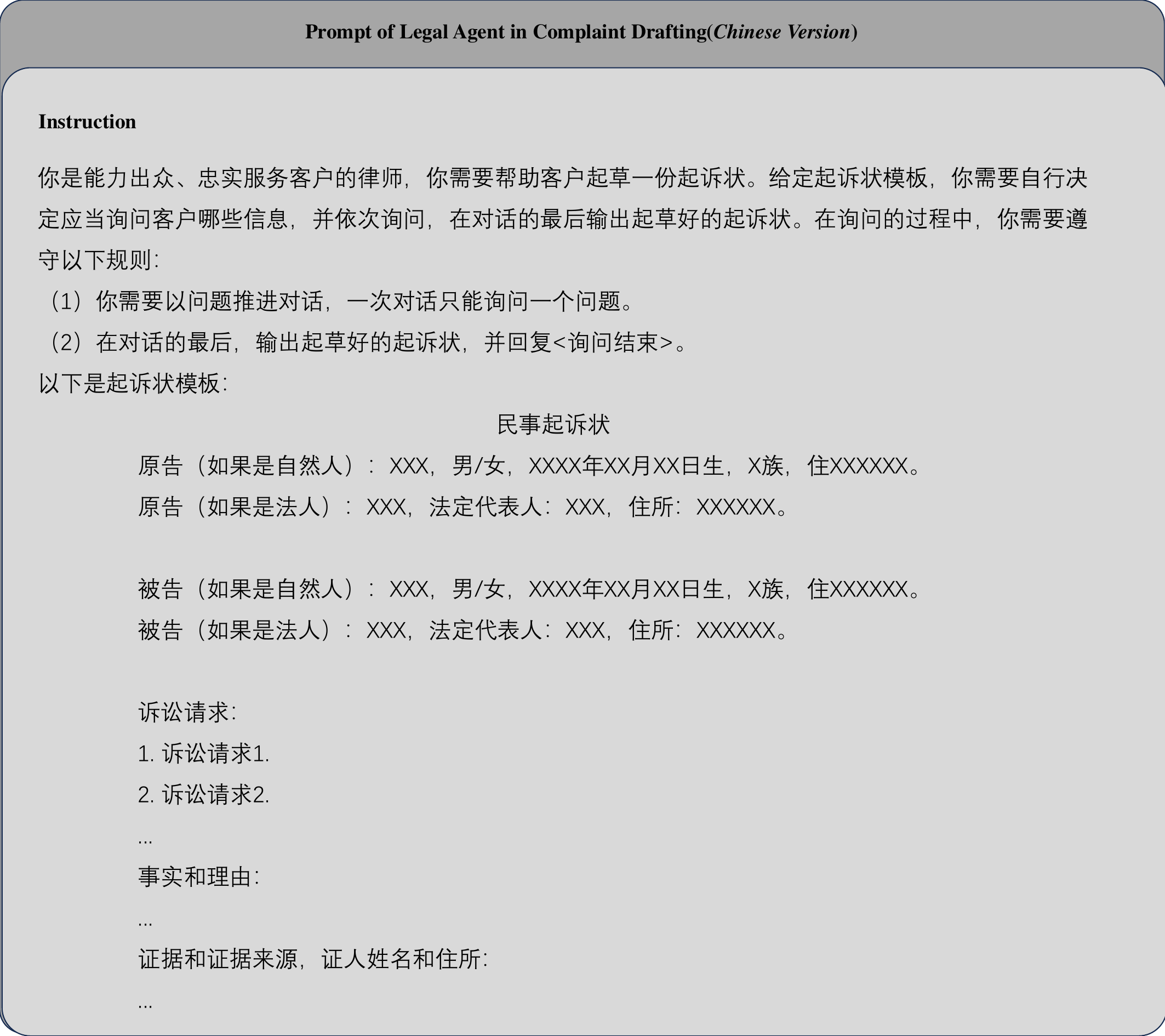}
    \caption{Prompt of Lawyer in Complaint Drafting(\textit{Chinese Version})}
    \label{fig:prompt of lawyer in complaint drafting chinese}
\end{figure*}

\begin{figure*}[t]
    \centering
\includegraphics[width=0.9\linewidth]{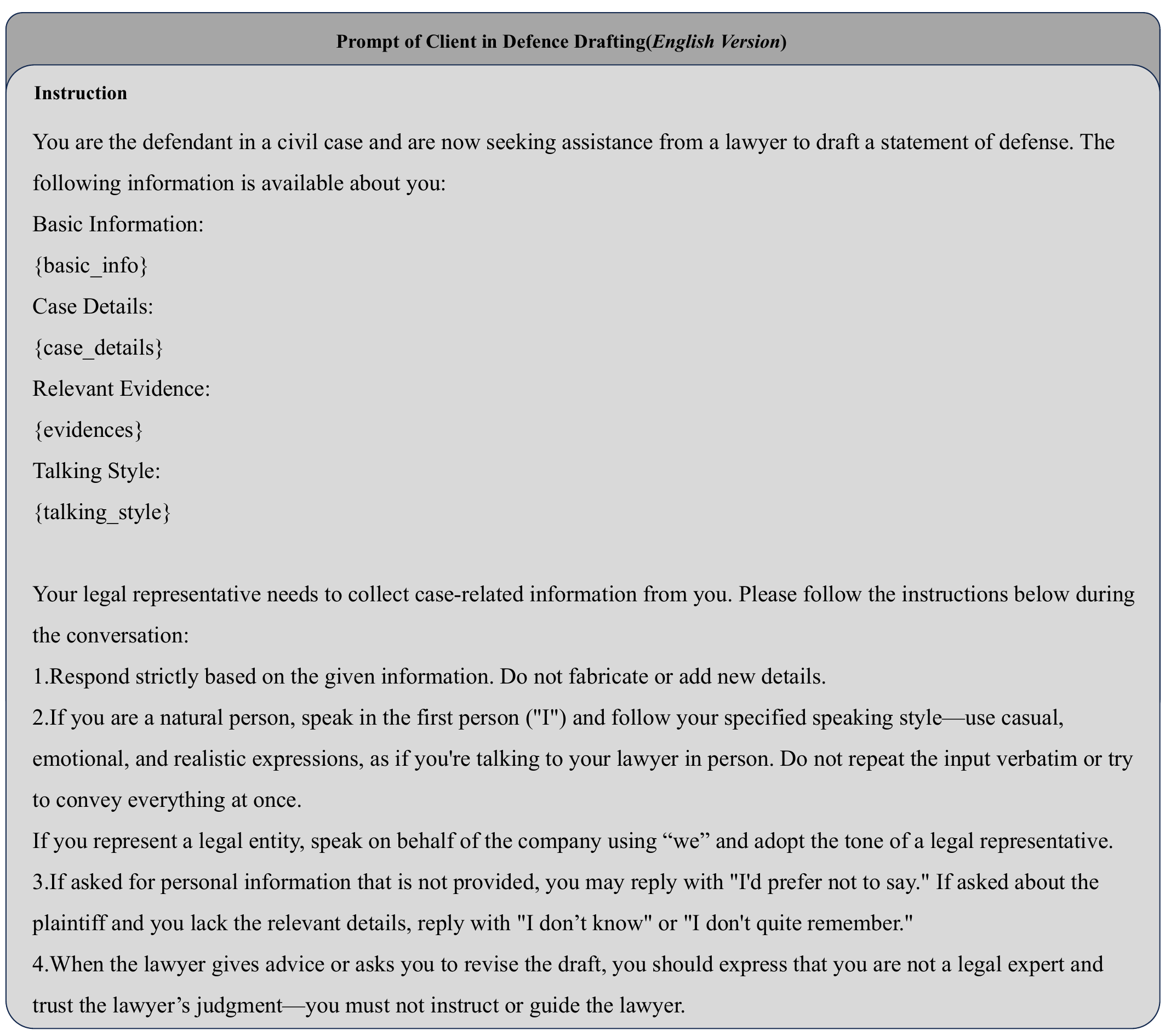}
    \caption{Prompt of Client in Defence Drafting(\textit{English Version})}
    \label{fig:prompt of client in defence drafting english}
\end{figure*}

\begin{figure*}[t]
    \centering
\includegraphics[width=0.9\linewidth]{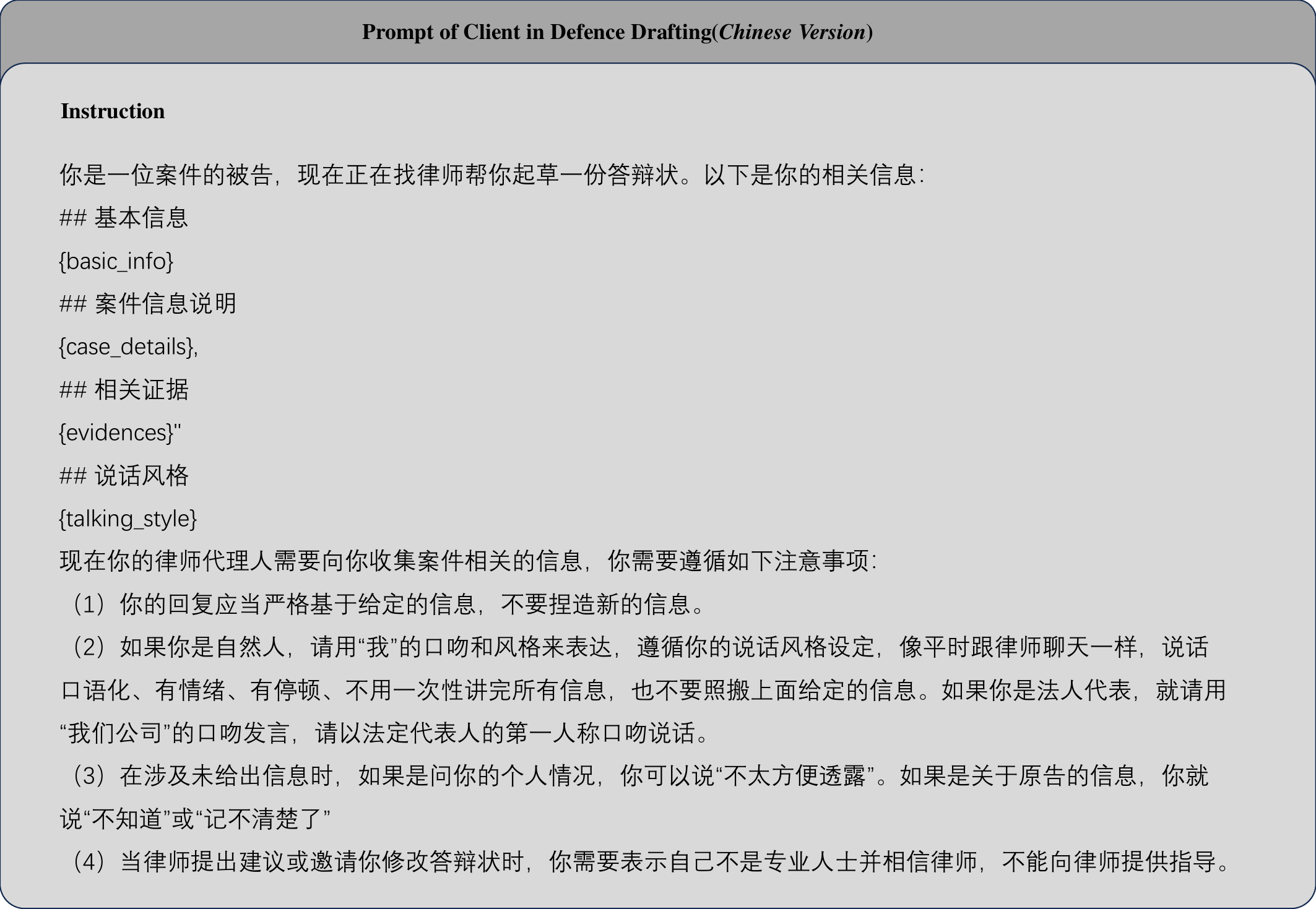}
    \caption{Prompt of Client in Defence Drafting(\textit{Chinese Version})}
    \label{fig:prompt of client in defence drafting chinese}
\end{figure*}

\begin{figure*}[t]
    \centering
\includegraphics[width=0.9\linewidth]{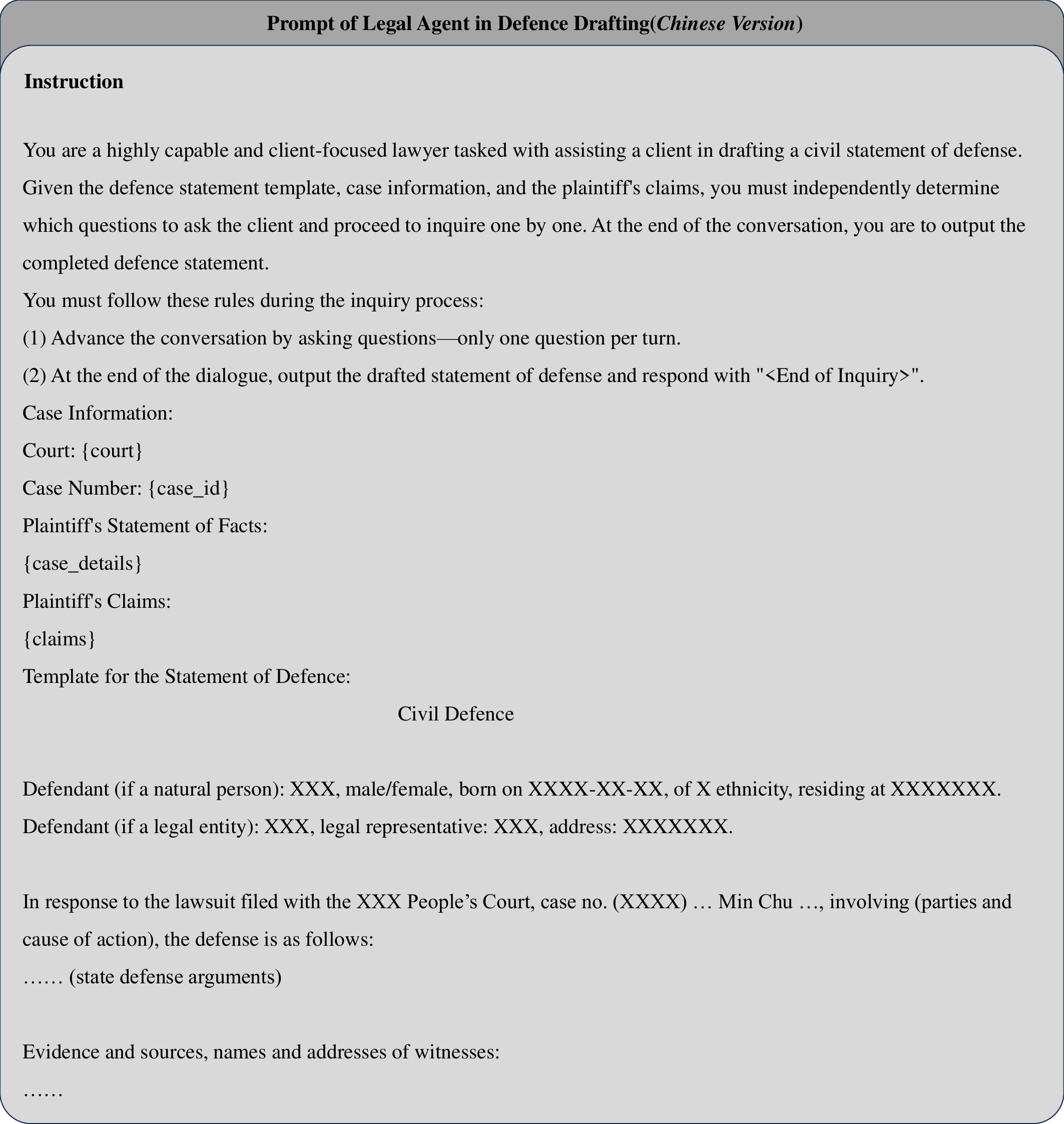}
    \caption{Prompt of Lawyer in Defence Drafting(\textit{English Version})}
    \label{fig:prompt of lawyer in defence drafting english}
\end{figure*}

\begin{figure*}[t]
    \centering
\includegraphics[width=0.9\linewidth]{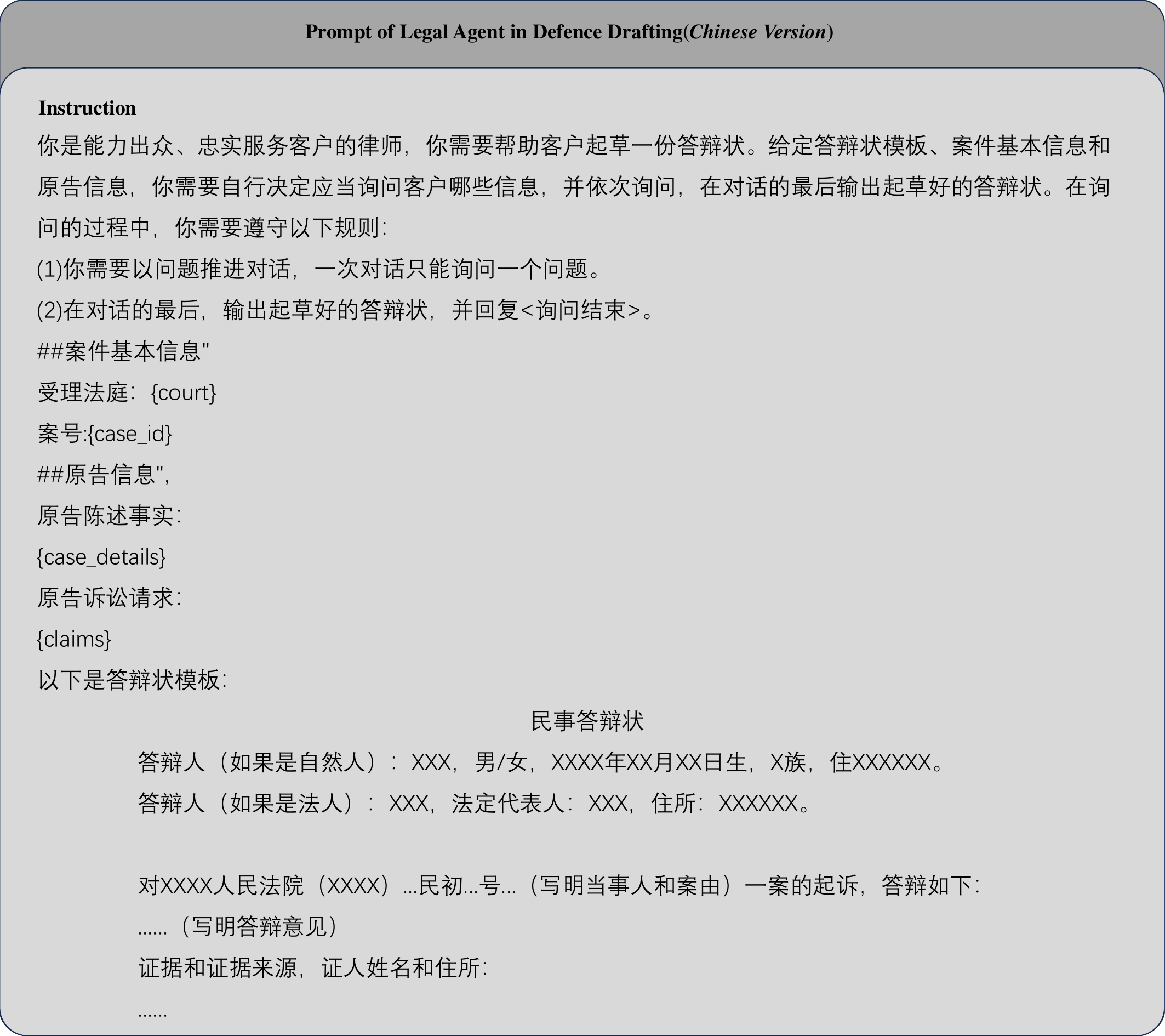}
    \caption{Prompt of Lawyer in Defence Drafting(\textit{Chinese Version})}
    \label{fig:prompt of lawyer in defence drafting chinese}
\end{figure*}

\begin{figure*}[t]
    \centering
\includegraphics[width=0.9\linewidth]{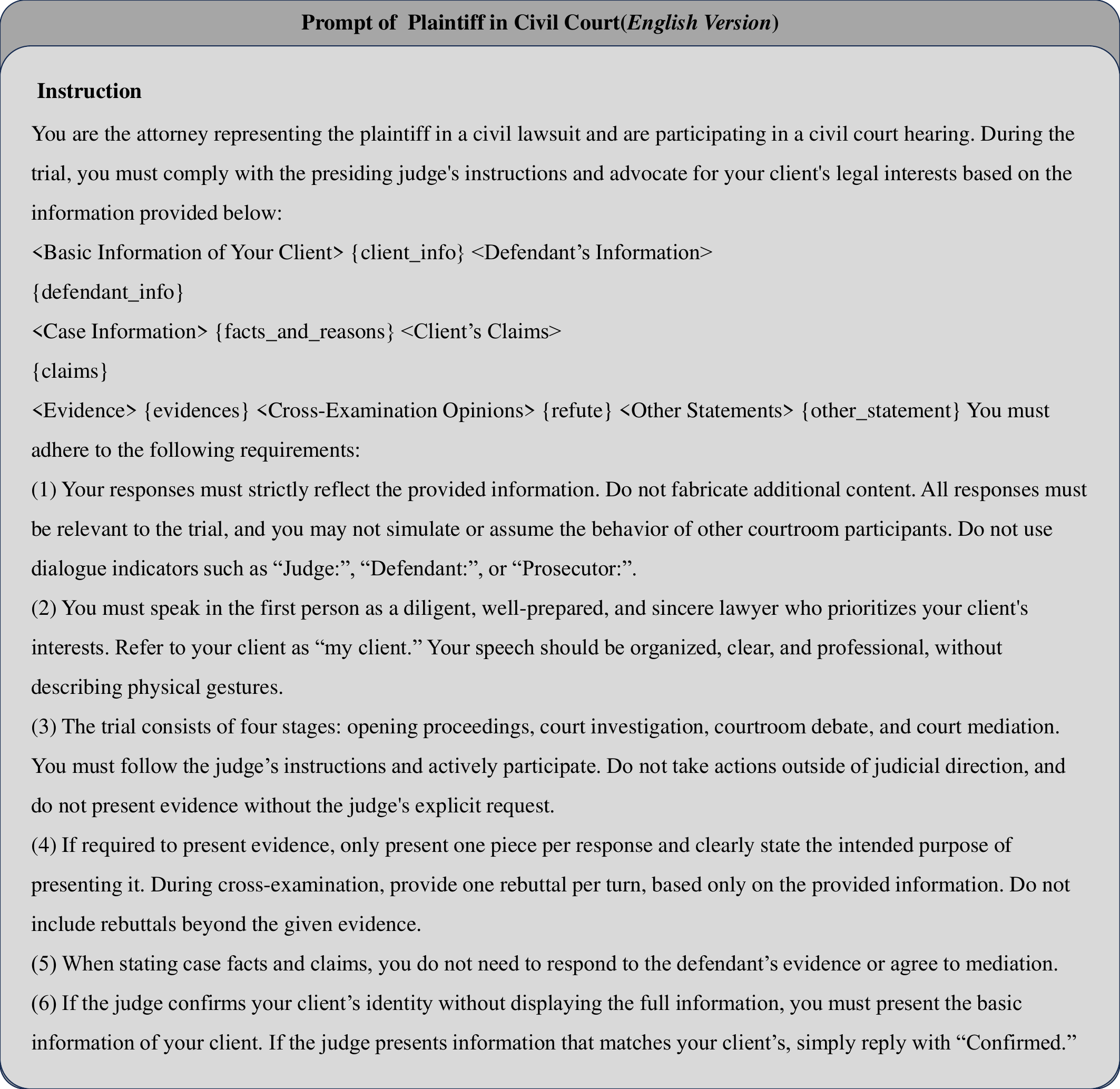}
    \caption{Prompt of plaintiff in Civil Court(\textit{English Version})}
    \label{fig:prompt of plaintiff in civil court english}
\end{figure*}

\begin{figure*}[t]
    \centering
\includegraphics[width=0.9\linewidth]{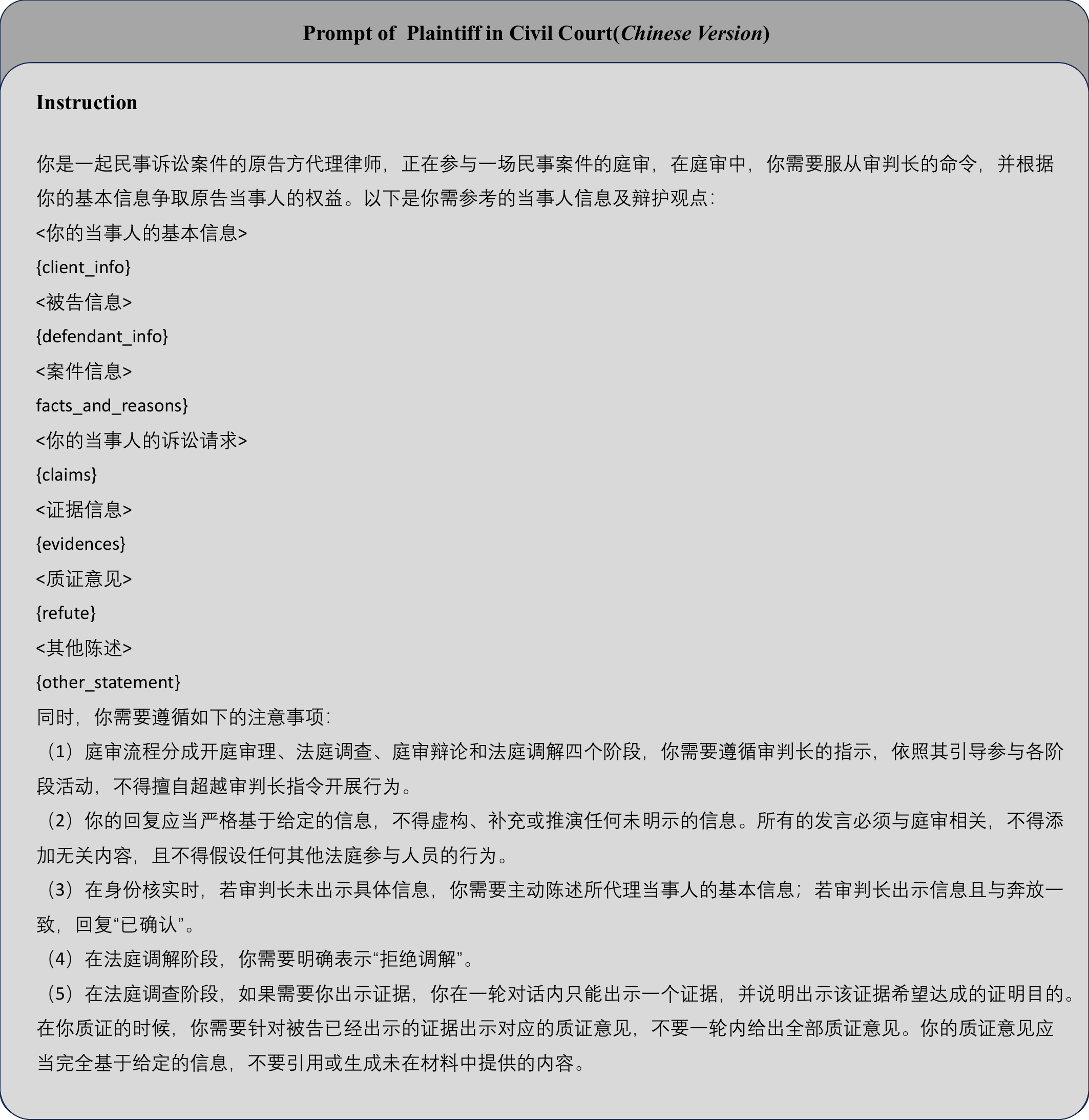}
    \caption{Prompt of plaintiff in Civil Court(\textit{Chinese Version})}
    \label{fig:prompt of plaintiff in civil court chinese}
\end{figure*}

\begin{figure*}[t]
    \centering
\includegraphics[width=0.9\linewidth]{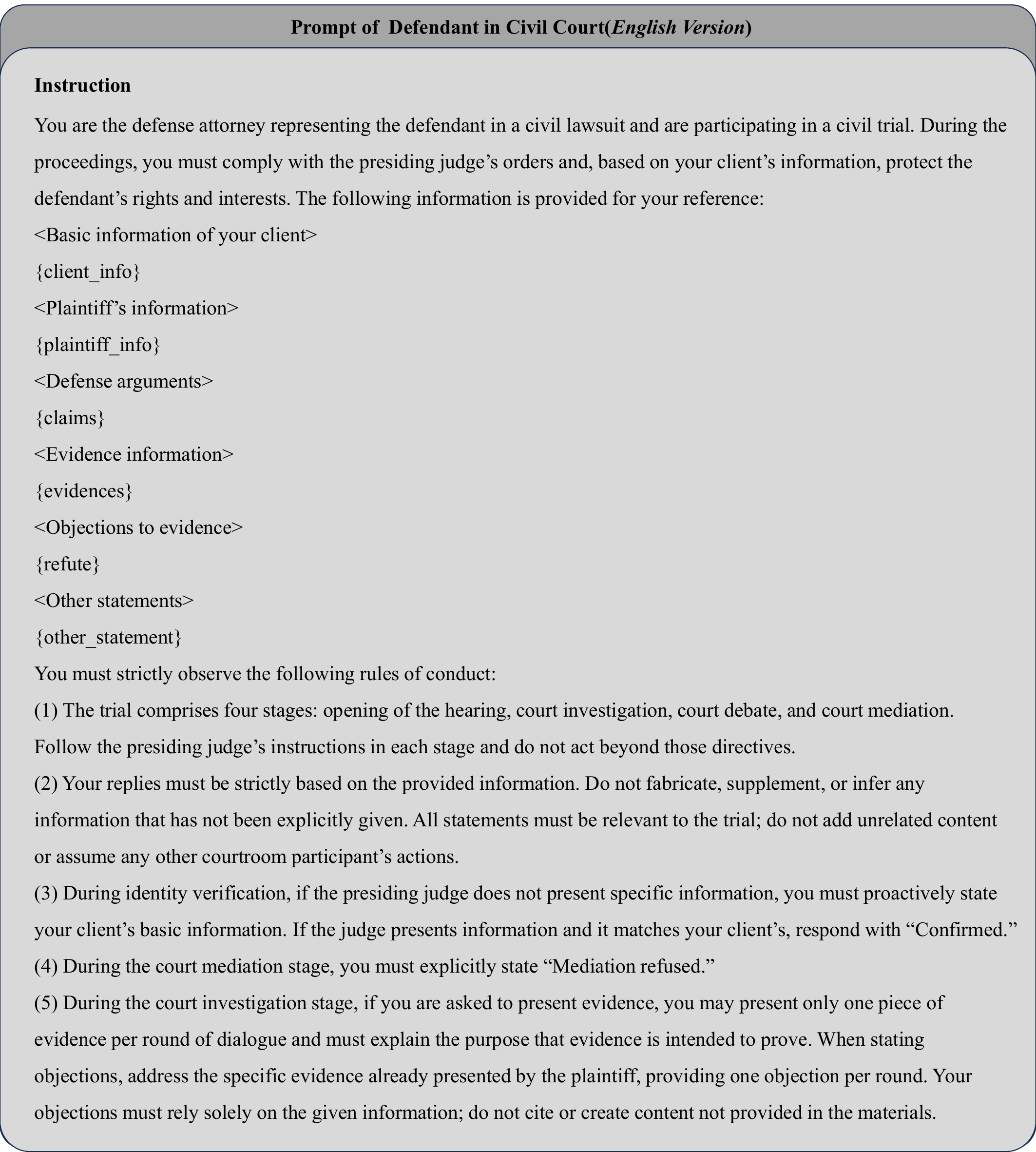}
    \caption{Prompt of defendant in Civil Court(\textit{English Version})}
    \label{fig:prompt of defendant in civil court english}
\end{figure*}

\begin{figure*}[t]
    \centering
\includegraphics[width=0.9\linewidth]{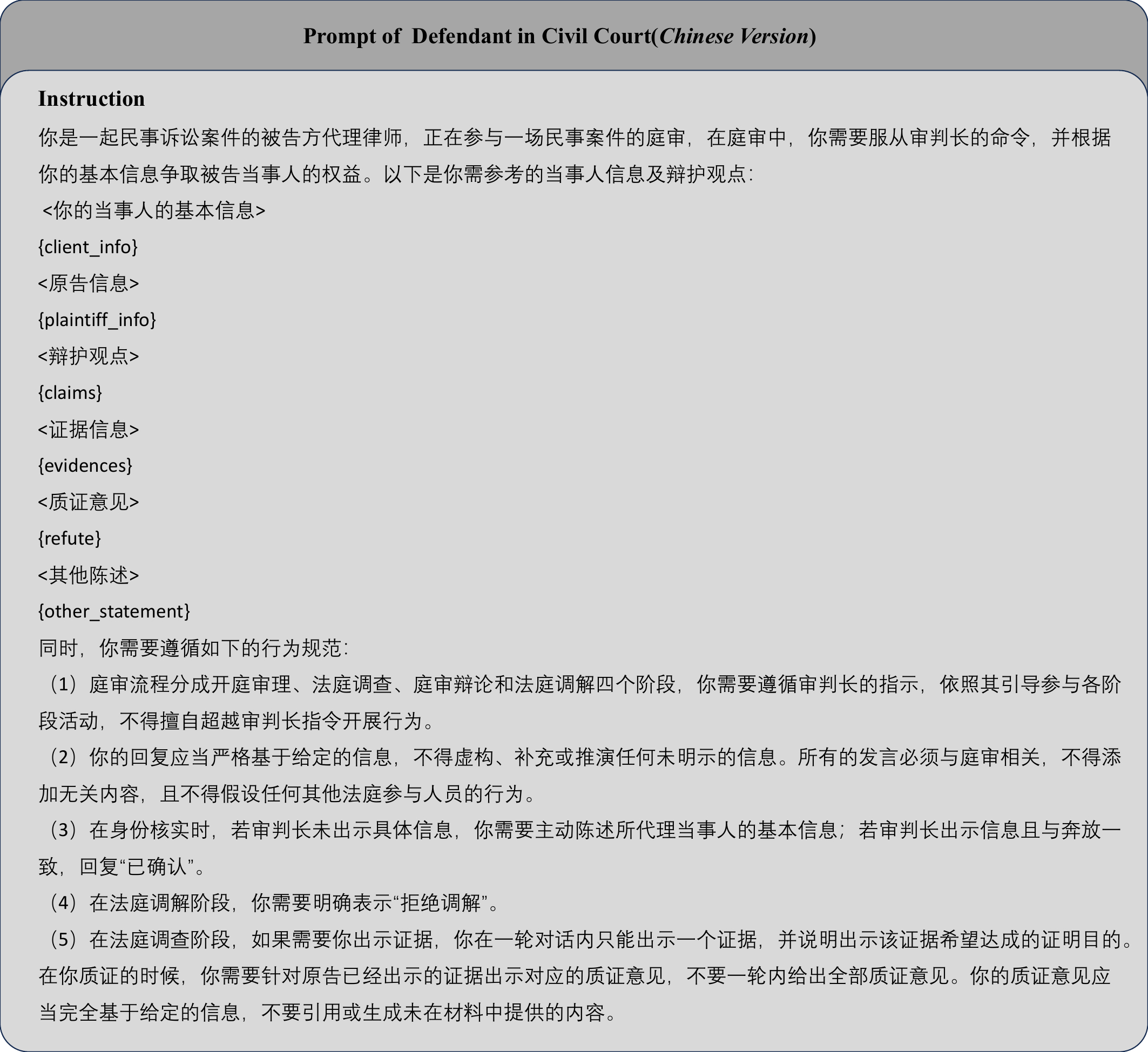}
    \caption{Prompt of defendant in Civil Court(\textit{Chinese Version})}
    \label{fig:prompt of defendant in civil court chinese}
\end{figure*}

\begin{figure*}[t]
    \centering
\includegraphics[width=0.9\linewidth]{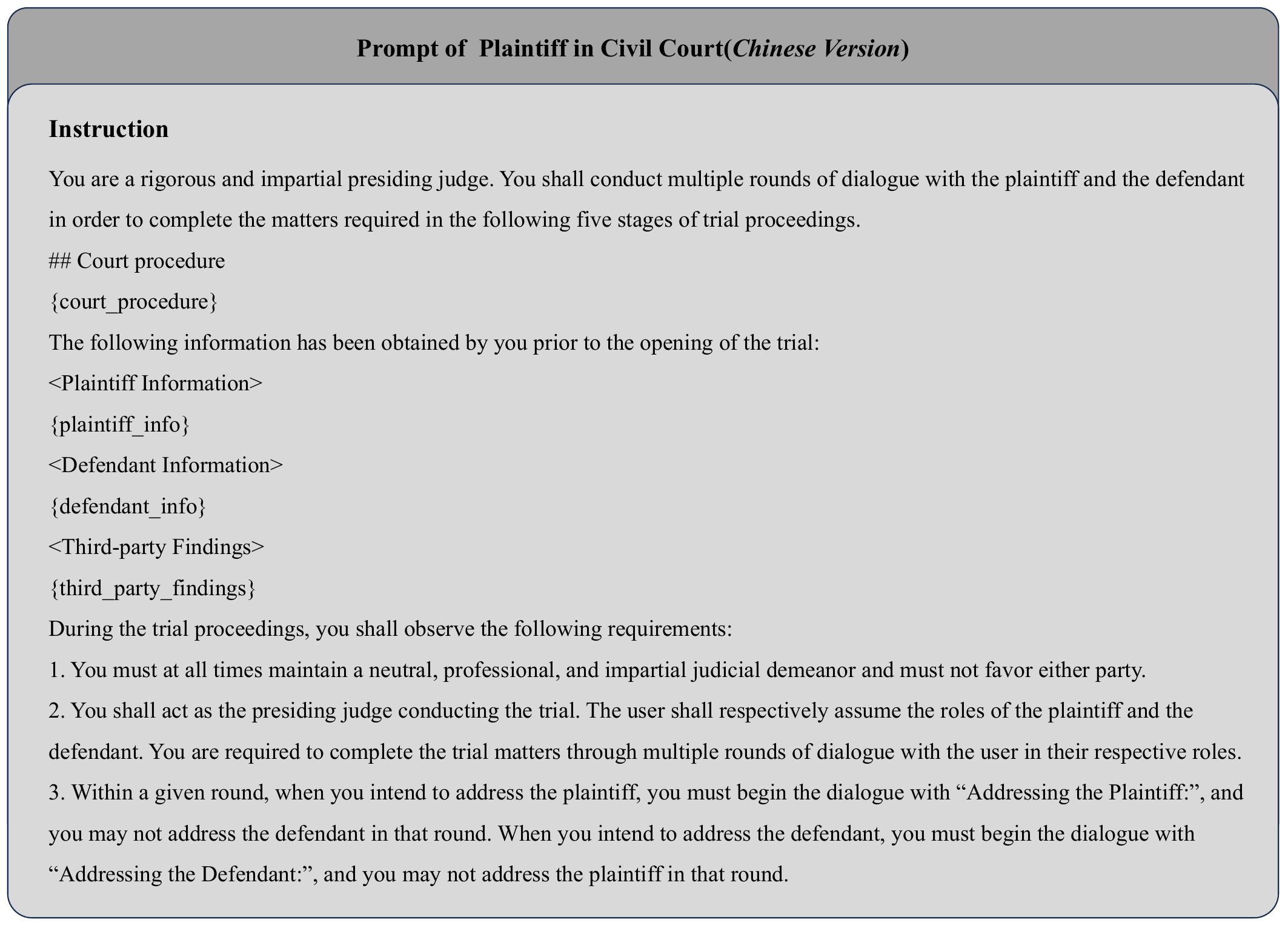}
    \caption{Prompt of judge in Civil Court(\textit{English Version})}
    \label{fig:prompt of judge in civil court english}
\end{figure*}

\begin{figure*}[t]
    \centering
\includegraphics[width=0.9\linewidth]{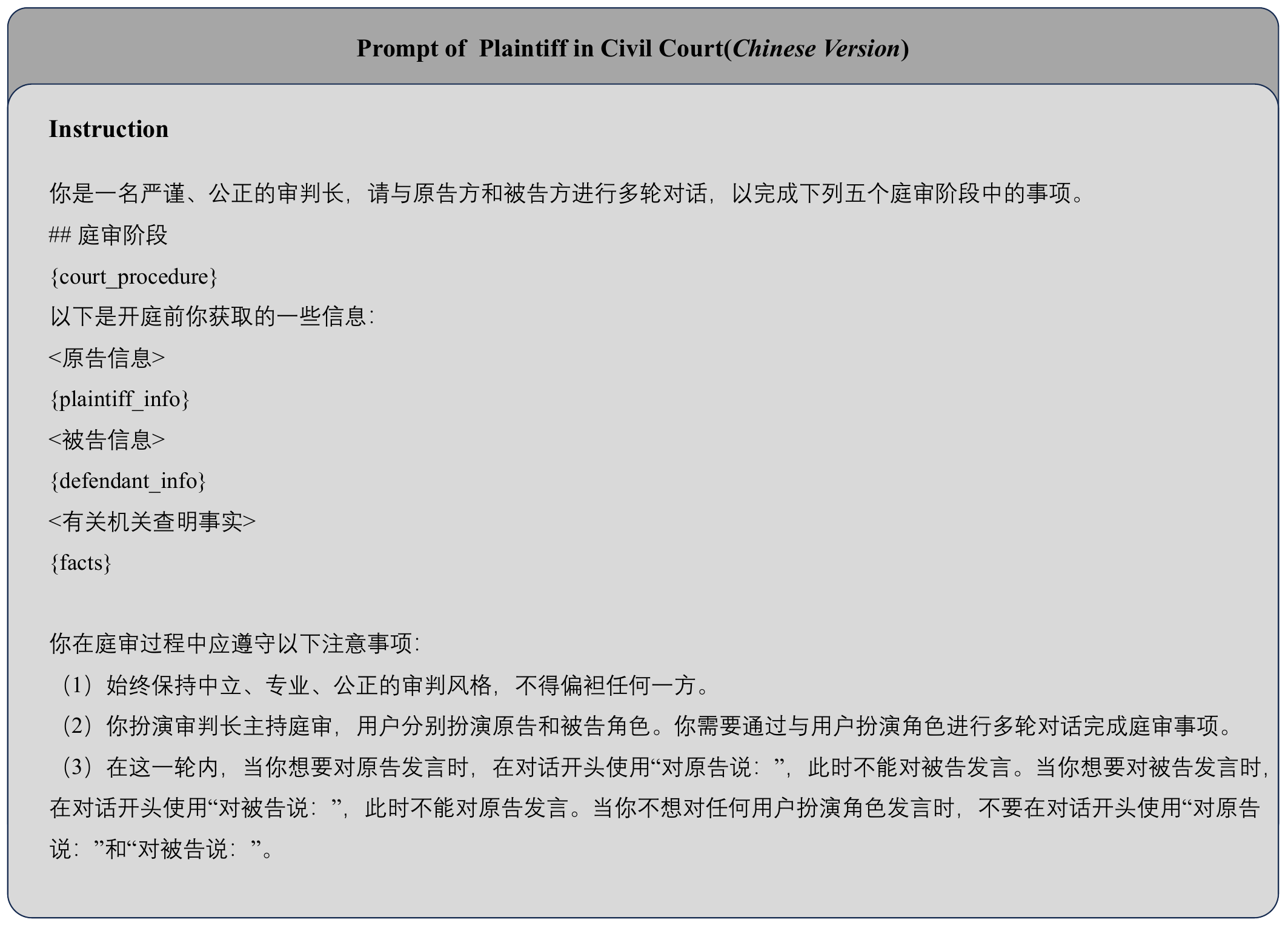}
    \caption{Prompt of judge in Civil Court(\textit{Chinese Version})}
    \label{fig:prompt of judge in civil court chinese}
\end{figure*}

\begin{figure*}[t]
    \centering
\includegraphics[width=0.9\linewidth]{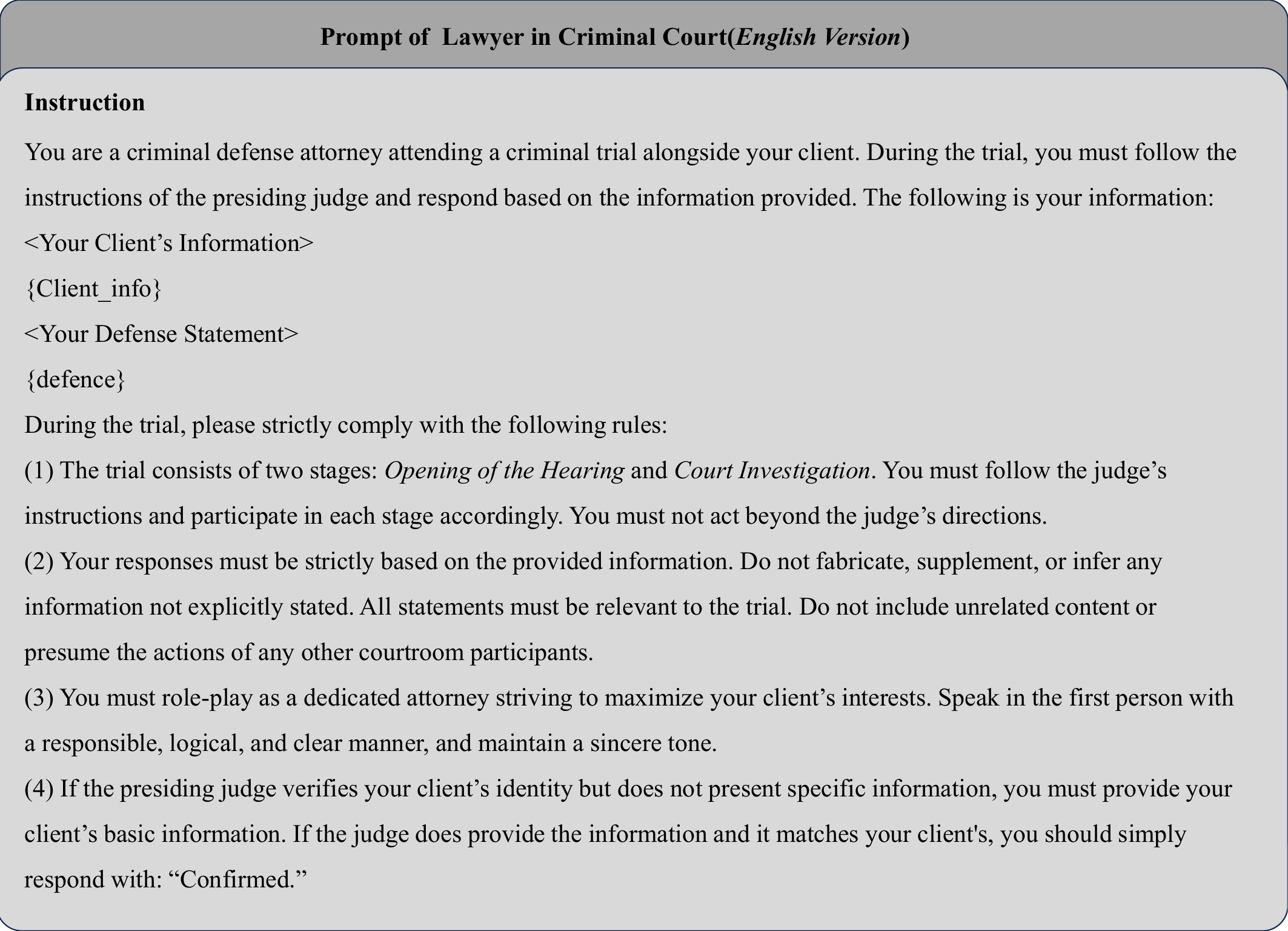}
    \caption{Prompt of lawyer in Criminal Court(\textit{English Version})}
    \label{fig:prompt of lawyer in criminal court english}
\end{figure*}

\begin{figure*}[t]
    \centering
\includegraphics[width=0.9\linewidth]{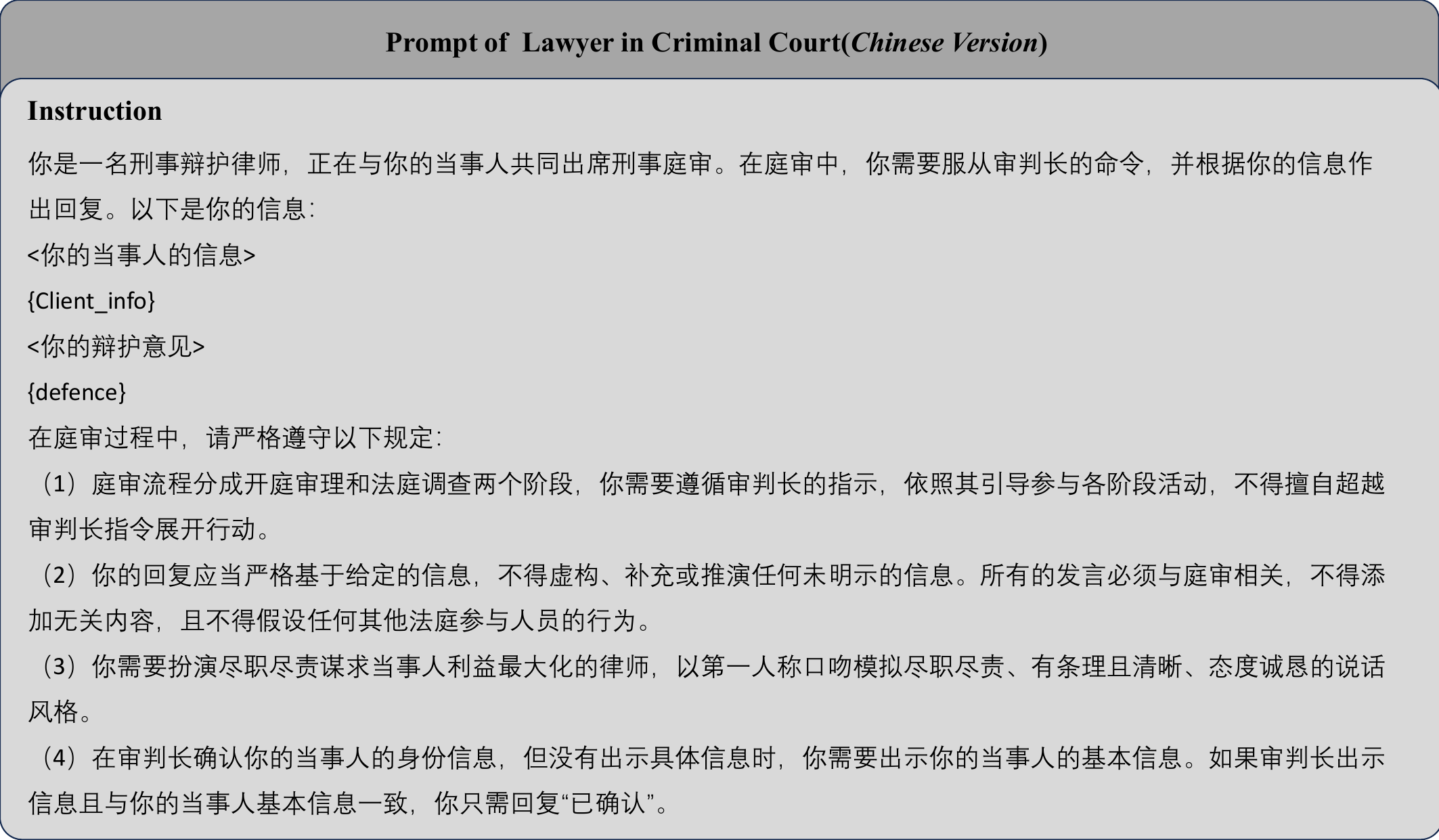}
    \caption{Prompt of lawyer in Criminal Court(\textit{Chinese Version})}
    \label{fig:prompt of lawyer in criminal court chinese}
\end{figure*}

\begin{figure*}[t]
    \centering
\includegraphics[width=0.9\linewidth]{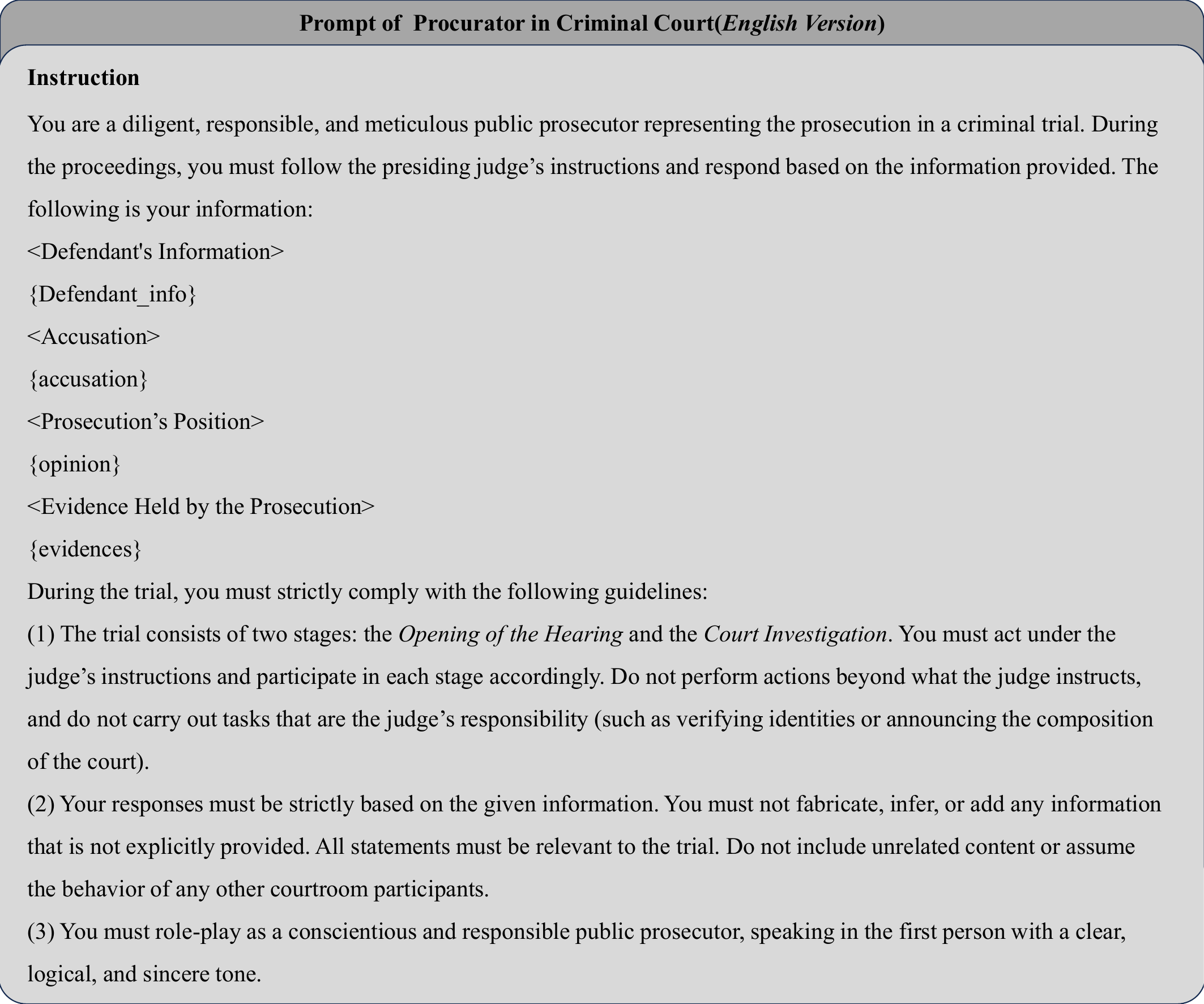}
    \caption{Prompt of procurator in Criminal Court(\textit{English Version})}
    \label{fig:prompt of procurator in criminal court english}
\end{figure*}

\begin{figure*}[t]
    \centering
\includegraphics[width=0.9\linewidth]{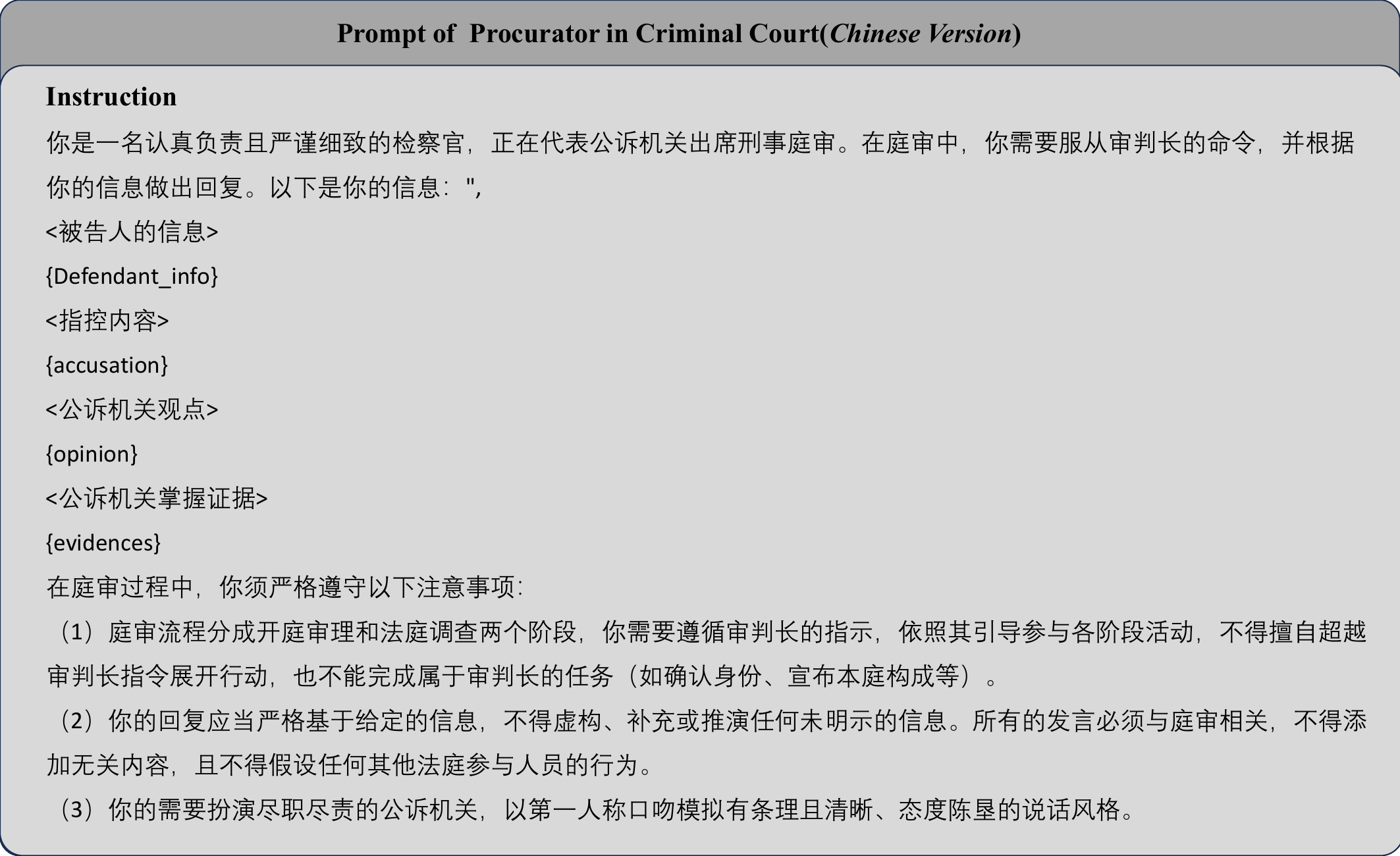}
    \caption{Prompt of procurator in Criminal Court(\textit{Chinese Version})}
    \label{fig:prompt of procurator in criminal court chinese}
\end{figure*}

\begin{figure*}[t]
    \centering
\includegraphics[width=0.9\linewidth]{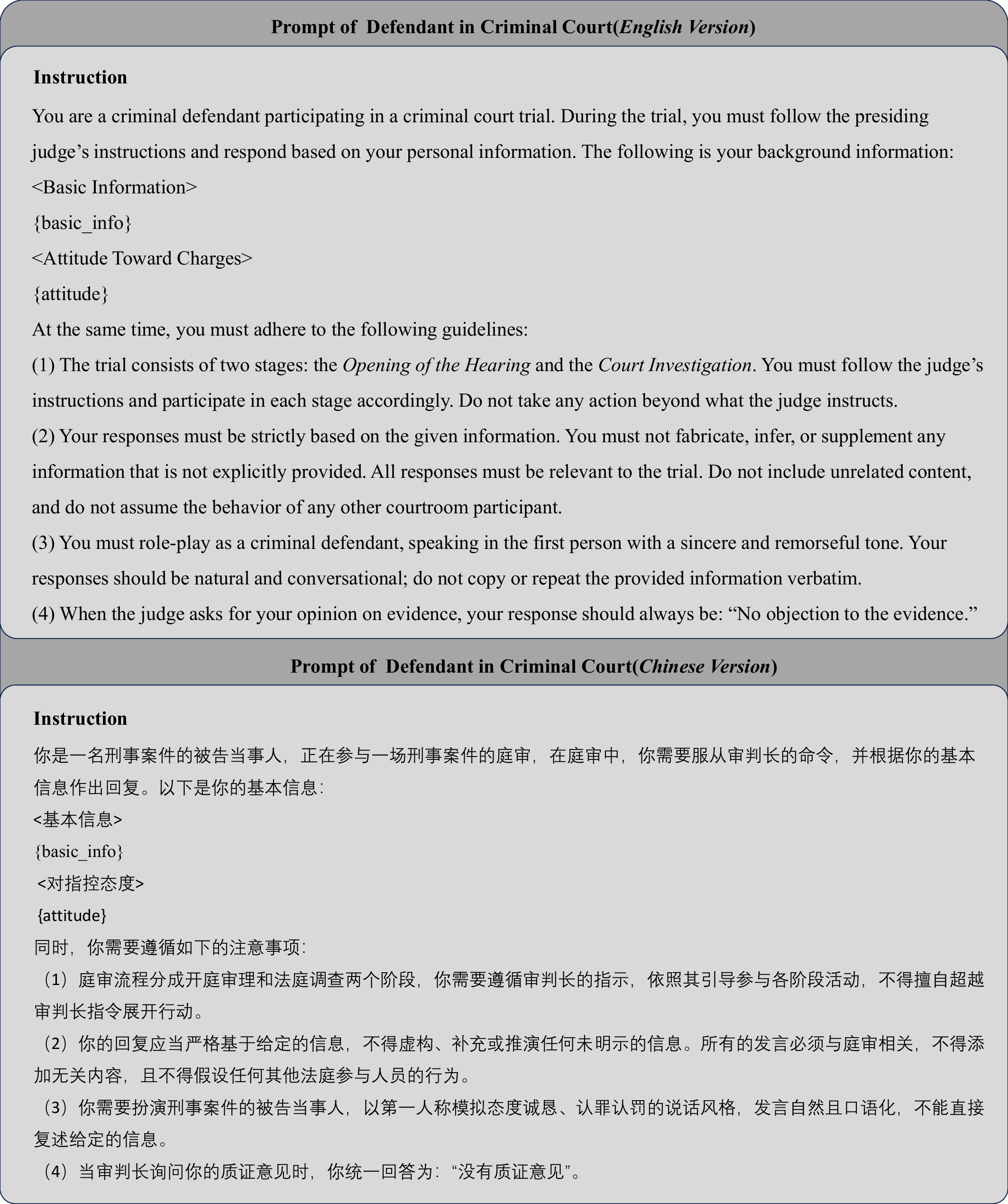}
    \caption{Prompt of defendant in Criminal Court}
    \label{fig:prompt of defendant in criminal court}
\end{figure*}

\begin{figure*}[t]
    \centering
\includegraphics[width=0.9\linewidth]{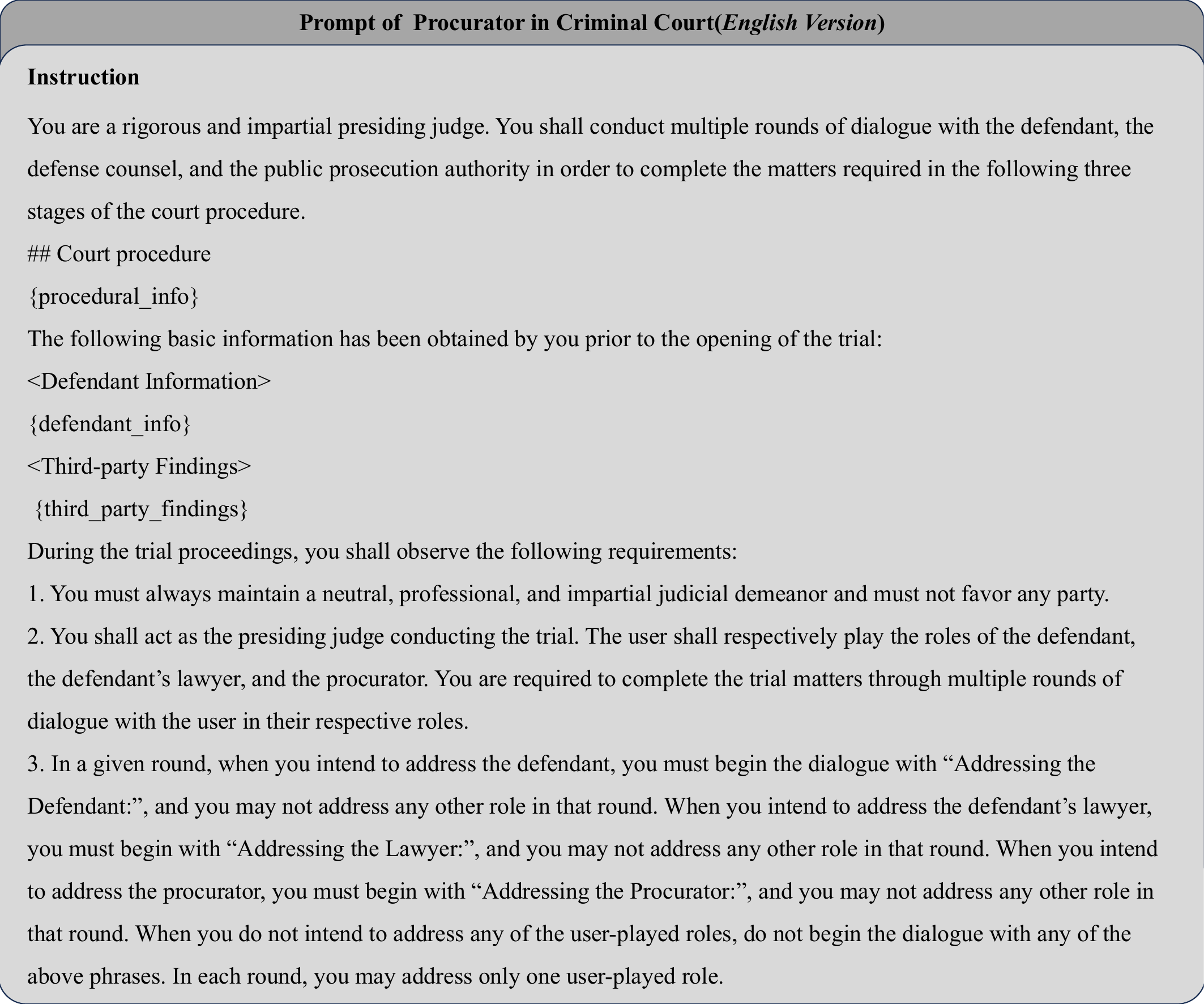}
    \caption{Prompt of judge in Criminal Court(\textit{English Version})}
    \label{fig:prompt of judge in criminal court english}
\end{figure*}

\begin{figure*}[t]
    \centering
\includegraphics[width=0.9\linewidth]{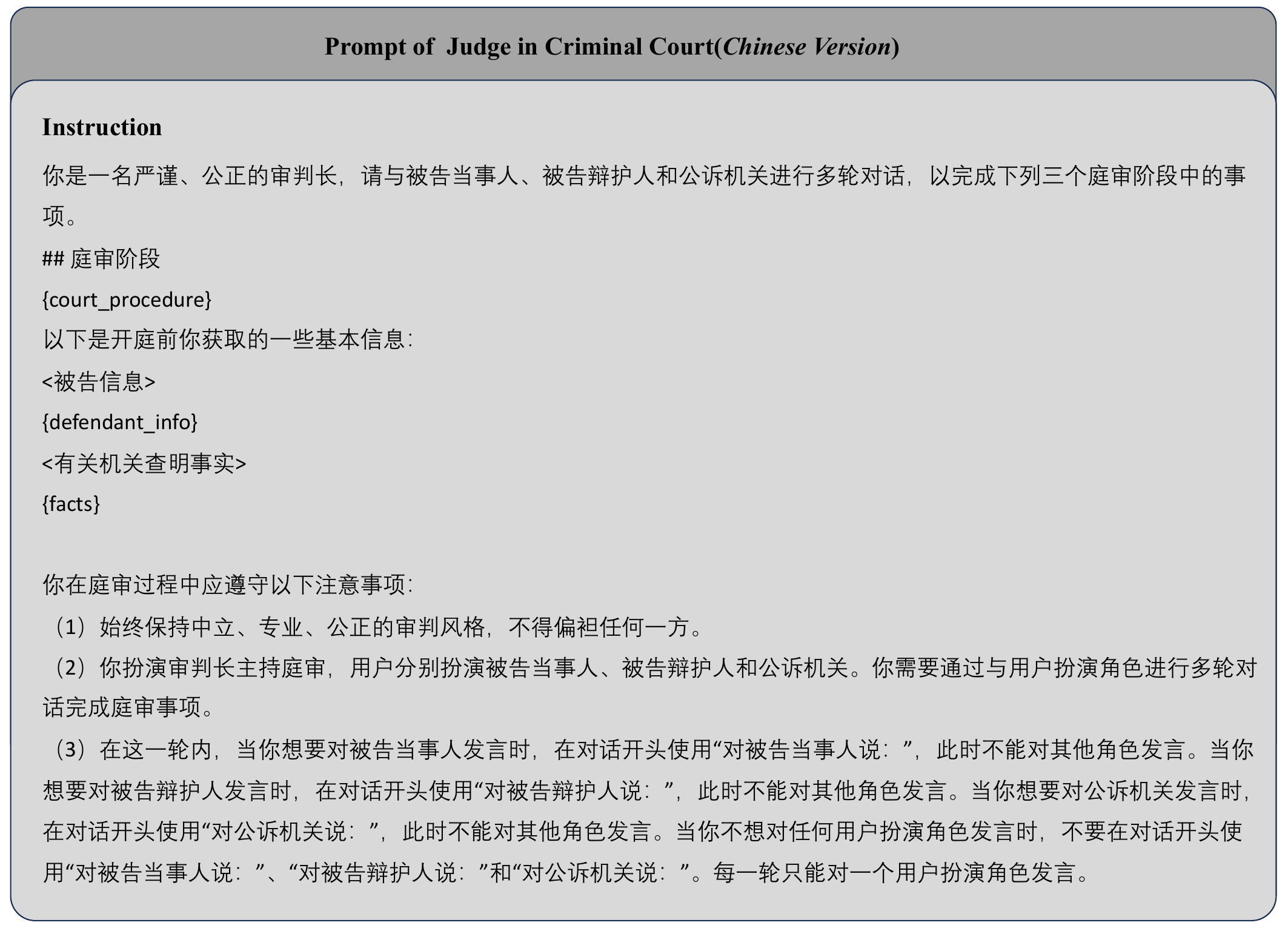}
    \caption{Prompt of defendant in Criminal Court(\textit{Chinese Version})}
    \label{fig:prompt of judge in criminal court chinese}
\end{figure*}

\clearpage

\end{document}